\documentclass[twoside,11pt]{article}

\usepackage{jair, theapa, rawfonts}
\usepackage{amsmath}
\usepackage{multirow}
\usepackage{bbm,url}
\usepackage{graphicx}
\usepackage{amssymb}
\usepackage{booktabs}
\usepackage{adjustbox}
\usepackage{xspace}
\usepackage{algpseudocode}
\usepackage{longtable}
\usepackage{makecell}
\usepackage{pgfplots} 

\usepackage{algorithm}
\usepackage{color, colortbl}
\usepackage{float}
\usepackage[caption = false]{subfig}
\usepackage{sectsty}
\allsectionsfont{\sloppy}

\usepackage{array}
\newcommand{\PreserveBackslash}[1]{\let\temp=\\#1\let\\=\temp}
\newcolumntype{C}[1]{>{\PreserveBackslash\centering}p{#1}}
\newcolumntype{R}[1]{>{\PreserveBackslash\raggedleft}p{#1}}
\newcolumntype{L}[1]{>{\PreserveBackslash\raggedright}p{#1}}

\newcommand{\secref}[2][]{Section#1~\ref{sec:#2}}

\newcommand{\tabref}[2][]{Table#1~\ref{tab:#2}}
\newcommand{\figref}[2][]{Figure#1~\ref{fig:#2}}

\newcommand{\model}[1]{\textsc{#1}\xspace}

\newcommand{\bertscore}{\model{BERTScore}}
\newcommand{\moverscore}{\model{MoverScore}}
\newcommand{\stsscore}{\model{STS-Score}}

\newcommand{\ffci}{\model{FFCI}}
\newcommand{\bert}{\model{BERT}}
\newcommand{\roberta}{\model{RoBERTa}}
\newcommand{\xlnet}{\model{XLNet}}
\newcommand{\gpt}{\model{GPT2}}
\newcommand{\bart}{\model{BART}}
\newcommand{\pegasus}{\model{PEGASUS}}

\definecolor{LightCyan}{rgb}{0.88,1,1}
\def\mybar#1{{\color{teal}\rule{8pt}{#1cm}}}

\newcommand{\precision}{\ensuremath{\mathcal{P}}}
\newcommand{\recall}{\ensuremath{\mathcal{R}}}
\newcommand{\fscore}{\ensuremath{\mathcal{F}}}

\newcommand{\rot}{\rotatebox[origin=c]{90}}
\newcommand{\x}{\checkmark}

\DeclareMathOperator{\METRIC}{METRIC}
\DeclareMathOperator{\NSP}{NSP}
\DeclareMathOperator{\AvgTop}{AvgTop-}
\newcommand{\AvgTopN}[1][n]{\ensuremath{\AvgTop#1}\xspace}

\newcommand{\RQ}[2][]{\textbf{[RQ#1]} \textit{#2}}

\jairheading{-}{2022}{}{07/2021}{03/2022}
\ShortHeadings{FFCI: A Framework for Interpretable Automatic Evaluation of Summarization}
{Koto, Baldwin, \& Lau}
\firstpageno{1}

\begin{document}
\emergencystretch 2em

\title{FFCI: A Framework for Interpretable Automatic \\Evaluation of Summarization}

\author{\name Fajri Koto \email ffajri@student.unimelb.edu.au \\
       \name Timothy Baldwin \email tb@ldwin.net\\
       \name Jey Han Lau \email jeyhan.lau@gmail.com \\
       \addr School of Computing and Information Systems\\
       The University of Melbourne\\
       Victoria 3010, Australia}


\maketitle

\begin{abstract}
In this paper, we propose \ffci, a framework for fine-grained
summarization evaluation that comprises four elements: faithfulness
(degree of factual consistency with the source), focus (precision of
summary content relative to the reference), coverage (recall of summary
content relative to the reference), and inter-sentential coherence
(document fluency between adjacent sentences). We construct a novel
dataset for focus, coverage, and inter-sentential coherence, and develop automatic methods for evaluating each of the four
dimensions of \ffci based on cross-comparison of evaluation metrics and
model-based evaluation methods, including question answering (QA)
approaches, semantic textual similarity (STS), next-sentence prediction (NSP), and scores derived
from 19 pre-trained language models.  We then apply the developed
metrics in evaluating a broad range of summarization models across two
datasets, with some surprising findings.
\end{abstract}


\section{Introduction}
\label{sec:Introduction}

Remarkable advances in abstractive summarization in recent years have unfortunately not been accompanied by commensurate improvements in automatic evaluation metrics. Most recent studies \shortcite{nallapati-etal-2016-abstractive,see-etal-2017-get,gehrmann-etal-2018-bottom,liu-lapata-2019-text,zhang2020pegasus,lewis2019bart} continue to rely on ROUGE \shortcite{lin-2004-rouge}, a lexical-overlap metric that is not capable of detecting paraphrases in abstractive summaries relative to reference summaries, with the only real mainstream alternative being manual evaluation \shortcite{hsu-etal-2018-unified,chen-bansal-2018-fast,hardy-vlachos-2018-guided,celikyilmaz-etal-2018-deep,krishna-srinivasan-2018-generating}. 

We identify four key dimensions across which to evaluate summaries: (1) \underline{f}aithfulness \shortcite{maynez-etal-2020-faithfulness}, (2) \underline{f}ocus, (3) \underline{c}overage, and (4) \underline{i}nter-sentential coherence; we label the combined approach ``\ffci''. Faithfulness measures the degree of factual consistency (and lack of hallucination) relative to the source document, and is especially important for abstractive methods. The other three aspects are inspired by manual evaluation in previous work \shortcite{peyrard-gurevych-2018-objective,hsu-etal-2018-unified,celikyilmaz-etal-2018-deep,narayan-etal-2018-dont,chen-bansal-2018-fast}, as summarized by \shortciteA{hardy-etal-2019-highres}.

We first revisit recent work on \textit{faithfulness}, and propose a simpler automatic evaluation scheme. Recent work has used question generation (QG) and question answering (QA) to evaluate faithfulness \shortcite{wang2020asking,durmus-etal-2020-feqa}. However, we argue that this approach is computationally expensive\footnote{For instance, to evaluate the 11,490 CNN/DailyMail test set \shortcite{hermann2015teaching} requires the generation of roughly 229,800 questions and answers.}
and critically depends on resources that are often unavailable in languages other than English. As an alternative, we extend the experiments of \shortciteA{zhang2020bertscore} and \shortciteA{durmus-etal-2020-feqa} in investigating scores from a broad range of pre-trained language models (computed between summary and article) and find them to be more reliable than QA-based methods.

\begin{figure}
	\begin{center}
		\includegraphics[width=4in]{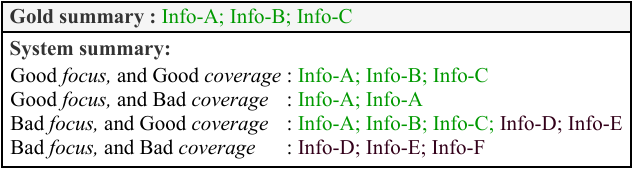}
	\end{center} 
	\caption{Illustration of focus and coverage.}
	\label{fig:foc_cov}
\end{figure}

Secondly, we propose \textit{focus} and \textit{coverage} relative to 
the reference summary. Both assess semantic equivalence, with focus 
evaluating the proportion of important information in the generated 
summary (= precision), and coverage evaluating the degree of salient 
information in the reference summary that the generated summary contains 
(= recall). In \figref{foc_cov}, we illustrate different scenarios for 
focus and coverage of system summaries.

Lastly, we address the automatic evaluation of linguistic quality in multi-sentence summaries. Previous work has looked at aspects including readability, fluency, and clarity \shortcite{hardy-etal-2019-highres}, but we argue that \textit{inter-sentential coherence} is more important for evaluating abstractive summaries for two reasons. First, modern pre-trained language models are highly adept at generating fluent sentences, but global coherence beyond the sentence is not a given. Second, inter-sentential coherence subsumes sub-sentence coherence, as disfluent sentences will break the global discourse coherence.

To summarize, our contributions are: (1) we release an annotated dataset for evaluating focus, coverage, and inter-sentential coherence; (2) for faithfulness, focus and coverage, we benchmark traditional metrics such as ROUGE, METEOR, and BLEU with model-based metrics, including question-answering (QA) methods, semantic textual similarity (STS), FactCC \shortcite{kryscinski-etal-2020-evaluating}, and scores from 19 different pre-trained language models; (3) we adapt the next sentence prediction (NSP) for evaluating inter-sentential coherence; and (4) we re-evaluate a broad range of contemporary summarization models over CNN/DailyMail and XSUM based on \ffci, with a number of surprising findings not captured by ROUGE. Data and code used in this paper can be accessed at \url{https://github.com/fajri91/ffci}.

\section{Related Work}
\label{sec:rel}





\subsection{Aspects on Summarization Evaluation}
\label{sec:rel_aspect}

Automatic evaluations of language generation systems have been based on the comparison of reference and system-generated text. BLEU \shortcite{papineni-etal-2002-bleu} is a precision-based metric in machine translation task, while ROUGE \shortcite{lin-2004-rouge} is the de facto metric for summarization systems  \shortcite{see-etal-2017-get,liu-lapata-2019-text,zhang2020pegasus}. 
In the other text generation tasks such as caption generation \shortcite{xu2015show} and question generation \shortcite{du-etal-2017-learning}, CIDEr \shortcite{vedantam2015cider} and SPICE \shortcite{anderson2016spice} are used to complement BLEU and ROUGE. Recently, pre-trained embedding based evaluation metrics such as \bertscore \shortcite{zhang2020bertscore} and \moverscore \shortcite{zhao-etal-2019-moverscore} have also been proposed.


{\setlength{\tabcolsep}{1.pt}
	\footnotesize
	\vspace{0.5cm}
	\begin{longtable}[]{L{5cm}|C{0.45cm}C{0.45cm}C{0.45cm}C{0.45cm}C{0.45cm}|C{0.45cm}|C{0.45cm}C{0.45cm}C{0.45cm}C{0.45cm}C{0.45cm}C{0.45cm}|C{0.45cm}C{0.45cm}|C{0.45cm}C{0.45cm}C{0.45cm}C{0.45cm}|C{0.45cm}}
		
		\hline
		\multirow{2}[27]{*}{\bf Paper} & \multicolumn{5}{c|}{\bf Automatic} & \multirow{2}{*}{\rot{\bf \medspace  No manual eval \medspace}} & \multicolumn{13}{c}{\bf Manual} \\
		\cline{2-6}\cline{8-20}
		& \rot{\bf ROUGE} & \rot{\bf METEOR} & \rot{\bf BLEU} & \rot{\bf BERTScore} & \rot{\bf \moverscore} &  & \rot{\bf Faithfulness} & \rot{\bf \bf Recall} & \rot{\bf Precision} & \rot{\bf Relevance} & \rot{\bf Coherence} & \rot{\bf Fluency} & \rot{\bf Relative} & \rot{\bf Absolute} & \rot{\bf SCU} & \rot{\bf reference} & \rot{\bf article} & \rot{\bf ref+article} & \rot{\bf \medspace Quality control\medspace}\\
		\hline
		\endhead
		\hline
		\multirow{2}[27]{*}{\bf Paper} & \multicolumn{5}{c|}{\bf Automatic} & \multirow{2}{*}{\rot{\bf \medspace  No manual eval \medspace}} & \multicolumn{13}{c}{\bf Manual} \\
		\cline{2-6}\cline{8-20}
		& \rot{\bf ROUGE} & \rot{\bf METEOR} & \rot{\bf BLEU} & \rot{\bf BERTScore} & \rot{\bf \moverscore} &  & \rot{\bf Faithfulness} & \rot{\bf \bf Recall} & \rot{\bf Precision} & \rot{\bf Relevance} & \rot{\bf Coherence} & \rot{\bf Fluency} & \rot{\bf Relative} & \rot{\bf Absolute} & \rot{\bf SCU} & \rot{\bf reference} & \rot{\bf article} & \rot{\bf ref+article} & \rot{\bf \medspace Quality control\medspace}\\
		\hline
		\endhead
		\hline
		\multicolumn{20}{c}{}\\\multicolumn{20}{c}{}\\
		\caption{Evaluation methods used in previous work (split over 3 pages)}
		\label{tab:prev_works}
		\endfoot
		\shortciteA{see-etal-2017-get} & \x & \x & & & & \x & & & & & & & & & & & & & \\
		\shortciteA{yang-etal-2017-detecting} & \x & & & & & & & \x & & & & & \x & & & \x & & &  \\
		\shortciteA{lin-etal-2018-global} & \x & & & & & \x & & & & & & & & & & & & &  \\
		\shortciteA{cohan-etal-2018-discourse} & \x & & & & & \x & & & & & & & & & & & & &  \\
		\shortciteA{liao-etal-2018-abstract} & \x & & & & & \x & & & & & & & & & & & & &  \\
		\shortciteA{kedzie-etal-2018-content} & \x & & & & & \x & & & & & & & & & & & & &  \\
		\shortciteA{amplayo-etal-2018-entity} & \x & & & & & & & & & \x & & \x & \x & & & & \x & &  \\
		\shortciteA{jadhav-rajan-2018-extractive} & \x & & & & & \x & & & & & & & & & & & & &  \\
		\shortciteA{li-etal-2018-guiding} & \x & & & & & \x & & & & & & & & & & & & &  \\
		\shortciteA{pasunuru-bansal-2018-multi} & \x & \x & & & & \x & & & & & & & & & & & & &  \\
		\shortciteA{cao-etal-2018-retrieve} & \x & & & & & \x & & & & & & & & & & & & &  \\
		\shortciteA{sakaue-etal-2018-provable} & \x & & & & & \x & & & & & & & & & & & & & \\
		\shortciteA{celikyilmaz-etal-2018-deep} & \x & & & & & & & \x & \x & & \x & & \x & \x & & & \x & & \\
		\shortciteA{chen-bansal-2018-fast} & \x & \x & & & & & \x & & & \x & \x & \x & \x & & & & & \x & \\
		\shortciteA{guo-etal-2018-soft} & \x & \x & & & & & & & & \x & \x & \x & \x & & & & & \x & \\
		\shortciteA{hardy-vlachos-2018-guided} & \x & & \x & & & & & & & & & \x & & \x & & & & & \\
		\shortciteA{hsu-etal-2018-unified} & \x & & & & & & & \x & \x & & & \x & & \x & & & \x & & \x \\
		\shortciteA{krishna-srinivasan-2018-generating} & \x & & & & & & & \x & & & & & \x & & & & \x & & \x \\
		\shortciteA{kryscinski-etal-2018-improving} & \x & & & & & & & \x & & & & \x & & \x & & & \x & & \\
		\shortciteA{li-etal-2018-ensure} & \x & & & & & & \x & & & & & & & \x & & & & & \\
		\shortciteA{narayan-etal-2018-document} & \x & & & & & & & \x & & & & \x & \x & & & & \x & & \\
		\shortciteA{narayan-etal-2018-dont} & \x & & & & & & & \x & & & & \x & \x & & & & \x & & \\
		\shortciteA{narayan-etal-2018-ranking} & \x & & & & & & & \x & & & & \x & \x & & & & \x & & \\
		\shortciteA{peyrard-gurevych-2018-objective} & \x & & & & & & & \x & \x & & & & & \x & & \x & & & \\
		\shortciteA{shafieibavani-etal-2018-summarization} & \x & & & & & & & \x & & & \x & \x & & \x & \x & & & & \\
		\shortciteA{song-etal-2018-structure} & \x & & & & & & \x & \x & & & & \x & & \x & & & \x & & \\
		\shortciteA{hardy-etal-2019-highres} & \x & & & & & & & \x & \x & & & \x & & \x & & & \x & & \\
		\shortciteA{makino-etal-2019-global} & \x & & & & & \x & & & & & & & & & & & & & \\
		\shortciteA{zhong-etal-2019-searching} & \x & & & & & \x & & & & & & & & & & & & & \\
		\shortciteA{peyrard-2019-simple} & & & & & & & & \x & \x & \x & & & & \x & \x & & & & \\
		\shortciteA{fabbri-etal-2019-multi} & \x & & & & & & & \x & \x & & & \x & \x & & & & \x & & \\
		\shortciteA{lev-etal-2019-talksumm} & \x & & & & & & &  & & & & & & & & & & & \\
		\shortciteA{you-etal-2019-improving} & \x & & & & & \x & & & & & & & & & & & & & \\
		\shortciteA{isonuma-etal-2019-unsupervised} & \x & & & & & \x & & & & & & & & & & & & & \\
		\shortciteA{wang-etal-2019-biset} & \x & & & & & & & \x & \x & & & \x & & \x & & & \x & & \x\\
		\shortciteA{lebanoff-etal-2019-scoring} & \x & & & & & & & \x & & & & & & & & \x & & & \\
		\shortciteA{li-etal-2019-keep} & \x & & \x & & & \x & & & & & & & & & & & & & \\
		\shortciteA{sharma-etal-2019-bigpatent} & \x & & & & & \x & & & & & & & & & & & & & \\
		\shortciteA{wang-etal-2019-self-supervised} & \x & & & & & \x & & & & & & & & & & & & & \\
		\shortciteA{abacha2019on} & \x & & & & & \x & & & & & & & & & & & & & \\
		\shortciteA{duan-etal-2019-zero} & \x & & & & & \x & & & & & & & & & & & & & \\
		\shortciteA{zhang-etal-2019-hibert} & \x & & & & & & &  & & & & & \x & & & & \x & & \\
		\shortciteA{kouris-etal-2019-abstractive} & \x & & & & & \x & & & & & & & & & & & & & \\
		\shortciteA{zhou-rush-2019-simple} & \x & & & & & \x & & & & & & & & & & & & & \\
		\shortciteA{zheng-lapata-2019-sentence} & \x & & & & & & \x & & & & & & & & & & & & \\
		\shortciteA{frermann-klementiev-2019-inducing} & \x & & & & & & & \x & & & & \x & \x & \x & & & \x & & \\
		\shortciteA{palaskar-etal-2019-multimodal} & \x & & \x & & & & & \x & & \x & \x & \x & & \x & & \x & & & \\
		\shortciteA{sun-nenkova-2019-feasibility} & \x & & & & & & & \x & \x & \x & & & & & & \x & & & \\
		\shortciteA{gui-etal-2019-attention} & \x & \x & & & & & & & & \x & & \x & & \x & & & & \x & \\
		\shortciteA{xiao-carenini-2019-extractive} & \x & \x & & & & \x & & & & & & & & & & & & & \\
		\shortciteA{luo-etal-2019-reading} & \x & & & & & & & \x & \x & & & & \x & & & & & \x & \\
		\shortciteA{duan-etal-2019-contrastive} & \x & & & & & \x & & & & & & & & & & & & & \\
		\shortciteA{zhu-etal-2019-ncls} & \x & & & & & & & \x & \x & & & \x & \x & \x & & & & & \\
		\shortciteA{liu-lapata-2019-text} & \x & & & & & & & \x & \x & & & \x & & & & & \x & & \\
		\shortciteA{gao-etal-2019-write} & \x & & & & & \x & & & & & & & & & & & & & \\
		\shortciteA{west-etal-2019-bottlesum} & \x & & & & & & & & \x & & \x & & \x & &  & & & & \\
		\shortciteA{shen-etal-2019-improving} & \x & & & & & & \x & & & & & \x & & & & & \x & & \\
		\shortciteA{parida-motlicek-2019-abstract} & \x & & \x & & & \x & & & & & & & & & & & & & \\
		\shortciteA{grenander-etal-2019-countering} & \x & & & & & \x & & & & & & & & & & & & & \\
		\shortciteA{li-etal-2019-deep} & \x & & & \x & & & & & & \x & & \x & \x & & & & \x & & \\
		\shortciteA{shapira-etal-2019-crowdsourcing} & & & & & & & & \x & & & & & & \x & \x & \x & & & \\
		\shortciteA{falke-gurevych-2019-fast} & \x & \x & & & & & & \x & \x & & & \x & \x & &  & & & & \\
		\shortciteA{liu-etal-2019-single} & \x & & & & & & & \x & \x & & & \x & \x & & & & \x & & \\
		\shortciteA{ouyang-etal-2019-robust} & \x & & & & & & & \x & & & & \x & & \x & & & \x & & \\
		\shortciteA{kim-etal-2019-abstractive} & \x & & & & & & & & & \x & & & \x & & & & \x & & \\
		\shortciteA{mendes-etal-2019-jointly} & \x & & & & & & & \x & & & & & & & & & & & \\
		\shortciteA{koto-etal-2020-liputan6} & \x & & & \x & & & & \x & \x & & & & & & & & & \x & \\
		\shortciteA{huang-etal-2020-generating} & \x & & & & & \x & & & & & & & & & & & & & \\
		\shortciteA{xiao-carenini-2020-systematically} & \x & & & & & \x & & & & & & & & & & & & & \\
		\shortciteA{lebanoff-etal-2020-cascade} & \x & & & & & \x & & & & & & & & & & & & & \\
		\shortciteA{xu-etal-2020-mixed} & \x & & & & & \x & & & & & & & & & & & & & \\
		\shortciteA{kano-etal-2020-identifying} & \x & & & & & \x & & & & & & & & & & & & & \\
		\shortciteA{gholipour-ghalandari-etal-2020-large} & \x & & & & & \x & & & & & & & & & & & & & \\
		\shortciteA{zhu-etal-2020-attend} & \x & & & & \x & & & \x & \x & & & \x & & \x &  & & & & \\
		\shortciteA{gholipour-ghalandari-ifrim-2020-examining} & \x & & & & & \x & & & & & & & & & & & & & \\
		\shortciteA{xu-etal-2020-self} & \x & & & & & & & \x & & & & \x & \x & & & & \x & & \\
		\shortciteA{gao-etal-2020-supert} & & & & & & & & \x & & & & & & \x & \x & & & & \\
		\shortciteA{sotudeh-gharebagh-etal-2020-attend} & \x & & & & & & \x & \x & & & & \x & & & & \x & & & \\
		\shortciteA{maynez-etal-2020-faithfulness} & \x & & & \x & & & \x & & & & & & & & & & \x & & \\
		\shortciteA{amplayo-lapata-2020-unsupervised} & \x & \x & & & & & \x & \x & & & \x & \x & & & & & & \x & \\
		\shortciteA{mao-etal-2020-facet} & \x & & & & & & & \x & & & & & \x & &  & & & & \\
		\shortciteA{xu-etal-2020-discourse} & \x & & & & & & & & & & \x & \x & & \x &  & & & & \\
		\shortciteA{schumann-etal-2020-discrete} & \x & & & & & & \x & & & & & \x & \x & & & & \x & & \\
		\shortciteA{ladhak-etal-2020-exploring} & \x & \x & & & & & & \x & & & & & & \x & \x & & & & \\
		\shortciteA{durmus-etal-2020-feqa} & \x & & & \x & & & \x & & & & & \x & & \x & & & \x & & \\
		\shortciteA{huang-etal-2020-knowledge} & \x & & & & & & \x & \x & & & & \x & & \x & & & \x & & \\
		\shortciteA{brazinskas-etal-2020-unsupervised} & \x & & & & & & & \x & \x & & \x & \x & & \x & & & \x & & \\
		\shortciteA{suhara-etal-2020-opiniondigest} & \x & & & & & & & \x & \x & & \x & & & \x & & & \x & & \\
		\shortciteA{li-etal-2020-composing} & \x & & & & & & & & \x & & & \x & \x & & & & \x & & \\
		\shortciteA{zhong-etal-2020-extractive} & \x & & & & & \x & & & & & & & & & & & & & \\
		\shortciteA{huang-etal-2020-achieved} & \x & & & & & & \x & & \x & & & \x & & \x & & & \x & & \x\\
		\shortciteA{wang-etal-2020-spectral} & \x & & & & & & & \x & \x & & & \x & & \x &  & & & & \\
		\shortciteA{wang-etal-2020-friendly} & \x & & & & & \x & & & & & & & & & & & & & \\
		\shortciteA{mao-etal-2020-multi} & \x & & & & & \x & & & & & & & & & & & & & \\
		\shortciteA{xiao-etal-2020-modeling} & \x & & & & & \x & & & & & & & & & & & & & \\
		\shortciteA{wu-etal-2020-unsupervised} & \x & \x & & \x & \x & & & \x & \x & \x & \x & \x & & \x & & & \x & & \x\\
		\shortciteA{jia-etal-2020-neural} & \x & & & & & & & \x & & & & \x & \x & & & & & \x & \\
		\shortciteA{xu-lapata-2020-coarse} & \x & & & & & & & & \x & & \x & & & \x &  & & & & \\
		\shortciteA{zou-etal-2020-pre} & \x & & & & & & & \x & \x & & & \x & \x & & & & \x & & \\
		\shortciteA{desai-etal-2020-compressive} & \x & & & & & & \x & & & & & \x & & \x & & & \x & & \x\\
		\shortciteA{cao-etal-2020-factual} & & & & & & & \x & & & & & & & \x & & & \x & & \\
		\shortciteA{tan-etal-2020-summarizing} & \x & & & & & & \x & & & \x & & \x & & \x &  & & & & \\
		\shortciteA{deng-etal-2020-multi} & \x & & & & & & \x & \x & \x & & & \x & & \x &  & & & & \\
		\shortciteA{scialom-etal-2020-mlsum} & \x & \x & & & & \x & & & & & & & & & & & & & \\
		\shortciteA{lu-etal-2020-multi-xscience} & \x & & & & & & &  & & & & & \x & & & & \x & & \\
		\shortciteA{pilault-etal-2020-extractive} & \x & & & & & & \x & \x & & & \x & \x & & \x &  & & & & \\
		\shortciteA{bhandari-etal-2020-evaluating} & & & & & & & & \x & & & & & & \x & \x & & & & \x\\
		\shortciteA{zhao-etal-2020-improving} & \x & & & & & & & \x & & & & \x & & \x & & & & \x & \\
		\shortciteA{cui-etal-2020-enhancing} & \x & & & & & \x & & & & & & & & & & & & & \\
		\shortciteA{lee-etal-2020-reference} & \x & & & & & & \x & & & \x & & \x & & \x & & & & \x &\\ 
		\shortciteA{he-etal-2020-tweetsum} & \x & & & & & \x & & & & & & & & & & & & & \\
		\hline
		\bf Total & \rot{ \medspace \bf 106 \medspace } &  \rot{\bf 11} &  \rot{\bf 4} & \rot{\bf 5} & \rot{\bf 2} &  \rot{\bf 40} &  \rot{\bf 18} &  \rot{\bf 46} &  \rot{\bf 25} &  \rot{\bf 12} &  \rot{\bf 13} &  \rot{\bf 45} &  \rot{\bf 26} &  \rot{\bf 37} &  \rot{\bf 6} &  \rot{\bf 7} &  \rot{\bf 34} &  \rot{\bf 9} &  \rot{\bf 7} \\	
		\hline
	
		\bf Percentage \color{white}\rule{8pt}{1.1cm} &  \mybar{0.95} & \mybar{0.09}  & \mybar{0.04}  & \mybar{0.05}  & \mybar{0.02}  & \mybar{0.36}  & \mybar{0.16}  & \mybar{0.41}  & \mybar{0.22}  & \mybar{0.11} & \mybar{0.12}  & \mybar{0.41}  & \mybar{0.23}  & \mybar{0.33}  & \mybar{0.05}  & \mybar{0.06}  & \mybar{0.31}  & \mybar{0.08}  & \mybar{0.06}  \\
		
	\end{longtable}%
}

To comprehensively understand how recent summarization work has performed evaluation, we followed the lead of \shortciteA{hardy-etal-2019-highres} in conducting a survey of 111 summarization papers from major NLP conferences over the period 2017--2020, and group them into automatic and manual evaluation methods in \tabref{prev_works}.\footnote{The first 26 rows are from \shortciteA{hardy-etal-2019-highres}. We manually re-examine these papers and found miss-annotation for \shortciteA{amplayo-etal-2018-entity}.} Here, we focus on text summarization and exclude multi-modal summarization systems, such as source code \shortcite{ahmad-etal-2020-transformer}, and screen-play summarization \shortcite{papalampidi-etal-2019-movie,papalampidi-etal-2020-screenplay}.

First, as expected, ROUGE is used by more than 95\% of papers, while other metrics such as METEOR, BLEU, \bertscore, and \moverscore are rarely used. Interestingly, 64\% of the surveyed papers used manual evaluation to analyze the strengths and weaknesses of the proposed model(s), an area where the single figure-of-merit output of ROUGE does not provide direct insights.

In \tabref{prev_works}, we summarize the 6 major dimensions of manual 
evaluation as faithfulness, recall, precision, relevance, coherence, and 
fluency. Faithfulness is the degree of factual consistency with respect 
to the source article. Recall, precision, and relevance measure the 
degree of salient and important information, where relevance is 
generally measured as the harmonic mean of precision and recall. 
According to our analysis, recall, precision, faithfulness, and fluency 
are the most frequent dimensions of human evaluation in recent work, 
from which we take inspiration in designing \ffci (with fluency defined 
as inter-sentential coherence, as discussed in \secref{Introduction}).

We also found that absolute scoring is more common than relative evaluation. Absolute benchmark is conducted by asking annotators to evaluate system-generated summaries based on a numeric scale, in isolation of any other summaries. With relative evaluation, on the other hand, annotators are asked to directly rank summaries generated by different methods.

Lastly, the manual evaluations in \tabref{prev_works} were conducted with different basis, namely, SCU (semantic content units, as defined in 
Pyramid \shortcite{nenkova-passonneau-2004-evaluating}), reference, article, and reference+article. SCU is clauses or sentences that are manually extracted from the ground-truth summary, and are used to evaluate content selection in summarization. Pyramid method is initially applied to aggregate the human summaries, however, previous work \shortcite{bhandari-etal-2020-evaluating} applied Pyramid in the single-reference setting.

We observe that most recent work has used the source article as 
 the basis in assessing faithfulness, precision, and recall, rather than 
reference summaries or SCUs.
 Intuitively, this is the best practice in human evaluation, especially 
for faithfulness, as generated summaries can technically contain details 
not found in reference summaries but are in the source article, and they 
should still be seen as faithful information in this case.
 However, for precision and recall, 
\shortciteA{nenkova-passonneau-2004-evaluating,fabbri2020summeval} have 
shown that using the source article rather than reference summaries 
leads to poor inter-annotator agreement as a result of the complication 
in the annotation scheme. Although most papers of the 71 papers in 
\tabref{prev_works} that perform manual evaluation base the evaluation 
on the source article, only 7 out of 71 papers describe explicit quality 
control mechanisms used in their experiments.\footnote{Quality control is a mechanism to measure the quality of the crowd-sourced annotation \shortcite{graham-etal-2016-glitters}.}





\subsection{Resource for Summarization Evaluation}

Best practice in assessing quality automatic metrics for text generation systems is by measuring correlation scores such as Pearson, Spearman, or Kendall between system-generated text and a reference. In machine translation (MT), BLEU \shortcite{papineni-etal-2002-bleu} and METEOR \shortcite{lavie-agarwal-2007-meteor} were validated based on WMT and LDC TIDES 2003 corpora, respectively. While MT metric evaluation resources have been developed progressively over time (e.g.\ the WMT Metrics Task has run annually since 2006), there has been a relative dearth of new evaluation datasets for summarization research, and only recently have  \shortciteA{bhandari-etal-2020-evaluating} and \shortciteA{fabbri2020summeval} released evaluation datasets based on summaries generated by a range of neural summarization models. 

\tabref{resource} comprehensively lists the available resources for summarization evaluation research, in which we observe an 11 year gap between DUC-TAC and the recent datasets. 
Because the summaries in the DUC\footnote{\url{https://duc.nist.gov/data.html}} and TAC\footnote{\url{https://tac.nist.gov/data/}} datasets are from more than 10 years ago, they are based on largely outdated extractive summarization systems. \shortciteA{bhandari-etal-2020-evaluating,fabbri2020summeval,maynez-etal-2020-faithfulness,wang2020asking} attempt to tackle this issue by releasing new data, although the dimensions of evaluation represented in those datasets do not fully align with the common dimensions of manual evaluation in \tabref{prev_works}. For instance, \shortciteA{bhandari-etal-2020-evaluating} only assess coverage based on SCUs, and \shortciteA{fabbri2020summeval} do not separate out precision and coverage. 

\begin{table*}[t]
	\begin{center}
		\begin{adjustbox}{max width=1\linewidth}
			\begin{tabular}{lclcp{4cm}p{2cm}}
				\toprule
				\multicolumn{2}{l}{\multirow{2}{*}{\bf Dataset}} & \multirow{2}{*}{\bf \#Systems} & \bf \#Summaries & \multirow{2}{*}{\bf Aspects} & \bf Annotation \\
				& & & \bf per System & & \bf benchmark \\
				\midrule
				DUC--2001 & \multirow{3}{*}{\rot{SDS}} & 12 & 149 & \multirow{3}{*}{Coverage} & \multirow{3}{*}{1 reference} \\
				DUC--2002 & & 14 & 295 & & \\
				DUC--2003 & & 14 & 624 & & \\
				\midrule
				DUC--2001 & \multirow{3}{*}{\rot{MDS}} & 42 & 29 & \multirow{3}{*}{Coverage} & \multirow{3}{*}{1 reference} \\
				DUC--2002 & & 36 & 59 &  &  \\
				DUC--2003 & & 18 & 30 &  &  \\
				\midrule
				TAC--2008 & \multirow{2}{*}{\rot{SDS}} & 54 + 4 ref & 480 & Pyramid (Coverage) & 1 reference \\
				TAC--2009 & & 55 + 4 ref & 440 &  Responsiveness & \\
				\midrule
				\shortciteA{bhandari-etal-2020-evaluating} (CNNDM) & \multirow{2}{*}{\rot{SDS}} & 25 & 100 & Pyramid (Coverage) & 1 reference \\
				\cmidrule{5-6}
				\shortciteA{fabbri2020summeval} (CNNDM) & & 23 & 100 & Faithfulness, Relevance, Coherence, Fluency & article \\
				
				\midrule
				
				\shortciteA{maynez-etal-2020-faithfulness} (XSUM) & \multirow{3}{*}{\rot{SDS}} & 4 & 500 & \multirow{3}{*}{Faithfulness} & \multirow{3}{*}{article} \\
				\shortciteA{durmus-etal-2020-feqa} (CNNDM, XSUM) & & N/A & 1034* & & \\
				\shortciteA{wang2020asking} (CNNDM, XSUM) & & 1 & 474 & & \\
				\bottomrule
			\end{tabular}
		\end{adjustbox}
	\end{center}
	\caption{Resources for summarization evaluation. MDS/SDS = Multi/Single Document Summarization. * indicates the total of summaries for all system as \#Systems is not reported by \shortciteA{durmus-etal-2020-feqa}. }
	\label{tab:resource}	
\end{table*}

\shortciteA{bhandari-etal-2020-evaluating} annotated 100 samples based
on the simplified Pyramid method
\shortcite{nenkova-passonneau-2004-evaluating}, where semantic content
units (SCUs) are manually extracted and crowd-workers then count the
appearance of SCUs in the summary. This annotation scheme is closely
related to coverage as proposed in this research, but does not consider
focus, faithfulness, and inter-sentential
coherence. \shortciteA{bhandari-etal-2020-evaluating} and
\shortciteA{peyrard-2019-studying} both found that evaluation metrics
developed based on older datasets do not necessarily perform well on modern
datasets with more modern summarization systems.

\shortciteA{fabbri2020summeval} assess four dimensions of summaries:
relevance, consistency, fluency, and coherence, by annotating 100
CNN/DailyMail samples. Our \ffci framework further decomposes relevance
into focus and coverage to provide a more fine-grained understanding of
content overlap, and replaces fluency --- which measures quality of
individual sentences --- with inter-sentential coherence, which measures
the quality of multi-sentence summaries more holistically. They evaluated summaries via
crowd-sourcing (Amazon MTurk) and expert (in-house) annotators, but
ultimately base all of their findings on the expert annotations, as they
found the crowd-sourced annotations to be highly inconsistent.
First, their annotation scheme is difficult for crowd-workers, as they
are asked to judge all four dimensions after reading an article and a
system-generated summary. Consistency (faithfulness) is found to be
particularly difficult (and subjective), and previous studies
\shortcite{maynez-etal-2020-faithfulness} have attempted to ease the
annotation burden by asking crowd-workers to highlight unfaithful spans
in the summary. Assessing relevance without a ground-truth summary is
also hard, as it requires crowd-workers to implicitly construct their
own summary of the article.  The second reason is that there is no
quality control to verify the quality of the annotations, which means
they may be potentially unreliable.  In this work, we use the
resource released by \shortciteA{maynez-etal-2020-faithfulness} to
study faithfulness, and use the customized \textit{Direct Assessment} framework
\shortcite{graham-etal-2015-accurate} to collect judgements across
three additional dimensions: focus, coverage, and inter-sentential coherence.
The annotation framework we use has the following benefits: a more
intuitive annotation scheme, better quality control, and better handling
of annotator variance (through \textit{z}-score normalization).

Perhaps more importantly, despite presenting extensive evaluation on a
number of state-of-the-art summarization systems using a wide range of
evaluation methods, \shortciteA{fabbri2020summeval} stop short of
providing guidance as to the best evaluation methods for assessing a
particular dimension of summary quality.  Our work addresses this gap by
providing practical advice on the best evaluation method for assessing
the four dimensions of \ffci.


\section{Existing Evaluation Metrics and Extensions}
\label{sec:metric}

In this section, we review the existing evaluation metrics that we use
for different dimensions of summarization evaluation (faithfulness,
focus, coverage, and inter-sentential coherence). We first introduce
traditional string overlap-based evaluation metrics, with a particular
focus on ROUGE \shortcite{lin-2004-rouge}, as it has become the de facto
standard for automatic summarization evaluation. We apply the
overlap-based metrics in evaluating all four dimensions of \ffci (see \tabref{prev_works}). Next,
we present the recently-proposed QAGS question answering-based framework
for evaluating faithfulness \shortcite{wang2020asking}, which we extend
to also evaluate focus and coverage (but not inter-sentential
coherence). We then introduce two general-purpose string similarity
metrics, namely the unsupervised \bertscore
\shortcite{zhang2020bertscore} and supervised \stsscore, which is
trained over STS data from successive SemEval tasks
\shortcite{agirre-etal-2012-semeval}. Both of these metrics are used to
evaluate all four dimensions of \ffci. Finally, we introduce the
coherence score of \shortciteA{nayeem-chali-2017-extract} and
\shortciteA{yin2020enhancing} as a specialized metric for evaluating
inter-sentential coherence.

\subsection{Traditional String Overlap-based Evaluation Metrics}

Despite its brittleness, the simplicity of ROUGE 
\shortcite{lin-2004-rouge} has made it a mainstay of summarization 
evaluation for over 15 years. ROUGE measures the overlap between a 
generated and reference summary in terms of unigram or bigram overlap 
(ROUGE-1 and ROUGE-2, respectively), or longest common subsequence 
(ROUGE-L). In another work, \shortciteA{ng-abrecht-2015-better} proposed 
ROUGE-WE as an extension of ROUGE which incorporates word2vec 
\shortcite{mikolov-etal-2013-linguistic} embeddings, but found it to 
perform similarly to the simpler ROUGE-1 and ROUGE-2 metrics in practice.

In most studies, the harmonic mean (F1) of each of the three main ROUGE 
variants (ROUGE-1, ROUGE-2 and ROUGE-L) is used for 
evaluation, which some studies 
\shortcite{see-etal-2017-get,pasunuru-bansal-2018-multi} complement with 
machine translation metrics such as BLEU 
\shortcite{papineni-etal-2002-bleu} and METEOR 
\shortcite{lavie-agarwal-2007-meteor}. 


\subsection{Question Answering-based Evaluation}
Recent work \shortcite{wang2020asking,durmus-etal-2020-feqa} has shown that factual consistency can be evaluated using a question answering (``QA'') task formulation. In this paper, we experiment with the QAGS framework \shortcite{wang2020asking}, which involves two components: (1) question generation (``QG''), and (2) question answering.

Let $X$, $Y$, and $Y'$ be the source document, reference summary, and system summary, respectively. For faithfulness, QAGS defines $p(Q|Y')$ as the distribution over questions $Q$ generated from system summary $Y'$. Answer $A$ is predicted based on two terms: $p(A|Q,X)$ and $p(A|Q,Y')$, representing the answer distribution based on the source document and system summary, respectively. Factual consistency is measured by the F1 score (or exact match) between the answers generated from the source document and system summary.

Evaluating faithfulness via QA-based evaluation such as QAGS is intuitive but has several drawbacks. First, QAGS requires careful tuning of hyperparameters such as the number of questions to generate, maximum token length for question generation, and also question filtering method. Secondly, QA-based evaluation is computationally expensive and hard to apply to languages other than English, due to the need for training data for the QA and QG models. 

Despite its drawbacks, previous work has reported encouraging results in the evaluation of faithfulness. In this work, we extend the QA-based method to two other elements of FFCI: focus and coverage. In this, we address the following research question: \RQ[1]{how effective is QAGS relative to other simpler methods for assessing faithfulness, and can it be applied to evaluate focus and coverage?}

\subsection{\bertscore}
Contextualized word embeddings have been shown to be a strong metric for
evaluating machine translation
\shortcite{mathur-etal-2019-putting,zhang2020bertscore}.
\shortciteA{zhang2020bertscore} proposed \bertscore as a means of
computing the similarity between \bert token embeddings of system and
reference texts, while in other work
\shortciteA{zhao-etal-2019-moverscore} proposed \moverscore as the
Euclidean distance between two contextualized \bert representations. We
use \bertscore rather than \moverscore in this study for two reasons: (1) \moverscore is
symmetric (i.e.\ MoverScore$(x, y) =$ MoverScore$(y, x)$), and as such
cannot easily be used to evaluate precision and recall separately; and (2) recent work
\cite{fabbri2020summeval} has shown that \bertscore is superior to \moverscore for summarization evaluation.

For $Y$ and $Y'$ as the reference and system summary, in the context of summarization, \bertscore is computed as follows: 
\begin{align*}
\precision_{\text{BERT}} &= \frac{1}{|Y'|} \sum_{t_i \in Y'} \max_{s_j \in Y} t_i^T s_j\\
\recall_{\text{BERT}} &= \frac{1}{|Y|} \sum_{s_j \in Y} \max_{t_i \in Y'} t_i^T s_j\\
\fscore_{\text{BERT}} &= 2\frac{\precision_{\text{BERT}}\cdot \recall_{\text{BERT}}} {\precision_{\text{BERT}}+\recall_{\text{BERT}}}
\end{align*}
where $s_j$ and $t_i$ are token embeddings of $Y$ and $Y'$.

In terms of hyperparameters, \bertscore is simpler than QA-based 
evaluation, with the main hyperparameter being layer selection: 
\shortciteA{zhang2020bertscore} found that selection of which transformer 
layer to source the token embeddings from is critical to performance. For machine 
translation and text generation evaluation, \shortciteA{zhang2020bertscore} 
recommend the use of $\fscore_{\text{BERT}}$ based on the 24th layer of 
\texttt{roberta-large}, on the basis of experiments over \bert 
\shortcite{devlin-etal-2019-bert}, \roberta \shortcite{liu2019roberta}, and \xlnet 
\shortcite{yang2019xlnet}.

Since layer selection in the original paper was based on machine translation datasets, we perform similar layer selection across the three sub-facets of \ffci, asking: \RQ[2]{which layer of which pre-trained language model is best for evaluating faithfulness, focus, and coverage of a summary?}

We perform a model--layer search to answer this question, extending the work of \shortciteA{zhang2020bertscore} to include other pre-trained models. In total, we examine 7 
model types, that can be categorized as follows: (1) encoder-only = \bert \shortcite{devlin-etal-2019-bert}, \roberta 
\shortcite{liu2019roberta}, and \xlnet \shortcite{yang2019xlnet}; (2) decoder-only = \gpt \shortcite{radford2019language}; and (3) encoder--decoder = T5 \shortcite{raffel2019exploring}, \bart 
\shortcite{lewis2019bart}, and \pegasus \shortcite{zhang2020pegasus}. For each of these, we experiment with different-sized pre-trained models, for a total of 19 models. For encoder--decoder models, we only perform layer selection over the encoder layers. 

\subsection{Model-based approaches}

\subsubsection{FactCC}
\shortciteA{kryscinski-etal-2020-evaluating} proposed a weakly-supervised, model-based approach for verifying factuality in abstractive summaries. The training data is generated based on transformation rules including paraphrasing, entity and number swapping, pronoun swapping, sentence negation, and noise injection. The goal is to estimate $P(y|A,c)$ where $y$ is a binary label of \texttt{CORRECT} and \texttt{INCORRECT}, $A$ is the source article, and $c$ is the transformed sentence (claim or summary). For training, \shortciteA{kryscinski-etal-2020-evaluating} simply fine-tuned \bert model and use [CLS] for classification (and denote this model as FactCC). Additionally, the training is extended by allowing the model to not only classify the claim consistency but also highlight a span in the source article as the supporting evidence (and denote this model as FactCCX).

\subsubsection{\stsscore}

We additionally experiment with \stsscore. Semantic textual similarity (STS) measures the relative semantic similarity of two short texts (often single sentences) on a continuous scale of $[0,5]$ \shortcite{agirre-etal-2012-semeval}. A broad range of STS approaches have been proposed, and datasets have been released for a number of different languages, predominantly through successive SemEval tasks.

Similar to \bertscore, STS-score is a similarity function, from which precision, recall, and F1 can be calculated as follows:
\begin{align*}
\precision_{\text{STS}} &= \frac{1}{|Y'|} \sum_{t_i \in Y'} \max_{s_j \in Y} \text{STS}(t_i,s_j)\\
\recall_{\text{STS}} &= \frac{1}{|Y|} \sum_{s_j \in Y} \max_{t_i \in Y'} \text{STS}(s_j,t_i) \\
\fscore_{\text{STS}} &= 2\frac{\precision_{\text{STS}}\cdot \recall_{\text{STS}}} {\precision_{\text{STS}}+\recall_{\text{STS}}}
\end{align*}
where $s_j$ and $t_i$ are segments within reference summary $Y$ and
system summary $Y'$, respectively. We experiment with three segment
granularities: (1) elementary discourse units (EDUs)
\shortcite{mann-1984-discourse};\footnote{Typically a clause or sentence
  that represents atomic information in discourse parsing. Reference and
  generated summaries may have different sentence lengths/granularities,
  and EDU-based segmentation is a possible alternative to sentence-based
  matching} (2) sentence; and (3) document. As our STS scorer, we use a
fine-tuned sentence transformer, based on the findings of
\shortciteA{reimers-gurevych-2019-sentence}.


\begin{table*}[t]

	\begin{center}
		\begin{adjustbox}{max width=1\linewidth}
			\begin{tabular}{p{5cm}L{4cm}L{4cm}L{4cm}}
				\toprule
				\textbf{Context} & \textbf{\shortciteA{maynez-etal-2020-faithfulness}} & \textbf{\shortciteA{durmus-etal-2020-feqa}} & \textbf{\shortciteA{wang2020asking}}\\
				\midrule
				XSUM samples & 2000 (4 models) & 286 (unknown) & 239 (1 model) \\
				CNN/DailyMail samples & --- & 748 (4 models) & 235 (1 model) \\
				Filtered data sampling & No & Yes (meaningful sentence) & No \\
				Quality control in annotation & Yes (pilot study) & None & Yes \\
				Evaluation against reference & ROUGE, \bertscore & ROUGE & ROUGE, \bertscore, BLEU, METEOR \\
				Evaluation against article & QA, Entailment & ROUGE, BLEU, \bertscore, Entailment, QA & QA \\
				Best evaluation & Entailment & QA & QA \\
				\bottomrule
			\end{tabular}
		\end{adjustbox}
	\end{center}
	\caption{Summary of previous work on faithfulness evaluation.}
	\label{tab:faithfulness}
	
\end{table*}

\subsection{Coherence Score}

\shortciteA{nayeem-chali-2017-extract} and \shortciteA{yin2020enhancing}
define coherence score as the weighted sum of similarity scores of two
adjacent sentences. For system summary $Y'$, the coherence score is computed as follows:
\begin{align*}
\text{coherence}(Y') &= \frac{1}{n-1} \sum_{i=1}^{n-1} \text{Sim}(t_i,t_{i+1})\\
\text{Sim}(t_i,t_{i+1}) &= \lambda \text{NESim}(t_i,t_{i+1}) + (1-\lambda) \text{CosSim}(t_i,t_{i+1})
\end{align*}
where NESim is the named entity overlap of two sentences $t_i$ and
$t_{i+1}$ in $Y'$, and cosine similarity is measured based on
pre-trained word embeddings.\footnote{We use GloVE embeddings
  \shortcite{pennington-etal-2014-glove} to compute coherence scores,
  consistent with \shortciteA{nayeem-chali-2017-extract}.} While this
method is commonly used to assess coherence in the sentence ordering
task \cite{shen2021simple}, this is the first paper to systematically evaluate its
effectiveness for summarization evaluation.

\section{FFCI Framework}

\subsection{Faithfulness}
\label{sec:faith}

Abstractive summarization is prone to ``hallucination'' or factual inconsistencies, where information is generated that does not exist in the source document \shortcite{maynez-etal-2020-faithfulness,wang2020asking}.
Three recent papers independently proposed to evaluate the degree of 
hallucination 
\shortcite{maynez-etal-2020-faithfulness,durmus-etal-2020-feqa,wang2020asking}, 
as detailed in \tabref{faithfulness}.

In terms of training data, \shortciteA{maynez-etal-2020-faithfulness} released
the largest dataset with 2,000 annotated summaries generated over
XSUM. \shortciteA{durmus-etal-2020-feqa} manually pre-filtered the data to
select ``meaningful'' sentences, making it difficult to fully automate
the method, and do not report on any quality controls in their human
annotation. On the other hand, \shortciteA{maynez-etal-2020-faithfulness} conducted a pilot
study to train their annotators, and \shortciteA{wang2020asking} applied
annotator attention checks, making us more confident in the quality of
the resultant dataset.

We argue that the best way to evaluate faithfulness is by comparing 
the generated summary with the source document (and not with the 
reference summary).\footnote{As we argued earlier (\secref{rel_aspect}), 
details in the generated summary that are not in the reference summary 
but in the source article should still be regarded as faithful 
information.}
  In \tabref{faithfulness}, \shortciteA{durmus-etal-2020-feqa} is the 
only paper to extensively measure traditional and model-based metrics 
against the source article.  However, because of concerns over their 
data, we revisit faithfulness evaluation using the dataset of 
\shortciteA{maynez-etal-2020-faithfulness}.\footnote{At the time this 
research was conducted, only \shortciteA{maynez-etal-2020-faithfulness} 
had released their data.} We score faithfulness by comparing summary 
sentences and the source document as follows:
\begin{subequations}
	\begin{align*} 
	\text{FA}_{\METRIC} &= \frac{1}{|Y'|} \sum_{t_i \in Y'} \text{A}(t_i,X, n) \\
	\text{A}(t_i,X, n) &= \underset{s_j\in X}\AvgTopN \;\METRIC(t_i,s_j) 
	\end{align*}
\end{subequations}
where $t_i$ and $s_j$ are sentences from the system summary $Y'$ and
source document $X$, respectively; $\METRIC$ $\in$ \{ROUGE, \stsscore,
\bertscore\}; and $n\in \mathbb{Z}^+$ is a hyperparameter.  \AvgTopN
matches sentence $t_i$ from the summary with each sentence $s_j$ in the
source document $X$, and returns the average score for the top-$n$
best-matching sentences. The intuition behind measuring across the
top-\textit{n} is that information in a summary sentence might
potentially be drawn from different sentences in the source article.

For faithfulness, ROUGE, \stsscore, and \bertscore are based on F1-scores.
In preliminary experiments, we compared $n\in\{1,2,3\}$ and found that $n=2$ works best for ROUGE, and $n=3$ works best for \stsscore and the pre-trained language model scores.

\subsection{Focus and Coverage}

\begin{table}[t]
	\begin{center}
		\begin{adjustbox}{max width=0.8\linewidth}
			\begin{tabular}{L{3.5cm}L{3cm}L{6cm}}
				\toprule
				\textbf{Metric} & \textbf{Question} & \textbf{Answer} \\
				\midrule
				Faithfulness & $p(Q|Y')$ & $p(A|Q,X)$ and $p(A|Q,Y')$\\
				Focus & $p(Q|Y')$ & $p(A|Q,Y)$ and $p(A|Q,Y')$\\
				Coverage & $p(Q|Y)$ & $p(A|Q,Y)$ and $p(A|Q,Y')$\\
				\bottomrule
			\end{tabular}
		\end{adjustbox}
		\caption{Probability distributions used in QAGS to evaluate faithfulness, focus, and coverage.}
		\label{tab:fact_cov}
	\end{center}
\end{table}

We test the ability of the evaluation metrics from \secref{metric} to
measure focus and coverage, arguing that it is important to separately
measure summaries in terms of precision and recall relative to a reference.

First, we adopt QAGS from faithfulness evaluation \shortcite{wang2020asking}, and extend it to evaluate focus and coverage based on the probability distributions in \tabref{fact_cov}. For focus, we generate questions $Q$ based on system summary ($p(Q|Y')$) similarly to faithfulness, and answer the questions based on $p(A|Q,Y)$ and $p(A|Q,Y')$. That is, we test the consistency of answers generated from the system and reference summaries, based on questions generated from the system summary (meaning we only evaluate information present in the \textit{system} summary as this is the source of the questions; hence focus). For coverage, on the other hand, we generate questions based on the \textit{reference} summary ($p(Q|Y)$), and answer those questions based on $p(A|Q,Y)$ and $p(A|Q,Y')$, in the same manner as focus (meaning we evaluate information present in the \textit{reference} summary; hence coverage). We return to discuss how to generate questions and answers in \secref{model-exp}.

Apart from QAGS, we also examine ROUGE, METEOR, BLEU, \bertscore, and \stsscore to
evaluate focus and coverage. For computing ROUGE, \stsscore, and \bertscore, we use the precision and recall for focus and coverage, respectively.

\subsection{Inter-Sentential Coherence}

We extend the \shortciteA{nayeem-chali-2017-extract} method to measure
inter-sentential coherence (IC) within system summary $Y'$, based on a
next-sentence-prediction (NSP) classifier as follows:
\begin{align*}
{\text{NSP}(Y')} &= \text{mean}_{t_i \in Y'} \NSP(t_i,t_{i+1})
\end{align*}
where each $t_i$ is a sentence in summary $Y'$, and $\NSP$ returns a 
probability of $t_{i+1}$ following $t_i$. We experimented with $\max$, 
$\min$, and mean aggregation, but found mean to produce the most robust 
results. 

Compared to \shortciteA{nayeem-chali-2017-extract}, our NSP score also assesses coherence between two adjacent sentences, but with a model-based system, rather than based on cosine similarity of pre-trained word embeddings.
This way, we argue that the NSP score can better assess the overall writing flow based on two adjacent sentences because it is not limited by the factual content of the sentences.\footnote{Having said which, although the facts in both sentences can be different, they should still have similar topics and flow coherently.}

Note that we do not use the NSP classifier in pretrained language 
models, as not all pretrained language models have this objective.  
Instead, we train a seperate NSP classifier (by fine-tuning pretrained 
language models) where positive examples are two consecutive sentences 
and negative samples are constructed using a range of strategies (e.g.\ 
by flipping the sentences), as detailed in \secref{model-exp}.

In contrast to ROUGE, \stsscore, and 
\bertscore, our proposed evaluation scheme for inter-sentential coherence is a reference-less metric. ROUGE, \stsscore, and \bertscore are reference-based metrics and designed for evaluating saliency and coverage when compared to a reference text. That said, reference-based metrics  might implicitly assess 
inter-sentential coherence because a system summary that is similar to the gold summary is likely to be coherent (since the human-written gold summary should be coherent).


\section{Experimental Setup}

\subsection{Data}
\label{sec:data_human}

In order to evaluate the different metrics and perform model--layer selection, we need gold-standard data for each of the four \ffci sub-tasks. 

\subsubsection*{Faithfulness}

We use the 2000 samples from \shortciteA{maynez-etal-2020-faithfulness}, which
is based on summaries generated by 4 neural models over XSUM
\shortcite{narayan-etal-2018-dont}: pointer generator network (``PG'':
\shortciteA{see-etal-2017-get}), Topic-aware convolutional Seq2Seq
(``TCONV'': \shortciteA{narayan-etal-2018-dont}), a transformer-based
model (``TRANS2S'': \shortciteA{vaswani2017attention}), and \bert
(``\bert'':
\shortciteA{devlin-etal-2019-bert,liu-lapata-2019-text}).\footnote{Note
  that, at the time of writing, there is no annotated dataset for
  faithfulness based on CNN/DailyMail, so we can only evaluate faithfulness
  over XSUM.}

\subsubsection*{Focus and coverage}

We annotate 1080 data--model pairs by randomly sampling 135 articles each from the test sets of CNN/DailyMail \shortcite{hermann2015teaching} and XSUM \shortcite{narayan-etal-2018-dont}, and generate summaries with two models: PG \shortcite{see-etal-2017-get} and \bert \shortcite{liu-lapata-2019-text}; this results in 540 summaries ($135 \times 2 \times 2$) which are assessed for focus and coverage. 

Note that for CNN/DailyMail, we use the PG+Coverage variant, while for XSUM, the basic PG model is used, as it produces better summaries \shortcite{see-etal-2017-get,narayan-etal-2018-dont}. We choose these data--model pairs for two reasons: (1) CNN/DailyMail and XSUM are benchmark abstractive summarization datasets which represent the most extractive and abstractive summarization corpora, respectively \shortcite{bommasani-cardie-2020-intrinsic}\footnote{Noting that contemporaneous works \shortcite{bhandari-etal-2020-evaluating,fabbri2020summeval} only use CNN/DailyMail.}; and (2) PG and \bert are representative of contemporary neural models from the attention-based recurrent model to pre-trained language model era.

\subsubsection*{Inter-sentential coherence}

We used the same 270 system summaries from CNN/DailyMail as for focus
and coverage.\footnote{Noting that XSUM summaries are single sentences,
  and thus inter-sentential coherence is not relevant.}

\subsection{Human Evaluation Task Design}
\label{sec:design}

\begin{figure}[t]
	\centering
	\subfloat[Focus and coverage annotation.]{\includegraphics[width = 4.75in]{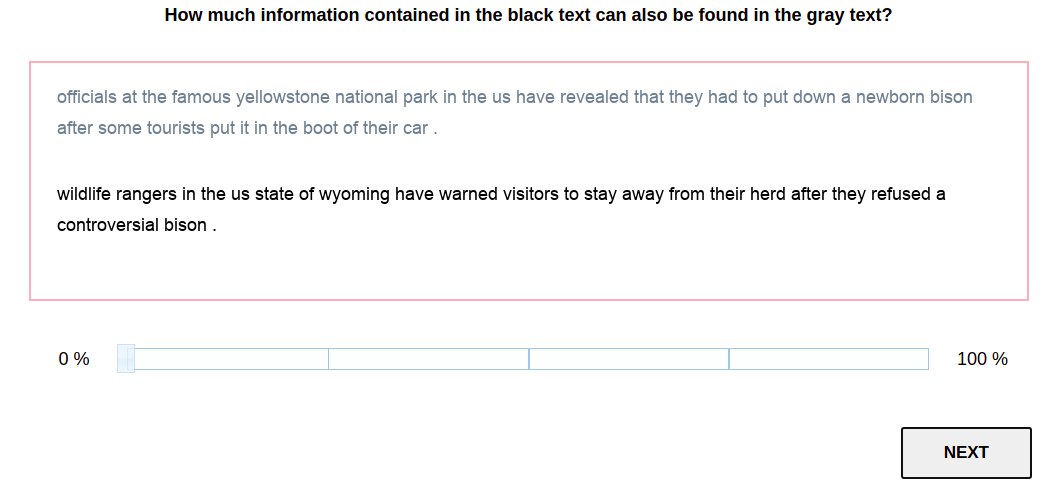}} \\
	\subfloat[Inter-sentential coherence annotation.]{\includegraphics[width = 4.75in]{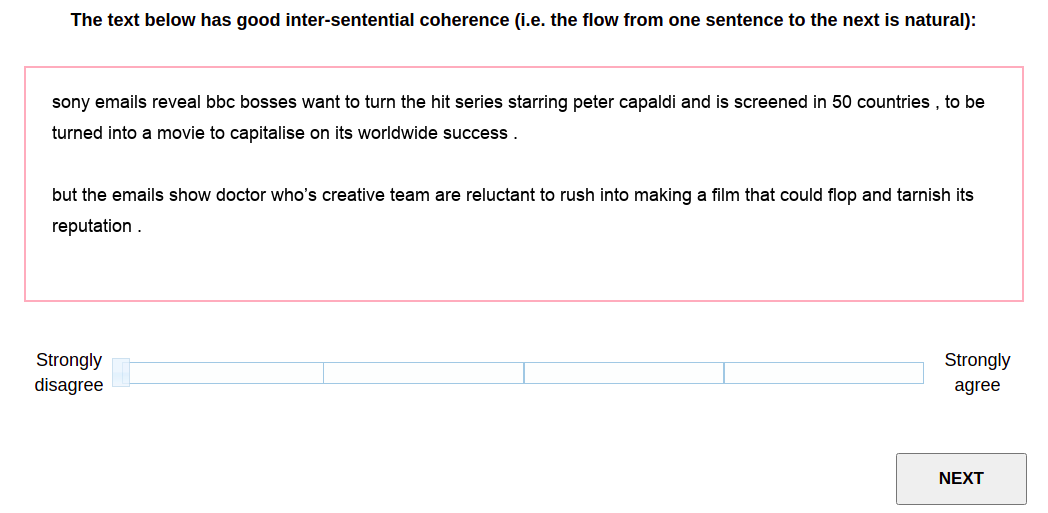}}\\
	\caption{MTurk annotation interface for focus, coverage, and inter-sentential coherence.}
	\label{fig:mturk}
	\vspace{1cm}
\end{figure}

We used Amazon Mechanical Turk\footnote{\url{https://www.mturk.com/}} and the customized Direct Assessment (``DA'') method \shortcite{graham-etal-2015-accurate,graham2017can}, which has become the de facto for MT evaluation in WMT. DA equips the annotation scheme with some pre-annotated samples for quality control, two texts (system and human translation), and a slider button (continuous scale with range 1--100) for annotation. In \figref{mturk}, we present the annotation interface of the customized DA method for summarization evaluation.

For focus and coverage, the annotation interface provides system and reference summary, and a question: \textit{How much information contained in the second text can also be found in the first text?} We were able to combine focus and coverage annotation, as the only thing
that differentiates them is the ordering of the system and reference summaries, which was invisible to annotators.\footnote{For focus, the first and second texts are the reference and system summaries, respectively. For coverage, the order is reversed, and they are the system and reference summaries, respectively.} For inter-sentential coherence, the annotators are given a single summary and asked to rate inter-sentential coherence directly.

While it may seem more natural and reliable to evaluate focus and coverage based on the source
document than the ground-truth summary, we use the ground-truth summary in this research for the following reasons. First, historically, validation of automatic summarization evaluation metrics has been based primarily on ground-truth summaries (not source documents). Second, previous work such as DUC (dataset for ROUGE), TAC (dataset for \moverscore), and \shortciteA{bhandari-etal-2020-evaluating} annotated coverage based on a single reference summary. Third, this work is based on single-document summarization systems, and we argue that the variance in content is actually not that
great.  Lastly, basing human evaluation (of focus and coverage) on the source article leads to more complicated annotation schemes, and has been shown to yield poor annotations (as discussed in \secref{rel_aspect}).

We posted separate HITs for focus + coverage vs.\ inter-sentential coherence, where a single HIT consisted of 100 annotation instances including 10 quality control instances. For focus + coverage, 5 samples are random pairs (should be scored 0) and the remaining 5 samples are repetitions with minor edits (should be scored 100). For inter-sentential coherence, 5 samples are random sentence pairs, and the remaining 5 are verbatim repeats (both of which should be scored 0).

We restricted the HITs to US-based workers with at least 10,000 approved
HITs. For each HIT, we pay USD\$5 and an additional \$8 bonus if they
pass the quality control checks (the minimum bar is to have at least 7
correct answers from the total of 10 quality control samples), to ensure
that workers are paid at a level that is comfortably above the minimum
wage in Australia.\footnote{For inter-sentential coherence annotation,
  we pay USD\$3 and an additional \$3 bonus.}

\begin{table}[t]
	\begin{center}
		\begin{adjustbox}{max width=\linewidth}
			\begin{tabular}{lccc}
				\toprule
				\multirow{2}{*}{\bf Aspect} & \bf Pearson correlation ($r$) & \bf Avg. Quality & \bf Avg. Working \\
				& \bf(agreement) & \bf score (\%) & \bf time (min) \\ 
				\midrule
				Focus + Coverage & 0.57 & 90.3 & 62.1 \\
				Inter-sentential coherence & 0.49 & 94.4 & 35.8 \\
				\bottomrule
			\end{tabular}
		\end{adjustbox}
		\caption{Statistics over the approved HITs on both criteria: Focus + Coverage and Inter-sentential coherence. Quality score is the average score of 10 quality control samples for each HIT.}
		\label{tab:annotation_stat}
	\end{center}
\end{table}

We collected 3 annotations per HIT which passed quality control (running new HITs in cases where HITs had to be discarded), and present the statistics over HITs in \tabref{annotation_stat}.
We achieved a mean Pearson's correlation between annotators of $r$ = 0.57 and 0.49, for focus + coverage and inter-sentential coherence, respectively. We also observe that the average quality score is high and the working time of both HITs is reasonable.
 To aggregate the scores, we standardized the scores of each worker into 
a $z$-score before averaging.

\subsection{Evaluation Metrics}
\label{sec:model-exp}

\subsubsection*{ROUGE, METEOR, and BLEU}
In this experiment, we use the original implementation of 
ROUGE\footnote{\url{https://github.com/bheinzerling/pyrouge}} and 
METEOR\footnote{text\url{http://www.cs.cmu.edu/∼alavie/METEOR/}}. For 
BLEU, it is based on SacreBLEU implementation 
\shortcite{post-2018-call}.\footnote{\url{https://github.com/mjpost/sacrebleu}}


\subsubsection*{QAGS}

We re-implemented QAGS \shortcite{wang2020asking} by training the question generator with \texttt{bart-large} on NewsQA \shortcite{trischler2017newsqa}, and QA model with \texttt{bert-large-wwm} on SQuAD2.0 \shortcite{jia2018know}, achieving similar results to \shortciteA{wang2020asking} on both tasks. 
We generate a maximum of 50 questions, and discard questions if the QA system cannot predict the correct answer based on the original context the question is generated from.

To validate our implementation, we tested the model at faithfulness
evaluation over XSUM using the dataset of
\shortciteA{maynez-etal-2020-faithfulness}, and achieved a correlation of
$r = 0.25$, 0.075 points higher than the original paper.\footnote{Noting
  that at the time of writing, the QAGS data and code have not been
  released,
  and that our evaluation is thus based on a different test dataset to
  the authors.}

\subsubsection*{\bertscore}

We sourced 19 pre-trained language models from
HuggingFace,\footnote{\url{https://huggingface.co/}}
and adjust the
\bertscore implementation to search for the best model--layer
combination.\footnote{\url{https://github.com/Tiiiger/bert_score}} The models include \bert \shortcite{devlin-etal-2019-bert}, \roberta \shortcite{liu2019roberta}, \xlnet \shortcite{yang2019xlnet}, \gpt \shortcite{radford2019language}, T5 \shortcite{raffel2019exploring}, \bart \shortcite{lewis2019bart}, and \pegasus \shortcite{zhang2020pegasus}.\footnote{\texttt{bert-base-uncased}, \texttt{bert-large-uncased}, \texttt{roberta-base}, \texttt{roberta-large}, \texttt{roberta-large-mnli}, \texttt{xlnet-base-cased}, \texttt{xlnet-large-cased}, \texttt{gpt2}, \texttt{gpt2-medium}, \texttt{gpt2-large}, \texttt{gpt2-xl}, \texttt{t5-small}, \texttt{t5-base}, \texttt{t5-large},  \texttt{bart-base}, \texttt{bart-large}, \texttt{pegasus-xsum}, \texttt{pegasus-cnn\_dailymail}, \texttt{pegasus-large}.} 

In selecting the best layers of each model for faithfulness, focus, and coverage, we do not merge datasets and systems, but instead select based on the averaged best results across different dataset--system combinations. We believe this is a robust method which does away with the need for a method such as cross-validation. Another reason not to merge datasets and systems (and compute correlation across summary-level data points) is because of the small number of systems: unlike DUC and TAC which have many systems, the faithfulness data \shortcite{maynez-etal-2020-faithfulness} only consists of 4 different systems from 1 dataset, while our focus and coverage data consist of 2 systems from 2 datasets.

\subsubsection*{FactCC}
We use the models and original implementation of FactCC and FactCCX from \shortciteA{kryscinski-etal-2020-evaluating} to evaluate faithfulness.\footnote{\url{https://github.com/salesforce/factCC}} We use the probability of class \texttt{CORRECT} (a summary being factually correct relative to the source article) as the final output for both FactCC and FactCCX.

\subsubsection*{\stsscore}

We use sentence-transformers \shortcite{reimers-2019-sentence-bert} with
\texttt{bert-large-nli}, using
\texttt{spaCy}\footnote{\url{https://spacy.io/}} and discourse
segmentation \shortcite{ji-eisenstein-2014-representation} to perform
sentence and EDU segmentation. We also experimented with other
pre-trained transformer models, but found there to be little difference
in the results.

\subsubsection*{NSP score}

For inter-sentential coherence, we fine-tune a pretrained language model
for NSP classification
on 100,000 sentence pairs (50K positive and 50K negative) automatically 
extracted from original XSUM articles. We experimented with four types 
of negative samples: \texttt{type1} = flipped sentence pairs; 
\texttt{type2} = pairs where the second sentence is randomly obtained 
from a different document; \texttt{type3} = pairs of corrupted repetitive 
sentences; and \texttt{type4} =  pairs where the second sentence is 
randomly picked from the same document in arbitrary position. 

We define 5 training data variants by combining the types as follow: (1)
50K $\times$ \texttt{type1}; (2) 50K $\times$ \texttt{type2}; (3) 25K
$\times$ \texttt{type1} + 25K $\times$ \texttt{type2}; (4) 25K $\times$
\texttt{type2} + 5K $\times$ \texttt{type3} + 20K $\times$
\texttt{type4}; and (5) 25K $\times$ \texttt{type1} + 5K $\times$
\texttt{type3} + 20K $\times$ \texttt{type4}.

In preliminary experiments, we tested seven models (\bert, \roberta,
\textsc{ALBERT}, \xlnet, \textsc{ELECTRA}, \gpt, and \bart) for fine-tuning NSP
score. However, the results indicated that \bert performs the best for
NSP score (see Appendix D), so this forms the basis of our primary
results in the paper for inter-sentential coherence. First, we partition
our data into training, development, and test splits based on a ratio of
80:10:10, respectively, and fine-tune \texttt{bert-base-uncased} with
learning rate = 5e-5, batch size = 40, and maximum epochs = 20.
We simply use the \texttt{[CLS]} encoding as the input to an MLP layer.  
During training, we use early stopping (patience = 5) based on the 
development set performance. We run all models 5 times and achieve 
varied averaged F1 scores ranging from 75\% to 93\%, but found variant-5 
to achieve the best overall Pearson correlation (see \tabref{ic_mean}).

\begin{table}[t]

	\begin{center}
		\begin{adjustbox}{width=0.6\linewidth}
			\begin{tabular}{R{3cm}C{3cm}C{3cm}}
				\toprule
				\textbf{Data Variant} & \textbf{C-PG} & \textbf{C-BT} \\
				\midrule 
				1 & 0.22$\pm$0.06 & 0.27$\pm$0.01 \\
				2 & 0.23$\pm$0.02 & 0.28$\pm$0.04 \\
				3 & 0.25$\pm$0.12 & 0.30$\pm$0.12 \\
				4 & \textbf{0.41$\pm$0.01} & 0.26$\pm$0.07 \\
				5 & 0.39$\pm$0.07 & \textbf{0.35$\pm$0.05} \\
				\bottomrule
			\end{tabular}
		\end{adjustbox}
	\end{center}
	\caption{Averaged Pearson correlation scores for inter-sentential coherence and NSP-Score over 5 run models for the C-PG (CNN/DailyMail-PG) and C-BT (CNN/DailyMail-BERT) data, based on the five training data variants. }
	\label{tab:ic_mean}
\end{table}

\section{Experimental Result}

\subsection{Faithfulness}
\label{sec:res_faith}

\begin{table}[t]
	\begin{center}
	\begin{adjustbox}{width=0.7\linewidth}
		\begin{tabular}{L{7cm}R{2cm}R{2cm}}
			\toprule
			\textbf{Metric} & $r$ & $\rho$  \\
			\midrule
			\multicolumn{3}{l}{\textit{Against reference}} \\
			ROUGE-1 & 0.199 & 0.199 \\
			ROUGE-2 & 0.116 & 0.161 \\
			BLEU-4 & 0.072 & 0.133 \\
			METEOR & 0.131 & 0.170 \\
            \bertscore{}* & 0.128 & 0.131 \\
			\midrule
			\multicolumn{3}{l}{\textit{Against source sentences}} \\
			ROUGE-1 & --0.047 & --0.028 \\
			ROUGE-2 & 0.179 & 0.221 \\
			BLEU-4 & --0.107 & --0.138 \\
			METEOR & --0.018 & 0.006 \\
			\bertscore{}* & --0.034 & 0.006 \\
			FactCC & 0.042 & 0.045 \\
			FactCCX & --0.027 & --0.017 \\
			QA \shortcite{maynez-etal-2020-faithfulness} & --- & 0.044 \\
			Entailment \shortcite{maynez-etal-2020-faithfulness} & --- & 0.431 \\
			QAGS (our implementation) & 0.250 & 0.270 \\
			$\text{FA}_\text{STS}$ & 0.260 & 0.258 \\
			$\text{FA}_\text{ROUGE-1}$ & 0.361 & 0.361 \\
			$\text{FA}_\text{ROUGE-2}$ & 0.311 & 0.315 \\
            $\text{FA}_\bertscore${}* & 0.178 & 0.179 \\
			$\text{FA}_\bertscore$ (ours) & \textbf{0.476} & \textbf{0.474} \\
			\bottomrule
		\end{tabular}
	\end{adjustbox}
	\end{center}
	\caption{Pearson ($r$) and Spearman ($\rho$) correlation coefficients for faithfulness, measured between human judgement and various automatic metrics. `*' denotes that \bertscore uses \texttt{roberta-large} (layer 24 as recommended by \shortciteA{zhang2020bertscore}) while ours uses \texttt{roberta-base} (layer 10).} 
\label{tab:faith_cor}
\end{table}

\begin{figure}[ht]
	\centering
	\includegraphics[width=3.2in]{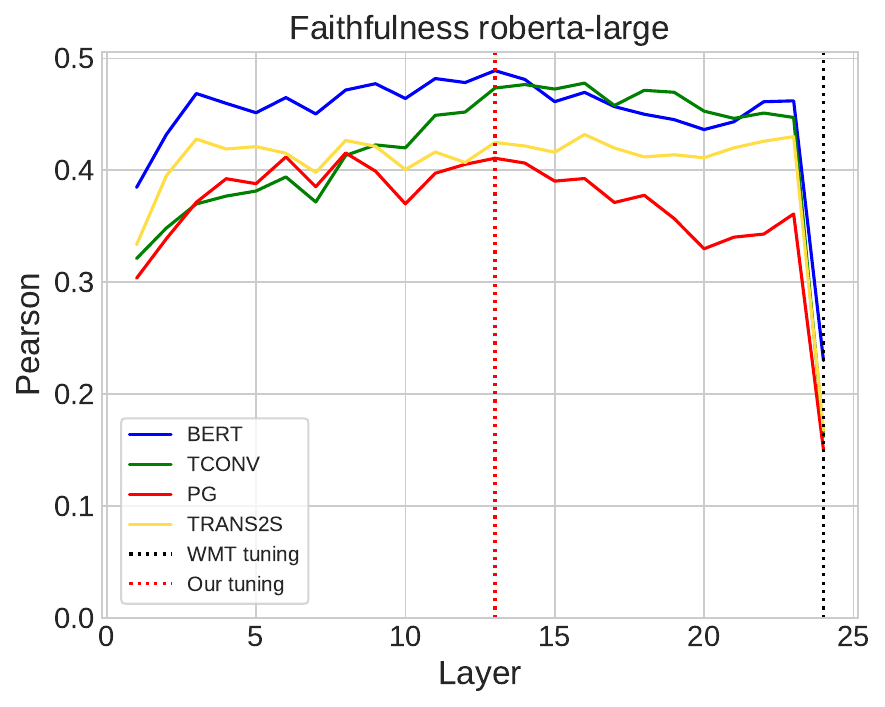}
	\caption{Pearson correlation based on each layer of \texttt{roberta-large} for faithfulness evaluation. Tuning refers to layer selection (i.e.\ model parameters are not updated.)}
	\label{fig:faith_roberta_large}
\end{figure}

In \tabref{faith_cor}, we show Pearson correlation scores ($r$) for 
faithfulness in two different forms: (1) evaluation against the reference; and (2) evaluation against the source. We additionally report the Spearman correlation ($\rho$), for direct comparison with 
\shortciteA{maynez-etal-2020-faithfulness}. We measure the correlation 
between human judgement (data in \secref{data_human}) and various automatic metrics. For evaluation against the reference, $\text{FA}_\text{ROUGE}$ and $\text{FA}_\bertscore$ are equivalent to ROUGE and \bertscore, respectively, because the XSUM dataset only consists of one-sentence summaries. 

As we can see in \tabref{faith_cor}, computing ROUGE, BLEU, METEOR and \bertscore conventionally (either against reference or full source sentences) yields low correlation scores.\footnote{For evaluation against the reference, we reproduce the Spearman correlation scores ($\rho$) of ROUGE-1 and ROUGE-2, but \bertscore is 0.131, slightly lower than \shortciteA{maynez-etal-2020-faithfulness} ($\rho=0.190$). We use the recommended model-layer by \shortciteA{zhang2020bertscore} while \shortciteA{maynez-etal-2020-faithfulness} do not report their \bertscore configuration (e.g. source code or model-layer used for evaluation).} We also found that our QAGS implementation \shortcite{wang2020asking} performs better than the QA system of \shortciteA{maynez-etal-2020-faithfulness} over their dataset. For FactCC and FactCCX, they have low correlation scores, with 0.04 being the highest correlation score. This is in line with a recent study by \shortciteA{pagnoni-etal-2021-understanding} who found that FactCC performs poorly (0.07 correlation score) over the XSUM dataset.

When we apply ROUGE, \stsscore, and \bertscore over the source document based on $\text{FA}_\text{METRIC}$ in \secref{faith}, we see that the $\text{FA}_\text{ROUGE-1}$ and $\text{FA}_\text{ROUGE-2}$ baselines actually outperform QAGS that is computationally expensive, but more importantly that the two versions of $\text{FA}_\bertscore$ perform differently, with our summarization-optimized version resulting in the best overall results. 
The first $\text{FA}_\bertscore$ uses the recommendations of
\shortciteA{zhang2020bertscore} (\texttt{roberta-large}; layer 24), but
we found this configuration to produce a lower correlation. The best layer for this model is layer-13, as depicted in \figref{faith_roberta_large}.

After conducting model-layer search over 19 models,\footnote{Emphasizing
	that there is no task-specific training for any of the \bertscore
	variants; we are simply selecting which layer of which pre-trained
	model to extract the word representations from.} we found that
\texttt{roberta-base} (layer 10) to result in the best correlation
(based on average rank across four summarization models).


Matching information in the summary sentence with the article sentence works best through $\text{FA}_{\bertscore}$  in \tabref{faith_cor}. Although $\text{FA}_{\bertscore}$  is computationally cheaper than question-answering based models such as QA and QAGS, we argue that deeper investigation is needed to more thoroughly evaluate $\text{FA}_{\bertscore}$, especially on different dataset with more varied article and summary lengths. As this paper is focused on introducing the FFCI framework, for faithfulness we use only the available data from \shortciteA{maynez-etal-2020-faithfulness}, and leave further validation to future work.

\subsection{Focus, Coverage, and Inter-sentential coherence}
\label{sec:visual}

\subsubsection{Dataset Visualization}

\begin{figure}[t]
	\centering
	\subfloat[]{\includegraphics[width=2.5in]{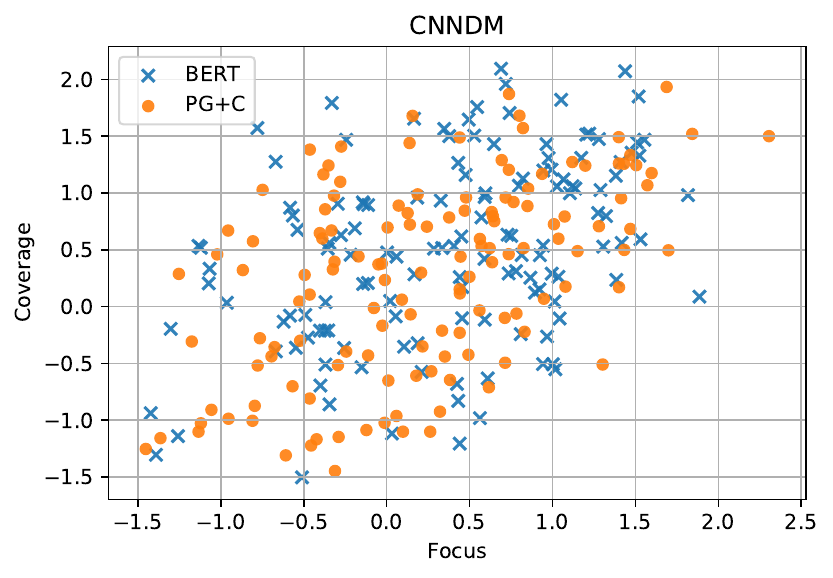}}
	\subfloat[]{\includegraphics[width=2.5in]{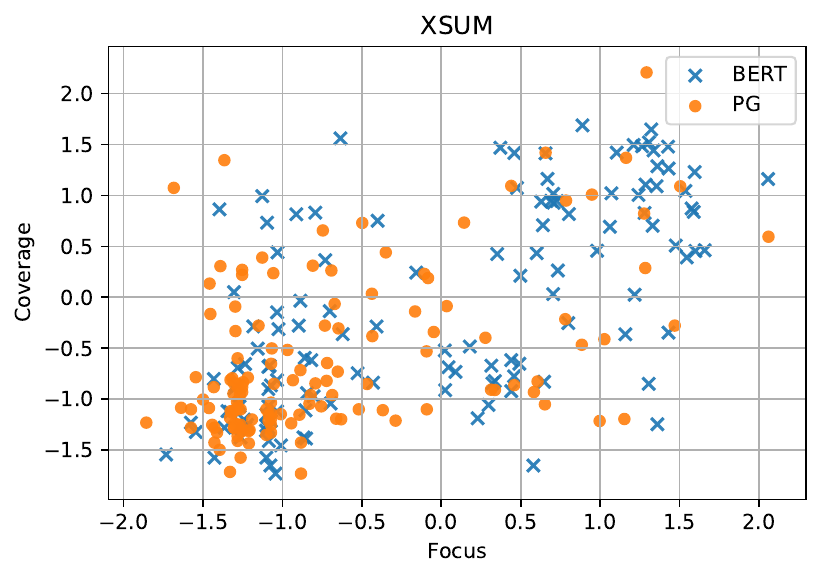}} \\
	\subfloat[]{\includegraphics[width=2in]{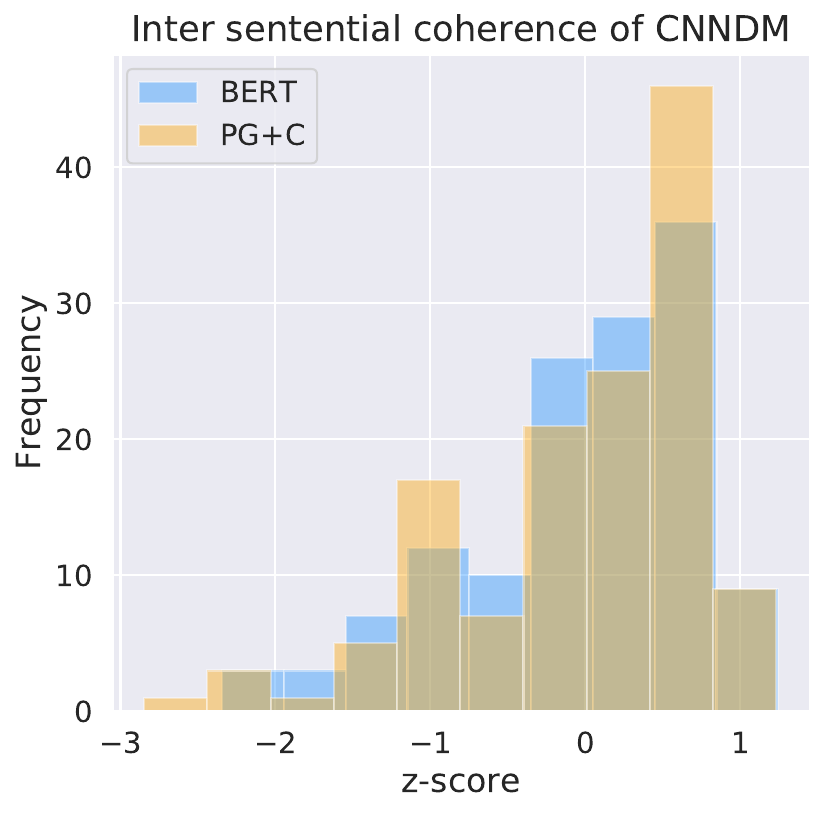}}
	\caption{Human annotation result for focus, coverage, and inter-sentential coherence.}
	\label{fig:annotation_res}
\end{figure}

In \figref{annotation_res}, we visualize the focus and coverage data for 
CNN/DailyMail and XSUM (after quality control, $z$-scoring, and averaging across 
annotators for a given summary), broken down across the two 
summarization systems that were used to generate the sample summaries. 
For CNN/DailyMail, the focus--coverage scores appear to be slightly 
higher (esp.\ for \bert), but are better separated over XSUM.
We also present the distribution of the inter-sentential coherence 
scores in \figref{annotation_res}, and again see that \bert and PG appear 
to be similar, with PG+Coverage appearing to be slightly better. We 
return to evaluate these trends more formally in \secref{leaderboard}.

\subsubsection{Metric Evaluation}

\begin{table*}[t!]
	\begin{center}
		\begin{adjustbox}{width=\linewidth}
			\begin{tabular}{lrrrrrrrrrrrr}
				\toprule
				\multirow{2}{*}{\textbf{Metric}} & \multicolumn{4}{c}{\textbf{Focus}} && \multicolumn{4}{c}{\textbf{Coverage}} && \multicolumn{2}{c}{\textbf{IC}} \\
				\cmidrule{2-5}
				\cmidrule{7-10}
				\cmidrule{12-13}
				& \textbf{C-PG} & \textbf{C-BT} & \textbf{X-PG} & \textbf{X-BT} & & \textbf{C-PG} & \textbf{C-BT} & \textbf{X-PG} & \textbf{X-BT} && \textbf{C-PG} & \textbf{C-BT} \\
				\midrule
				ROUGE-1 & 0.607 & 0.623 & 0.540 & 0.562 && 0.592 & 0.641 & 0.480 & 0.514 && 0.097 & 0.138 \\
				ROUGE-2 & 0.595 & 0.552 & 0.564 & 0.454 && 0.547 & 0.569 & 0.463 & 0.437 && $-$0.004 & 0.083 \\
				ROUGE-LCS & 0.604 & 0.619 & 0.528 & 0.552 && 0.581 & 0.636 & 0.482 & 0.487 && 0.088 & 0.114 \\
				METEOR & --- & --- & --- & --- && 0.597 & 0.660 & 0.523 & 0.601 && 0.061 & 0.143 \\
				BLEU-4 & 0.511 & 0.442 & 0.526 & 0.304 && --- & --- & --- & --- && $-$0.030 & 0.090 \\
				\midrule
				QAGS & 0.543 & 0.611 & 0.541 & 0.527 && 0.570 & 0.608 & 0.452 & 0.513 && --- & --- \\
				\midrule
				\stsscore (EDU) & 0.525 & 0.591 & 0.317 & 0.527 && 0.559 & 0.551 & 0.461 & 0.551 && --- & --- \\
				\stsscore (sentence) & 0.524 & 0.526 & 0.444 & \textbf{0.617} && 0.559 & 0.572 & 0.559 & \textbf{0.641} && --- & --- \\
				\stsscore (doc) & 0.524 & 0.569 & 0.444 & \textbf{0.617} && 0.513 & 0.508 & 0.559 & \textbf{0.641} && 0.124 & 0.197 \\ %
				\midrule
				\bertscore *  & 0.552 & 0.519 & 0.427 & 0.406 && 0.549 & 0.579 & 0.363 & 0.359 && 0.042 & 0.152 \\
				\bertscore  (Ours) & \textbf{0.665} & \bf 0.625 & \bf 0.577 & {0.581} && \textbf{0.680} & \bf 0.695 & \textbf{0.617} & 0.623 && 0.055 & 0.132 \\
				\midrule
				\shortciteA{nayeem-chali-2017-extract} & --- & --- & --- & --- && --- & --- & --- & --- && $-$0.275 & 0.166 \\
				NSP & --- & --- & --- & --- && --- & --- & --- & --- && \textbf{0.388} & \textbf{0.351} \\
				\bottomrule
			\end{tabular}
		\end{adjustbox}
	\end{center}
    \caption{Pearson correlation for focus, coverage, and 
    inter-sentential coherence, measured between human judgement and 
various automatic metrics. (``C-PG'' = CNN/DailyMail-PG; ``C-BT'' = 
CNN/DailyMail-BERT; ``X-PG'' = XSUM-PG; ``X-BT'' = XSUM-BERT; ``IC'' = 
Inter-sentential Coherence). All focus metrics are precision based, 
coverage metrics are recall based, and baselines for IC use F1. `*' uses 
\texttt{roberta-large} (layer 24), while ours use \texttt{gpt2-xl} 
(focus: layer 29, coverage: layer 4, IC: layer 47). 
\shortciteA{nayeem-chali-2017-extract} use $\lambda=0.5$.}
	\label{tab:result}
\end{table*}


In \tabref{result}, we present the meta-evaluation results for the primary metrics over focus, coverage, and inter-sentential coherence. 
We measure Pearson's $r$ between human judgements and the various automatic metrics discussed in \secref{metric}.

First, we observe that ROUGE, METEOR, and BLEU perform worse than the
model-based metrics in all cases.\footnote{ROUGE and METEOR scores are
	calculated based on the original implementations, while BLEU is based
	on SacreBLEU \shortcite{post-2018-call}.} For inter-sentential coherence in
particular, these baseline metrics perform expectedly badly, around
random. Our second observation is that QAGS performs poorly for focus
and coverage, compared to traditional metrics.\footnote{We experimented
	with different numbers of questions \textit{K}, ranging from 10 to 50,
	and also with different methods for pruning ill-formed questions.} The
correlation for QAGS is comparable to ROUGE-1 and only slightly better
than ROUGE-2. For \stsscore, the best segment granularity is sentence,
although there is little difference between the three granularities. The results for
\stsscore are excellent for \bert over XSUM, but appreciably worse for
other data--model pairs.


\begin{figure}[ht]
	\subfloat[]{\includegraphics[width = 0.5\linewidth]{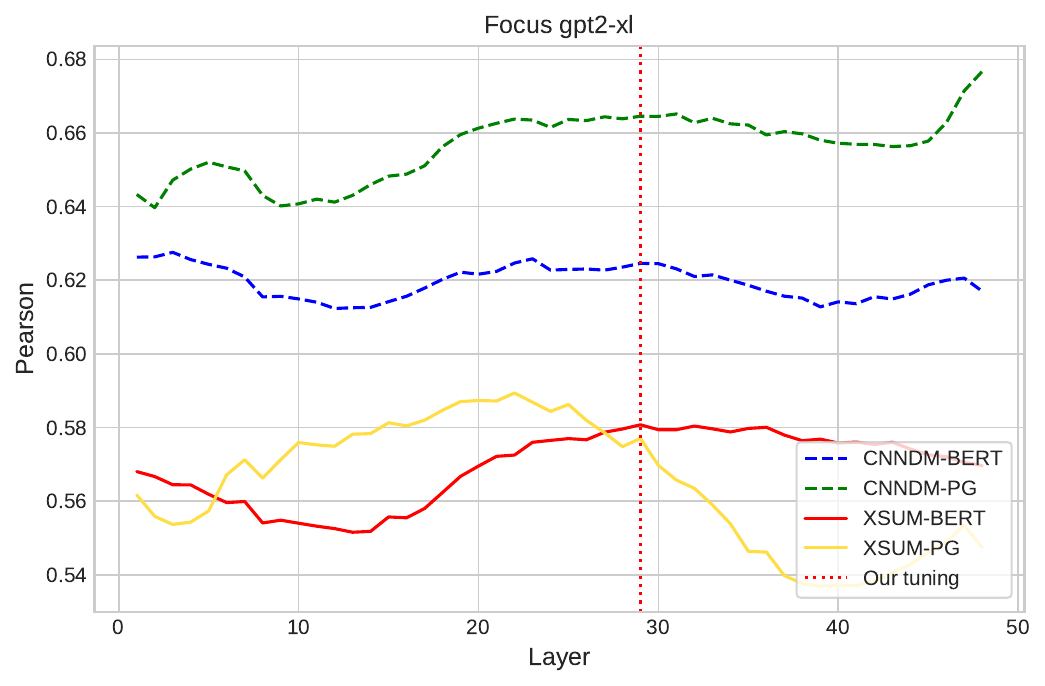}}
	\subfloat[]{\includegraphics[width = 0.5\linewidth]{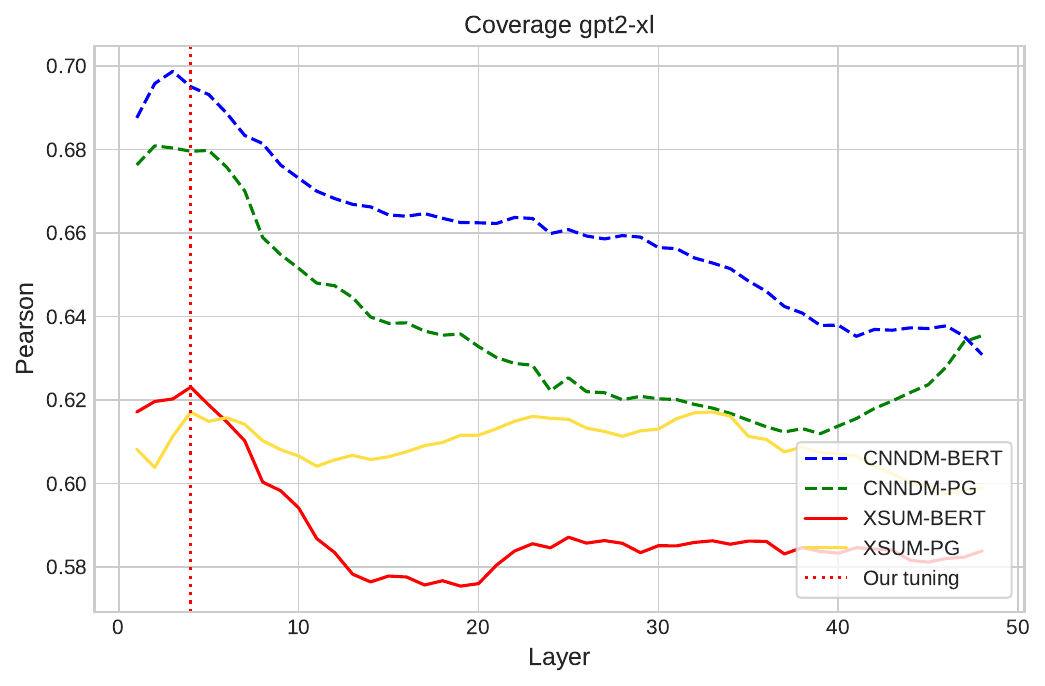}}\\
	\caption{Pearson correlation for each layer of \texttt{gpt2-xl} for focus and coverage evaluation. Tuning refers to layer selection (i.e.\ model parameters are not updated.) \shortciteA{zhang2020bertscore} does not report WMT tuning for this model.}
	\label{fig:gpt2xl}
\end{figure}

Our optimized version of \bertscore performs
better than the original due to the task-specific layer selection.
Similar to faithfulness (\secref{res_faith}), layer selection is 
conducted by selecting the average rank of 2 data $\times$  2 
summarization models = 4 options, for each 19 pre-trained models. We 
found that our \bertscore (\texttt{gpt2-xl}) performs the best for focus 
(layer 29) and coverage (layer 4) as depicted in \figref{gpt2xl}. We 
refer readers who are interested in other pre-trained language model 
scores to the Appendix .


Finally, we show the effectiveness of NSP prediction for
inter-sentential coherence. Computing \bertscore against the reference
results in low correlation, around random for CNN/DailyMail-PG. We found
that a simple NSP score consistently outperforms coherence score
\shortcite{nayeem-chali-2017-extract} at nearly double the correlation score (for
CNN/DailyMail-BERT). For CNN/DailyMail-PG,
\shortciteA{nayeem-chali-2017-extract} produces negative correlation
$r=-0.28$ which we suspect is due to severe repetition in PG
summaries, and the influence of NESim in the coherence score equation. Our
proposed NSP score handles this case better and achieves $r=0.388$ for
CNN/DailyMail-PG.

\subsection{Summarization Leaderboard Using FFCI}
\label{sec:leaderboard}

Having motivated the \ffci framework and developed robust metrics for each of the four elements, we next apply them in evaluating a broad range of contemporary methods over CNN/DailyMail and XSUM.

First, we collect summaries from the different abstractive and extractive summarization models either by downloading test outputs provided by the authors or applying checkpoint models from the authors to the test data to generate summaries. In each case, we ensure the test summaries result in similar ROUGE scores to those reported by the authors.

\begin{table*}[t]\footnotesize	
	\begin{center}
		\begin{adjustbox}{width=1\linewidth}
			\begin{tabular}{lrrrrrrrr}
				\toprule
				\multirow{2}{*}{\textbf{Method}} & \multicolumn{3}{c}{\textbf{ROUGE}} && \multicolumn{4}{c}{\textbf{FFCI}} \\
				\cmidrule{2-4}
				\cmidrule{6-9}
				& \textbf{R-1} & \textbf{R-2} & \textbf{R-L} && \textbf{Fa} &  \textbf{Fo} & \textbf{C} & \textbf{IC} \\
				\midrule
				\textsc{Lead3} & 40.1 & 17.3 & 36.3 && 91.2 & 49.2 & 70.9 & 65.3 \\
				\midrule
				\textbf{Abstractive} &  &  &  &&  &  &  & \\
				PG \shortcite{see-etal-2017-get} & 36.4 & 15.7 & 33.4 && 90.9 & 52.1 & 65.6 & 52.8 \\
				PG+C \shortcite{see-etal-2017-get} & 39.5 & 17.3 & 36.4 && \textbf{91.1} & 52.4 & 68.6 & 67.2 \\
				rnn+RL+rerank \shortcite{chen-bansal-2018-fast} & 40.9 & 17.8 & 38.5 && 89.6 & 53.4 & 70.2 & 56.4 \\
				\textsc{Bottom-Up} \shortcite{gehrmann-etal-2018-bottom} & 41.5 & 18.7 & 38.6 && 90.0 & 55.3 & 68.5 & 65.3 \\
				\textsc{BertSumExtAbs} \shortcite{liu-lapata-2019-text} & 42.1 & 19.4 & 39.1 && 89.8 & 51.9 & 68.7 & 65.7 \\
				BART \shortcite{lewis2019bart} & 44.3 & 21.1 & 41.2 && 89.5 & 52.6 & 69.5 & 69.6 \\
				PEGASUS \shortcite{zhang2020pegasus} & 44.4 & 21.5 & 41.4 && 89.9 & \textbf{56.0} & 70.8 & 69.5 \\
				\textsc{ProphetNet} \shortcite{yan2020prophetnet} & 44.4 & 21.2 & 41.5 && 89.9 & 55.9 & \textbf{72.0} & \textbf{70.0} \\
				\midrule
				\textbf{Extractive} &  &  &  &  &  &  &  \\
				BanditSum \shortcite{dong2018banditsum} & 41.6 & 18.7 & 37.9 && 91.8 & 51.5 & 71.6 & 61.5 \\
				PNBERT \shortcite{zhong-etal-2019-searching} & 42.7 & 19.5 & 38.8 && \textbf{91.9} & 51.9 & \textbf{73.5} & \textbf{66.2} \\
				\textsc{BertSumExt} \shortcite{liu-lapata-2019-text} & 43.3 & 20.2 & 39.7 && 91.8 & 52.2 & 73.0 & 61.8 \\
				\textsc{MatchSum} \shortcite{zhong-etal-2020-extractive} & 44.4 & 20.8 & 40.6 && 91.9 & \textbf{53.3} & 72.4 & 62.5 \\
				\bottomrule
			\end{tabular}
		\end{adjustbox}
	\end{center}	
	\caption{ROUGE and \ffci scores for various summarization models over CNN/DailyMail (``Fa'' = faithfulness; ``Fo'' = focus; ``C'' = coverage; and ``IC'' = inter-sentential coherence).}
\label{tab:leaderboard_CNN/DailyMail}
\end{table*}

\begin{table*}[t]\footnotesize
	\begin{center}
		\begin{adjustbox}{width=\linewidth}
			\begin{tabular}{lrrrrrrrr}
				\toprule
				\multirow{2}{*}{\textbf{Method}} & \multicolumn{3}{c}{\textbf{ROUGE}} && \multicolumn{4}{c}{\textbf{FFCI}} \\
				\cmidrule{2-4}
				\cmidrule{6-9}
				& \textbf{R-1} & \textbf{R-2} & \textbf{R-L} && \textbf{Fa} &  \textbf{Fo} & \textbf{C} & \textbf{IC} \\
				\midrule
				\textsc{Lead1} & 16.3 & 1.6 & 12.0 && 90.3 & 35.3 & 50.1 & --- \\
				\midrule
				PG  \shortcite{see-etal-2017-get} & 29.7 & 9.2 & 23.2 && 85.2 & 45.0 & 57.1 & --- \\
				TCONV  \shortcite{narayan-etal-2018-dont} & 31.9 & 11.5 & 25.8 && 85.2 & 49.4 & 57.7 & --- \\
				\textsc{BertSumExtAbs} \shortcite{liu-lapata-2019-text} & 38.8 & 16.5 & 31.3 && 85.6 & 53.7 & 62.3 & --- \\
				BART \shortcite{lewis2019bart} & 45.1 & 22.3 & 37.3 && \textbf{86.6} & 61.9 & {69.0} & --- \\
				PEGASUS \shortcite{zhang2020pegasus} & 47.2 & 24.6 & 39.3 && 86.5 & \textbf{64.6} & \textbf{69.5} & --- \\
				\bottomrule
			\end{tabular}
		\end{adjustbox}
	\end{center}
	\caption{ROUGE and \ffci scores for various summarization models over XSUM (``Fa'' = faithfulness; ``Fo'' = focus; ``C'' = coverage; and ``IC'' = inter-sentential coherence).}
	\label{tab:leaderboard_xsum}	
\end{table*}

In \tabref[s]{leaderboard_CNN/DailyMail} and \ref{tab:leaderboard_xsum} we 
provide FFCI-based results over CNN/DailyMail and XSUM respectively. At a 
glance, we can see that, despite the lacklustre results for ROUGE in our 
meta-evaluation, model development based on ROUGE has broadly led to 
positive progress in summarization, but we get a richer picture of the 
relative advantages of different methods.

First, the upper bound of faithfulness in CNN/DailyMail is around 91.0, as 
indicated by \textsc{Lead3} and the extractive models. Most abstractive 
models except PG+C obtain a faithfulness score lower than 91.0, but none 
lower than 89.0. As such, for CNN/DailyMail, faithfulness appears not to be a 
differentiating factor, although there has been a slight downward creep 
with recent abstractive methods. For XSUM, the upper bound for 
faithfulness is 90.3, for \textsc{Lead1}. We observe the faithfulness 
gap between \textsc{Lead1} and neural models is bigger than CNN/DailyMail, at 
around 5--6 points, but there has been a very slight upward trend in 
faithfulness for abstractive models.

In terms of coverage for abstractive methods, for CNN/DailyMail  there has been little improvement in recent years, with all models achieving below the \textsc{Lead3} baseline until the recently-proposed \textsc{ProphetNet} \shortcite{yan2020prophetnet}. Where progress has occurred for abstractive models over CNN/DailyMail is in focus, although results have fluctuated, with \textsc{BertSumExtAbs} \shortcite{liu-lapata-2019-text} performing notably badly in terms of focus, and to a lesser degree, coverage. This is despite ROUGE suggesting that the model performs better than \textsc{Bottom-Up} \shortcite{gehrmann-etal-2018-bottom}, for example. Another example of a substantial change-up in results is \bart vs. \textsc{PEGASUS}, where our \ffci framework shows that \textsc{PEGASUS} is substantially better in terms of both focus and coverage, despite the ROUGE scores being almost identical.

In contrast with the abstractive models, the extractive models tend to have higher coverage and lower focus. While \textsc{MatchSum} \shortcite{zhong-etal-2020-extractive} is state of the art in terms of ROUGE, based on our evaluation, coverage is actually markedly lower than competitor methods but focus is high.

For XSUM, we can observe large improvements in focus in particular, and relatively smaller but still clear improvements in coverage. Faithfulness, on the other hand, has improved only slightly, and there is clear room for improvement. Once again, our framework clearly shows that \textsc{PEGASUS} achieves better focus than \bart, but that they are closer in terms of coverage. 

Lastly, we look at inter-sentential coherence for CNN/DailyMail.\footnote{Recalling that XSUM summaries are single sentence, and thus inter-sentential coherence is irrelevant.} Overall, three abstractive models achieve IC score close to \textsc{Lead3}: PG+C, \textsc{Bottom-Up}, and \textsc{BertSumExtAbs}. PG and rnn+RL+rerank \shortcite{chen-bansal-2018-fast} result in very poor inter-sentential coherence, which we suspect is due to severe repetition at the decoding stage. \bart, \textsc{PEGASUS} and \textsc{ProphetNet} achieve higher scores than \textsc{Lead3}, with a gap of around 4 points. As expected, the extractive methods tend to result in poorer  inter-sentential coherence than \textsc{Lead3}, with the exception of \textsc{PNBERT}.

\subsection{Analysis of Inter-sentential Coherence}

One surprising observation from \tabref{leaderboard_CNN/DailyMail} is
that the IC score of \textsc{Lead3} is actually lower than that of
\textsc{ProphetNet}, despite \textsc{Lead3} consisting of the first
three sentences of the source document (which we would assume have high
IC). Additionally, the raw correlation numbers for the IC experiments
from \tabref{result} are low, suggesting that the scores from the IC
metrics are prone to noise and potentially unreliable.  Given this, we
performed additional manual annotations for IC over four summarization
methods to better understand the results: \textsc{Lead3}, PG+C,
\textsc{BertSumExtAbs}, and \textsc{ProphetNet}. We use the same 135
CNN/DailyMail samples as in \secref{data_human}, and ask three workers
to annotate IC using the same procedure as \secref{design} (135
documents $\times$ 4 models $\times$ 3 annotators, resulting in 1,620
annotations). In this annotation, we achieve a quality score of 96.7 and
average working time of 31.7 minutes.

\begin{table*}[t]\footnotesize
	\begin{center}
		\begin{adjustbox}{width=0.6\linewidth}
			\begin{tabular}{lcrr}
				\toprule
				\multirow{3}{*}{\bf Model} & \multicolumn{3}{c}{\bf IC Score} \\
				\cmidrule{2-4}
				& \multirow{2}{*}{\bf Automatic} & \multicolumn{2}{c}{\bf Human ($z$-score)} \\
				\cmidrule{3-4}
				& & \bf Sum & \bf Avg \\
				\midrule
				\textsc{Lead3} & 65.3 & 0.45 & $-$0.0030 \\
				PG+C & 67.2 & $-$3.40 & $-$0.0026 \\
				\textsc{BertSumExtAbs} & 65.7 & $-$4.50 & $-$0.0330 \\
				\textsc{ProphetNet} & \bf 70.0 & \bf 7.50 & \bf 0.0600 \\
				\bottomrule
			\end{tabular}
		\end{adjustbox}
	\end{center}
	\caption{Inter-sentential coherence (IC) based on automatic and manual scores.}
	\label{tab:ic_analysis}	
\end{table*}

\begin{table*}[t]\footnotesize
	\begin{center}
		\begin{adjustbox}{width=0.9\linewidth}
			\begin{tabular}{C{7cm}C{7cm}}
				\toprule
				\bf \textsc{Lead3} & \bf \textsc{ProphetNet}  \\
				\midrule
				\begin{itemize}
					\item This is the breathtaking moment a diver came face to face with a 'fish tornado' off the mexican coast.
					\item Tori hester, 25, from san diego, california, was diving in cabo pulmo when the huge school of trevally fish began circling above her.
					\item Husband jeff, 26, was on hand to capture the incredible moment using his underwater camera.
				\end{itemize}				
				& 
				\begin{itemize}
					\item Tori hester, 25, from san diego, california, was diving in cabo pulmo.
					\item Huge school of trevally fish began circling above her.
					\item Husband jeff, a marine scientist, was on hand to capture the moment.
				\end{itemize} \\
				\midrule
				\begin{itemize}
					\item London (CNN).
					\item Congolese immigrant tarsis mboma thale has a small business selling t-shirts in johannesburg, south africa.
					\item Thale's job normally requires him to walk the streets of the city he has called home for the past few years.
				\end{itemize}
				&
				\begin{itemize}
					\item Wave of anti-immigrant violence has swept south africa in recent days, leaving several dead.
					\item Some blame alleged inflammatory comments about foreign nationals from the zulu king.
					\item Others say a labor dispute between locals and foreigners back in march turned nasty.
				\end{itemize} \\
				\bottomrule
			\end{tabular}
		\end{adjustbox}
	\end{center}
	\caption{Example \textsc{Lead3} and \textsc{ProphetNet} summaries for CNN/DailyMail.}
	\label{tab:ic_example}	
\end{table*}

Firstly, in \tabref{ic_analysis} we observe that the manual annotations
for PG+C and \textsc{BertSumExtAbs} are consistent with
\figref{annotation_res} from \secref{visual}, in that PG+C tends to have
slightly higher inter-sentential coherence. More importantly,
\tabref{ic_analysis} confirms that \textsc{ProphetNet} does indeed outperform all
other models including \textsc{Lead3}. The reason that \textsc{Lead3} has lower IC is that the resultant summaries often contain subtle
disfluencies, as shown in \tabref{ic_example}. The first \textsc{Lead3}
example contains a ``teaser'' first sentence (which is disconnected from the next two sentences), while the second example has metadata in the first sentence. Compared with this, the \textsc{ProphetNet} examples are more fluent in terms of information structure.

Although the analysis in \tabref{ic_analysis} is highly promising in 
terms of the veracity of the automatic metric, this is the 
least-developed of the four \ffci metrics with plenty of room for further 
improvement (owing to its low absolute correlation values (0.35--0.38), 
as seen in \tabref{result}).


\section{Conclusion}
We introduce the \ffci evaluation framework for summarization evaluation, based on the four elements of: faithfulness, focus, coverage, and inter-sentential coherence. We have shown that \bertscore (\texttt{roberta-base}) is the most robust metric for evaluating faithfulness, \bertscore (\texttt{gpt2-xl})  for focus and coverage, and NSP-score for inter-sentential coherence.

Our general finding is that ROUGE has lead to positive progress in 
modern summarization systems but lacks fine-grained interpretability.  
\ffci shows that  since the LSTM-based seq2seq, modern abstractive 
summarization systems over CNN/DailyMail have largely improved on focus, 
with coverage not being much better than \textsc{Lead3} until recent 
systems (e.g.\ \textsc{ProphetNet}). Our \ffci framework found  three 
competitive state-of-the-art systems: \bart, \textsc{PEGASUS}, and 
\textsc{ProphetNet}, with \textsc{PEGASUS} and \textsc{ProphetNet} having 
generally higher focus and coverage respectively.

Lastly, although FFCI was designed based on our survey of evaluation approaches in previous work (\tabref{prev_works}), we believe there are some additional aspects that should be addressed in future work such as redundancy and relevance. Our work and the aforementioned survey are mostly based on news datasets such as CNN/DM and XSUM with relatively short summaries. We believe the redundancy in particular becomes very important as summarisation research shifts focus to longer documents and summaries.



\acks{In this research, the first author is
	supported by the Australia Awards Scholarship (AAS), funded by the Department of Foreign Affairs and Trade (DFAT), Australia.
	This research was undertaken using the LIEF HPC-GPGPU Facility hosted at The University of Melbourne. This facility was established with the assistance of LIEF Grant LE170100200.
}

\newpage
\appendix
\section*{Appendix A. Recommended Layers for Faithfulness, Focus, and Coverage}

\begin{table}[ht]
	\footnotesize
	\begin{center}
		\begin{adjustbox}{width=0.45\linewidth}
			\begin{tabular}{lrrr}
				\toprule
				\textbf{Model} & \textbf{Fa} & \textbf{Fo} & \textbf{C}  \\
				\midrule
				\texttt{bert-base-uncased} & 6 & 1 & 2 \\
				\texttt{bert-large-uncased} & 11 & 9 & 9 \\
				\midrule
				\texttt{roberta-base} &\cellcolor{LightCyan} 10 & 9 & 2 \\
				\texttt{roberta-large} & 13 & 13 & 3 \\
				\texttt{roberta-large-mnli} & 14 & 15 & 3 \\
				\midrule
				\texttt{xlnet-base-cased} & 6 & 4 & 2 \\
				\texttt{xlnet-large-cased} & 7 & 7 & 5 \\
				\midrule
				\texttt{gpt2} & 1 & 3 & 3 \\
				\texttt{gpt2-medium} & 8 & 5 & 1 \\
				\texttt{gpt2-large} & 2 & 21 & 3 \\
				\texttt{gpt2-xl} & 2 & \cellcolor{LightCyan}29 & \cellcolor{LightCyan}4 \\
				\midrule
				\texttt{t5-small} & 2 & 3 & 2 \\
				\texttt{t5-base} & 3 & 4 & 4 \\
				\texttt{t5-large} & 10 & 13 & 10 \\
				\midrule
				\texttt{bart-base} & 1 & 3 & 1 \\
				\texttt{bart-large} & 2 & 5 & 2 \\
				\midrule
				\texttt{pegasus-xsum} & 8 & 11 & 6 \\
				\texttt{pegasus-cnn\_dailymail} & 12 & 11 & 5 \\
				\texttt{pegasus-large} & 3 & 4 & 4 \\ 			
				
				\bottomrule
			\end{tabular}
		\end{adjustbox}
	\end{center}
	\caption{Recommended layers for faithfulness (Fa), focus (Fo), and coverage (Co). }
	\label{tab:all_layers}
\end{table}

As discussed by \shortciteA{zhang2020bertscore} and \shortciteA{reimers-2019-sentence-bert}, layer selection in \bert model is important. \bertscore was designed to maximize the Pearson correlation between $\fscore_{\text{BERT}}$ and WMT16, which is potentially less than optimal for evaluating focus and coverage in summarization. 

\begin{figure}[ht]
	\subfloat[]{\includegraphics[width = 0.33\linewidth]{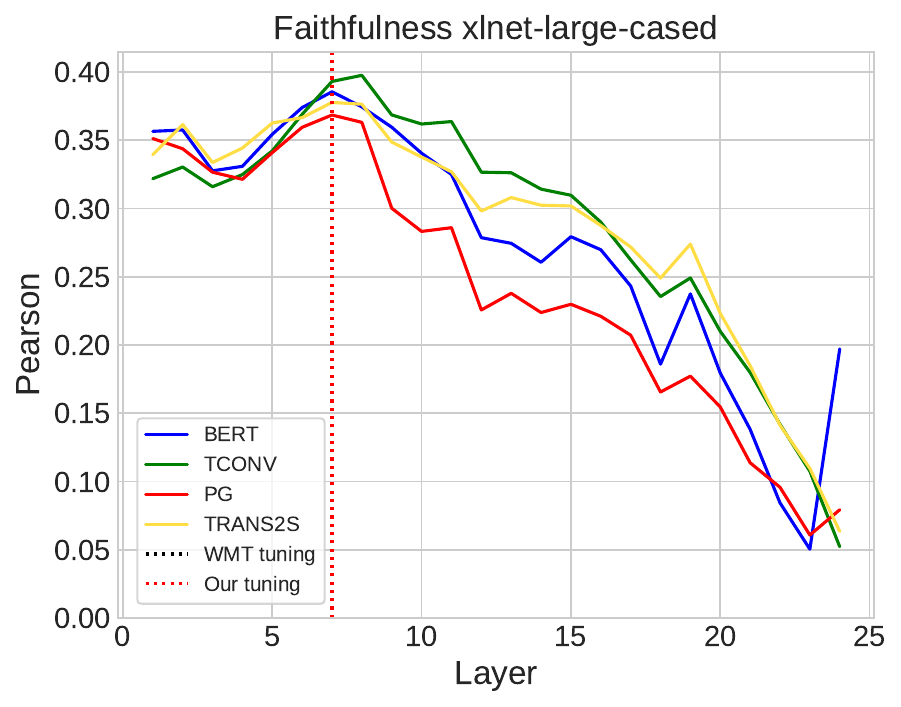}}
	\subfloat[]{\includegraphics[width = 0.33\linewidth]{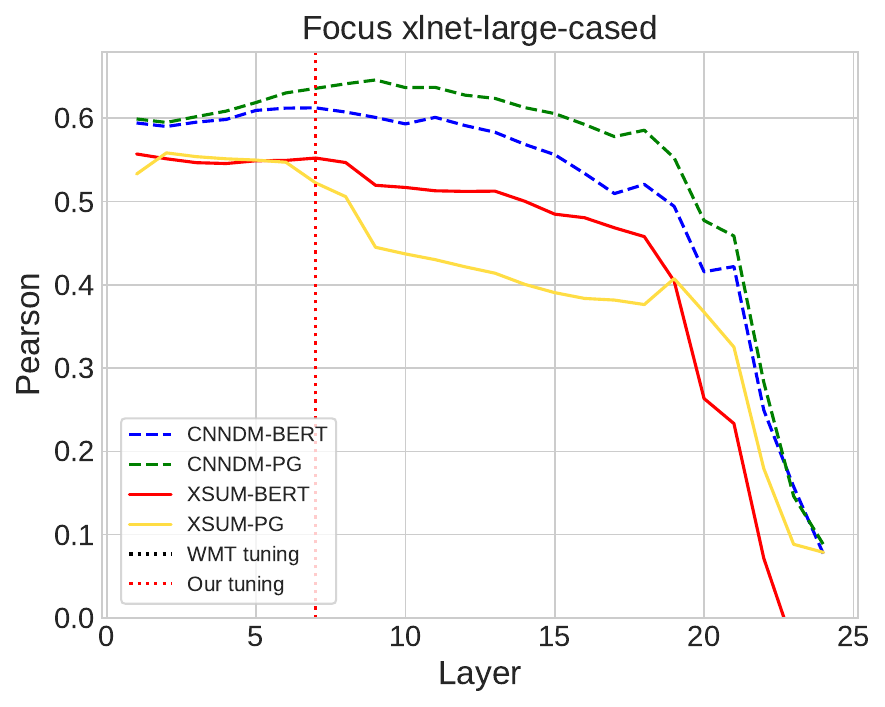}}
	\subfloat[]{\includegraphics[width = 0.33\linewidth]{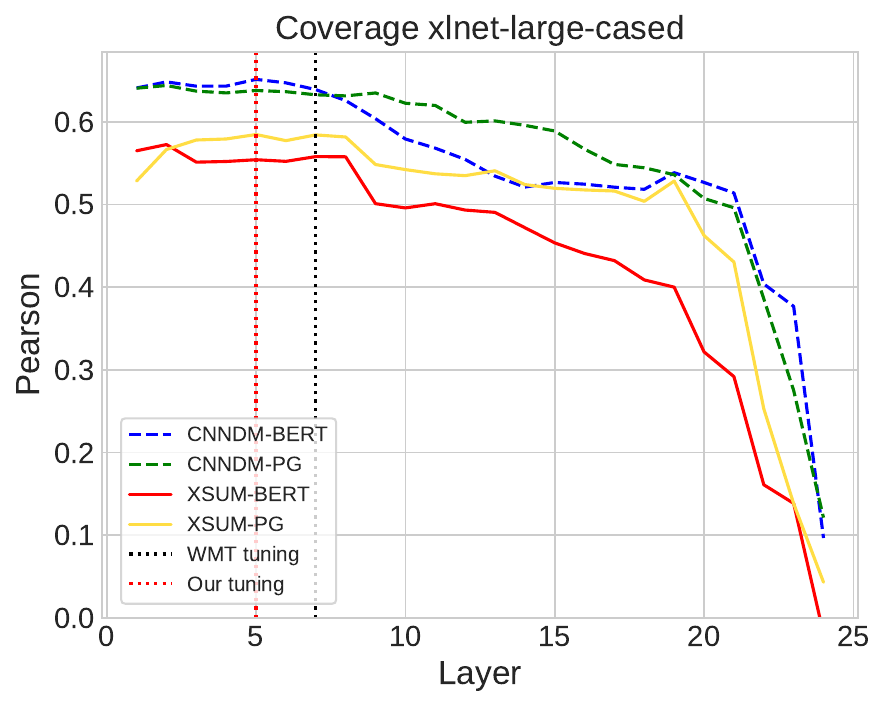}}\\
	\caption{\texttt{xlnet-large-cased}.}
	\label{fig:xlnet-large-cased}
\end{figure}

\begin{figure}[ht]
	\subfloat[]{\includegraphics[width = 0.33\linewidth]{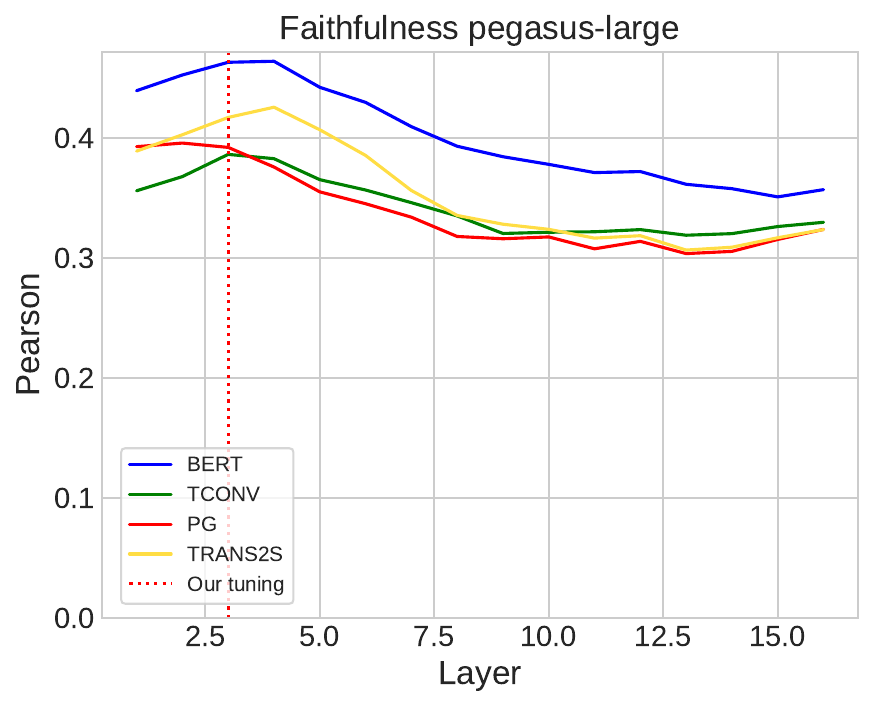}}
	\subfloat[]{\includegraphics[width = 0.33\linewidth]{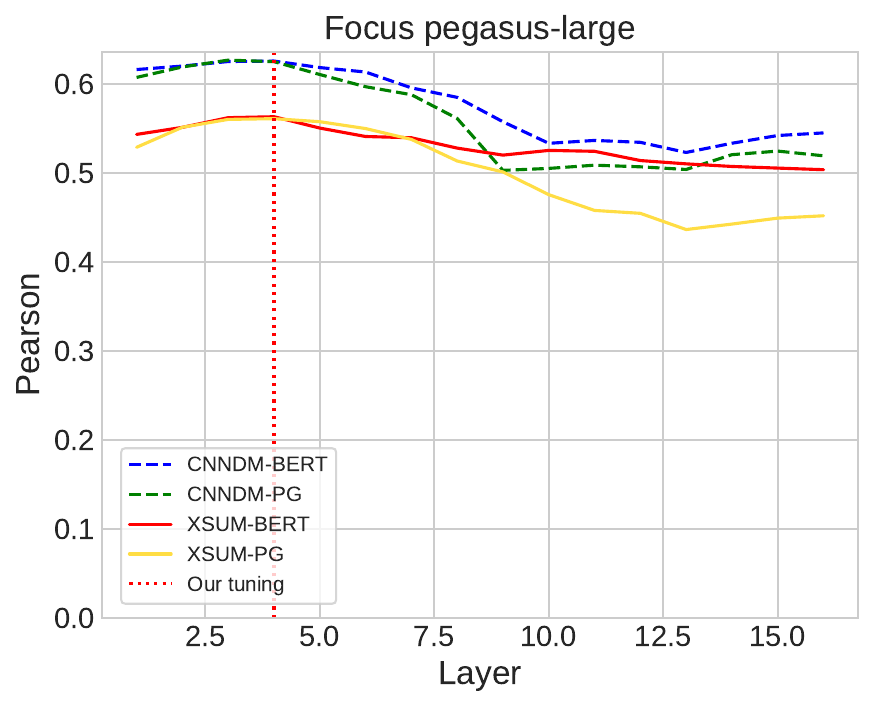}}
	\subfloat[]{\includegraphics[width = 0.33\linewidth]{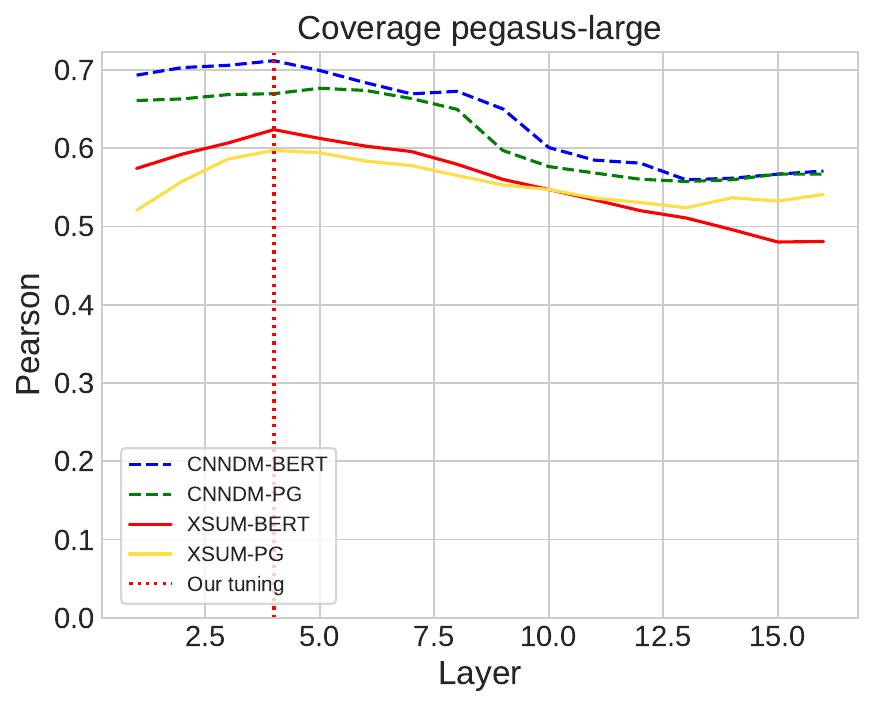}}\\
	\caption{\texttt{pegasus-large}.}
	\label{fig:pegasus-large}
\end{figure}

In \tabref{all_layers}, we present 19 models and their recommended layer number for evaluating faithfulness, focus, and coverage. Specifically, we use $\text{FA}_{\text{MODEL}}$, $\precision_{\text{MODEL}}$ and $\recall_{\text{MODEL}}$ to calculate the Pearson correlation, respectively. To pick the best layer, we simply average the rank of each data--model based on the outputs of a given layer. We observe that almost all selected layers are different to \bertscore (see \figref[s]{bert-base-uncased}--\ref{fig:xlnet-large-cased} for examples). The optimal layer for focus tends to be one of the last layers, while earlier layers tend to work better for coverage. We also present the layer selection plots for the non-\bert models (\figref[s]{gpt2}--\ref{fig:pegasus-large})

\newpage
\section*{Appendix B. Pre-trained Language Model Scores of Faithfulness }

\begin{table*}[ht]
	\footnotesize
	\begin{center}
		\begin{adjustbox}{width=1\linewidth}
			\begin{tabular}{lC{2cm}C{2cm}C{2cm}C{2cm}}
				\toprule
				\multirow{2}{*}{\textbf{Model}} & \multicolumn{4}{c}{\textbf{Faithfulness}} \\
				\cmidrule{2-5}
				& \textbf{PG} & \textbf{TRANS2S} & \textbf{TCONV} & \textbf{BERT} \\
				\midrule
				\multicolumn{5}{l}{\textit{Results based on our layer selection}}\\
				\midrule
				\texttt{bert-base-uncased} & 0.424 & 0.394 & 0.460 & 0.463 \\
				\texttt{bert-large-uncased} & 0.420 & 0.406 & 0.436 & 0.473 \\
				\midrule
				\rowcolor{LightCyan}
				\texttt{roberta-base} & \bf 0.459 & \bf 0.450 & \bf 0.519 & 0.475 \\
				\texttt{roberta-large} & 0.411 & 0.425 & 0.474 & 0.489 \\
				\texttt{roberta-large-mnli} & 0.437 & 0.415 & 0.489 & \textbf{0.477} \\
				\midrule
				\texttt{xlnet-base-cased} & 0.355 & 0.347 & 0.372 & 0.373 \\
				\texttt{xlnet-large-cased} & 0.369 & 0.378 & 0.393 & 0.386 \\
				\midrule
				\texttt{gpt2} & 0.299 & 0.341 & 0.331 & 0.367 \\
				\texttt{gpt2-medium}  & 0.329 & 0.412 & 0.357 & 0.396 \\
				\texttt{gpt2-large} & 0.360 & 0.399 & 0.386 & 0.439 \\
				\texttt{gpt2-xl} & 0.357 & 0.392 & 0.381 & 0.431 \\
				\midrule
				\texttt{t5-small} & 0.326 & 0.307 & 0.330 & 0.328 \\
				\texttt{t5-base} & 0.334 & 0.332 & 0.346 & 0.353 \\
				\texttt{t5-large} & 0.346 & 0.344 & 0.355 & 0.354 \\
				\midrule
				\texttt{bart-base} & 0.370 & 0.383 & 0.381 & 0.421 \\
				\texttt{bart-large} & 0.375 & 0.412 & 0.405 & 0.452 \\
				\midrule
				\texttt{pegasus-xsum} & 0.406 & 0.410 & 0.437 & 0.417 \\
				\texttt{pegasus-cnn\_dailymail} & 0.401 & 0.406 & 0.432 & 0.413 \\
				\texttt{pegasus-large} & 0.392 & 0.417 & 0.387 & 0.463 \\
				\midrule
				\multicolumn{5}{l}{\textit{Results based on recommended layers by \shortciteA{zhang2020bertscore}}} \\
				\midrule
				\texttt{bert-base-uncased}  & 0.386 & 0.377 & 0.435 & 0.440 \\
				\texttt{bert-large-uncased}  & 0.251 & 0.335 & 0.380 & 0.378 \\
				\midrule
				\rowcolor{LightCyan}
				\texttt{roberta-base}  & 0.459 & 0.450 & 0.519 & 0.475 \\
				\texttt{roberta-large}  & 0.150 & 0.168 & 0.162 & 0.230 \\
				\texttt{roberta-large-mnli}  & 0.370 & 0.340 & 0.440 & 0.423 \\
				\midrule
				\texttt{xlnet-base-cased}  & 0.281 & 0.303 & 0.355 & 0.328 \\
				\texttt{xlnet-large-cased}  & 0.369 & 0.378 & 0.393 & 0.386 \\
				\bottomrule
			\end{tabular}
		\end{adjustbox}
	\end{center}
	\caption{Pearson correlation of all experimental results on pre-trained language model scores for faithfulness (XSUM data). We highlight models with the highest average across data--model pairs. \texttt{roberta-base} of \shortciteA{zhang2020bertscore} and ours use the same layer-10. Please note that \shortciteA{zhang2020bertscore} final recommendation is to use \texttt{roberta-large} (layer-24).}
	\label{tab:all_result_faith}
\end{table*}

\newpage
\section*{Appendix C. Pre-trained Language Model Scores of Focus and Coverage }

\begin{table*}[ht!]
	\footnotesize
	\begin{center}
		\begin{adjustbox}{width=\linewidth}
			\begin{tabular}{lrrrrrrrrr}
				\toprule
				\multirow{2}{*}{\textbf{Model}} & \multicolumn{4}{c}{\textbf{Focus}} && \multicolumn{4}{c}{\textbf{Coverage}}  \\
				\cmidrule{2-5}
				\cmidrule{7-10}
				& \textbf{C-PG} & \textbf{C-BT} & \textbf{X-PG} & \textbf{X-BT} &&  \textbf{C-PG} & \textbf{C-BT} & \textbf{X-PG} & \textbf{X-BT} \\
				\midrule
				\multicolumn{9}{l}{\textit{Results based on layer selection}} \\
				\midrule
				\texttt{bert-base-uncased} & 0.623 & 0.647 & 0.513 & 0.529 && 0.636 & 0.680 & 0.587 & 0.578 \\
				\texttt{bert-large-uncased} & 0.627 & 0.641 & 0.547 & 0.563 && 0.648 & 0.689 & 0.608 & 0.609 \\
				\midrule
				\texttt{roberta-base} & 0.621 & 0.636 & 0.531 & 0.550 && \textbf{0.698} & 0.707 & 0.553 & 0.596 \\
				\texttt{roberta-large} & 0.634 & 0.643 & 0.583 & 0.552 && 0.674 & \textbf{0.712} & 0.588 & 0.603 \\
				\texttt{roberta-large-mnli} & 0.647 & \textbf{0.658} & 0.573 & 0.557 && 0.677 & 0.706 & 0.591 & 0.610 \\
				\midrule
				\texttt{xlnet-base-cased} & 0.640 & 0.603 & 0.508 & 0.531 && 0.636 & 0.639 & 0.566 & 0.533 \\
				\texttt{xlnet-large-cased} & 0.636 & 0.612 & 0.522 & 0.552 && 0.638 & 0.651 & 0.585 & 0.554 \\
				\midrule
				\texttt{gpt2} & 0.621 & 0.620 & 0.493 & 0.528 && 0.648 & 0.678 & 0.534 & 0.562 \\
				\texttt{gpt2-medium} & 0.636 & 0.616 & 0.579 & 0.544 && 0.667 & 0.687 & 0.552 & 0.603 \\
				\texttt{gpt2-large} & \textbf{0.668} & 0.629 & \textbf{0.577} & 0.571 && 0.676 & 0.689 & \textbf{0.614} & 0.612 \\
				\rowcolor{LightCyan}
				\texttt{gpt2-xl} & \textbf{0.665} & 0.625 & \textbf{0.577} & 0.581 && 0.680 & 0.695 & \textbf{0.617} & \textbf{0.623} \\
				\midrule
				\texttt{t5-small} & 0.632 & 0.603 & 0.529 & 0.582 && 0.641 & 0.636 & 0.580 & 0.591 \\
				\texttt{t5-base} & 0.632 & 0.608 & 0.548 & 0.586 && 0.641 & 0.653 & 0.580 & 0.596 \\
				\texttt{t5-large} & 0.643 & 0.615 & 0.544 & \textbf{0.591} && 0.646 & 0.641 & 0.600 & 0.604 \\
				\midrule
				\texttt{bart-base} & 0.664 & 0.629 & 0.541 & 0.546 && 0.674 & 0.688 & 0.577 & 0.578 \\
				\texttt{bart-large} & 0.655 & 0.627 & 0.550 & 0.561 && 0.676 & 0.690 & 0.599 & 0.609 \\
				\midrule
				\texttt{pegasus-xsum} & 0.630 & 0.643 & 0.556 & 0.563 && 0.664 & 0.676 & 0.587 & 0.603 \\
				\texttt{pegasus-cnn\_dailymail} & 0.652 & 0.640 & 0.534 & 0.571 && 0.668 & 0.700 & 0.580 & 0.604 \\
				\texttt{pegasus-large} & 0.625 & 0.626 & 0.561 & 0.563 && 0.669 & \textbf{0.712} & 0.597 & \textbf{0.624} \\ 			
				\midrule
				\multicolumn{9}{l}{\textit{Results based on recommended layers by \shortciteA{zhang2020bertscore}}} \\
				\midrule
				\texttt{bert-base-uncased} & 0.613 & 0.624 & 0.478 & 0.532 && 0.583 & 0.665 & 0.601 & 0.553 \\
				\texttt{bert-large-uncased} & 0.602 & 0.630 & 0.502 & 0.524 && 0.610 & 0.648 & 0.619 & 0.524 \\
				\midrule
				\texttt{roberta-base} & 0.616 & 0.638 & 0.516 & 0.556 && 0.677 & 0.667 & 0.544 & 0.559 \\
				\texttt{roberta-large} & 0.619 & 0.641 & 0.584 & 0.548 && 0.694 & 0.691 & 0.562 & 0.577 \\
				\texttt{roberta-large-mnli} & 0.608 & 0.653 & 0.529 & 0.537 && 0.628 & 0.668 & 0.562 & 0.554 \\
				\midrule
				\texttt{xlnet-base-cased} & 0.644 & 0.608 & 0.461 & 0.507 && 0.603 & 0.628 & 0.529 & 0.489 \\
				\texttt{xlnet-large-cased} & 0.636 & 0.612 & 0.522 & 0.552 && 0.633 & 0.639 & 0.584 & 0.558 \\
				\bottomrule
			\end{tabular}
		\end{adjustbox}
	\end{center}
	\caption{Pearson correlation of pre-trained language model scores for focus and coverage (``C-PG'' = CNNDM-PG; ``C-BT''= CNNDM-BERT; ``X-PG'' = XSUM-PG; ``X-BT'' = XSUM-BERT). We highlight models with the highest average across data--model pairs.}
	\label{tab:all_result_focus_coverage}
\end{table*}

\newpage
\section*{Appendix D. Full Results Over Inter-Sentential Coherence}
\label{app:nsp_details}
\begin{table*}[ht]
	\footnotesize
	\begin{center}
		\begin{adjustbox}{width=0.85\linewidth}
			\begin{tabular}{lC{3cm}C{3cm}}
				\toprule
				\textbf{Model} & \textbf{PG} & \textbf{BERT} \\
				\midrule
				\multicolumn{3}{l}{\bf NSP-Score (mean)}\\
				\midrule
				 \cellcolor{LightCyan}\texttt{bert-base-uncased} &  \cellcolor{LightCyan} 0.388 $\pm$ 0.069 &  \cellcolor{LightCyan} \bf 0.351 $\pm$ 0.051 \\
				\texttt{roberta-base} &  0.339 $\pm$ 0.037 & 0.230 $\pm$ 0.061\\
				\texttt{albert-base-v2} & 0.331 $\pm$ 0.049 & 0.200 $\pm$ 0.045 \\
				\texttt{xlnet-base-cased} & 0.365  $\pm$ 0.051 & 0.235 $\pm$ 0.070 \\
				\texttt{electra-base-discriminator} & \bf 0.389 $\pm$ 0.053 & 0.305 $\pm$ 0.038 \\
				\texttt{gpt2} & 0.313 $\pm$ 0.008 &	0.114 $\pm$ 0.024 \\
				\texttt{bart-base} & 0.357 $\pm$ 0.069 & 0.256 $\pm$ 0.068 \\
				\midrule
				\multicolumn{3}{l}{\bf NSP-Score (max)}\\
				\midrule
				\texttt{bert-base-uncased} & 0.269 $\pm$ 0.085 & 0.342 $\pm$ 0.045 \\
				\texttt{roberta-base} &  0.339 $\pm$ 0.037 & 0.230 $\pm$ 0.061\\
				\texttt{albert-base-v2} & 0.336 $\pm$ 0.052 & 0.243 $\pm$ 0.054 \\
				\texttt{xlnet-base-cased} & 0.247  $\pm$ 0.059 & 0.219 $\pm$ 0.069 \\
				\texttt{electra-base-discriminator} & 0.338 $\pm$ 0.032 & 0.311 $\pm$ 0.050 \\
				\texttt{gpt2} & 0.212 $\pm$ 0.013 &	0.064 $\pm$ 0.046 \\
				\texttt{bart-base} & 0.292 $\pm$ 0.055 & 0.267 $\pm$ 0.097 \\
				\midrule
				\multicolumn{3}{l}{\bf NSP-Score (min)}\\
				\midrule
				\texttt{bert-base-uncased} & 0.375 $\pm$ 0.049 & 0.245 $\pm$ 0.052 \\
				\texttt{roberta-base} &  0.261 $\pm$ 0.053 & 0.148 $\pm$ 0.075\\
				\texttt{albert-base-v2} & 0.293 $\pm$ 0.064 & 0.151	 $\pm$ 0.032 \\
				\texttt{xlnet-base-cased} & 0.349  $\pm$ 0.046 & 0.136 $\pm$ 0.064 \\
				\texttt{electra-base-discriminator} & 0.306 $\pm$ 0.060 & 0.227 $\pm$ 0.039 \\
				\texttt{gpt2} & 0.356 $\pm$ 0.018 &	0.142 $\pm$ 0.021 \\
				\texttt{bart-base} & 0.317 $\pm$ 0.067 & 0.171 $\pm$ 0.037 \\
				\midrule
				\multicolumn{3}{l}{\bf \shortciteA{nayeem-chali-2017-extract}}\\
				\midrule
				$\lambda = $0 & 0.046 & 0.131 \\
				$\lambda = $0.3 & --0.193 & 0.160 \\
				$\lambda = $0.5 & --0.275 & 0.166 \\
				$\lambda = $0.7 & --0.312 & 0.156 \\
				$\lambda = $1.0 & --0.334 & 0.128 \\
				\bottomrule
			\end{tabular}
		\end{adjustbox}
	\end{center}
	\caption{Pearson correlation of all experimental results on inter-sentential coherence. NSP-Score is computed 5 times over data variant-5 (see \secref{model-exp}).}
	\label{tab:all_ic}
\end{table*}

\newpage
\section*{Appendix E. Pre-trained Language Model Scores in different layers over Faithfulness, Focus, and Coverage}

\begin{figure}[ht]
	\subfloat[]{\includegraphics[width = 0.33\linewidth]{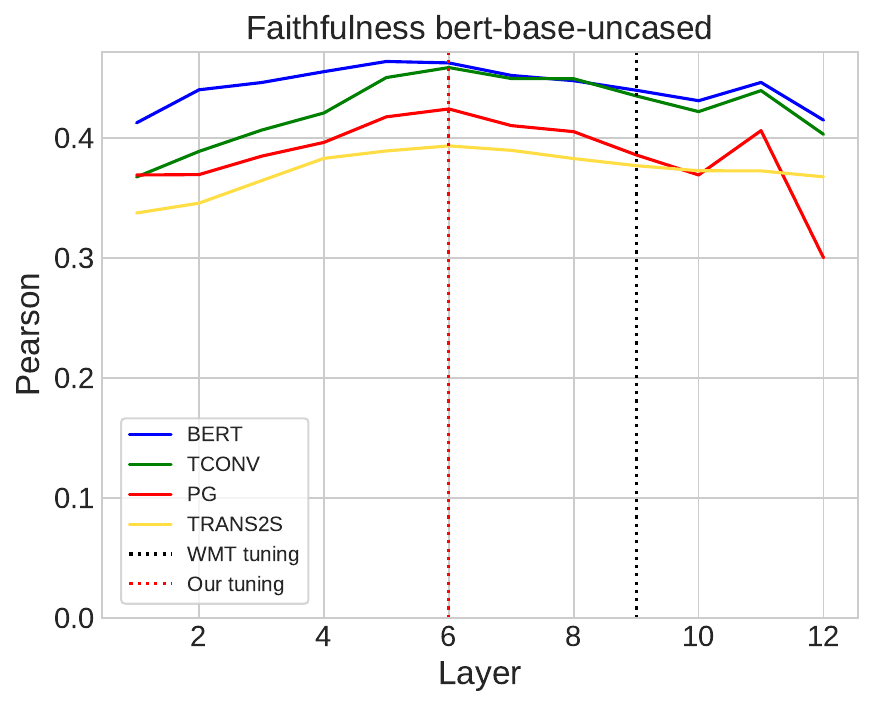}}
	\subfloat[]{\includegraphics[width = 0.33\linewidth]{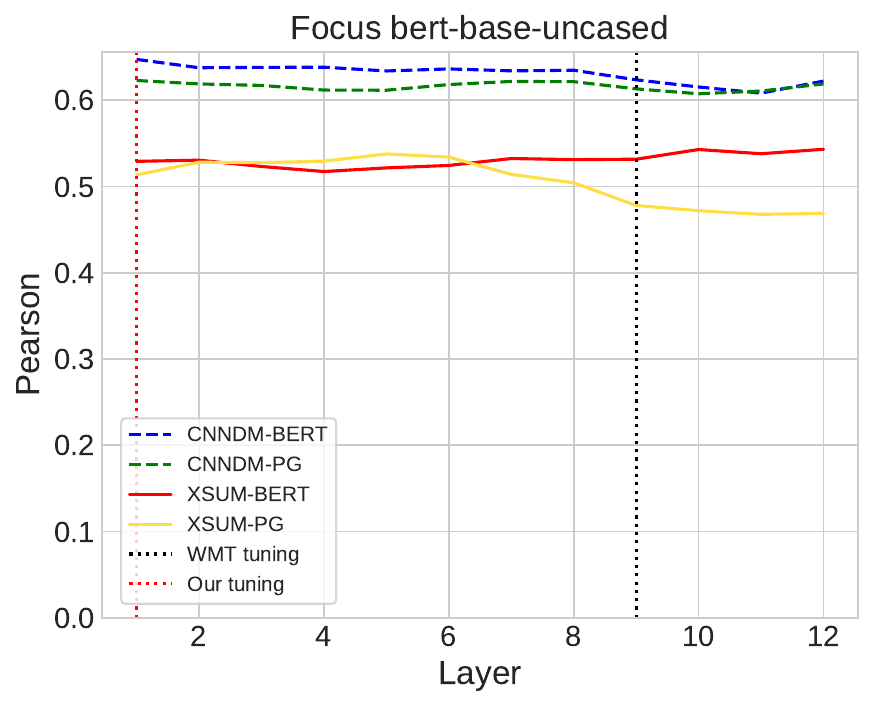}}
	\subfloat[]{\includegraphics[width = 0.33\linewidth]{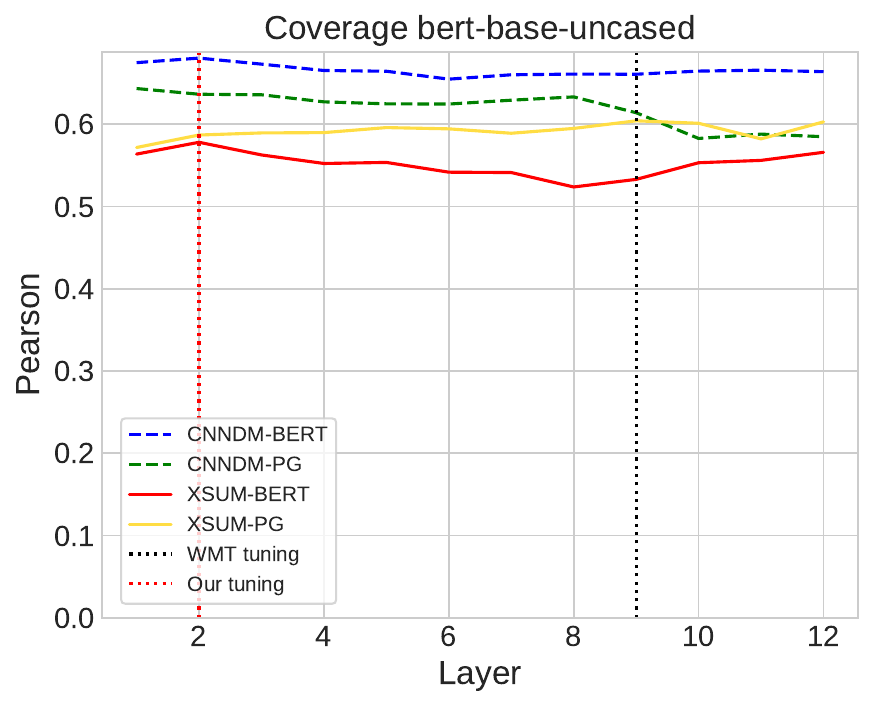}}\\
	\caption{\texttt{bert-base-uncased}}
	\label{fig:bert-base-uncased}
\end{figure}

\begin{figure}[ht]
	\subfloat[]{\includegraphics[width = 0.33\linewidth]{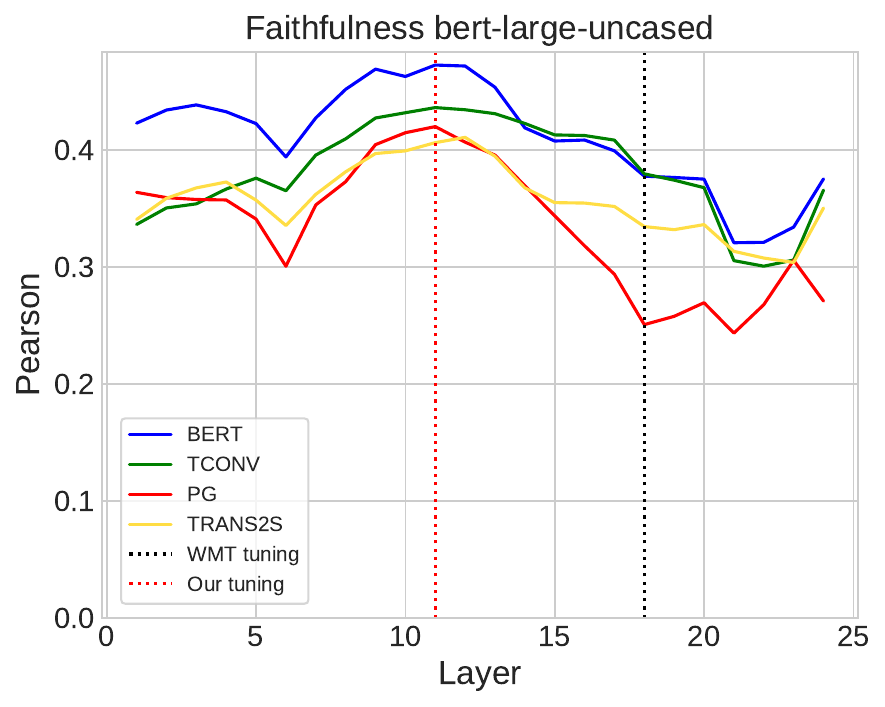}}
	\subfloat[]{\includegraphics[width = 0.33\linewidth]{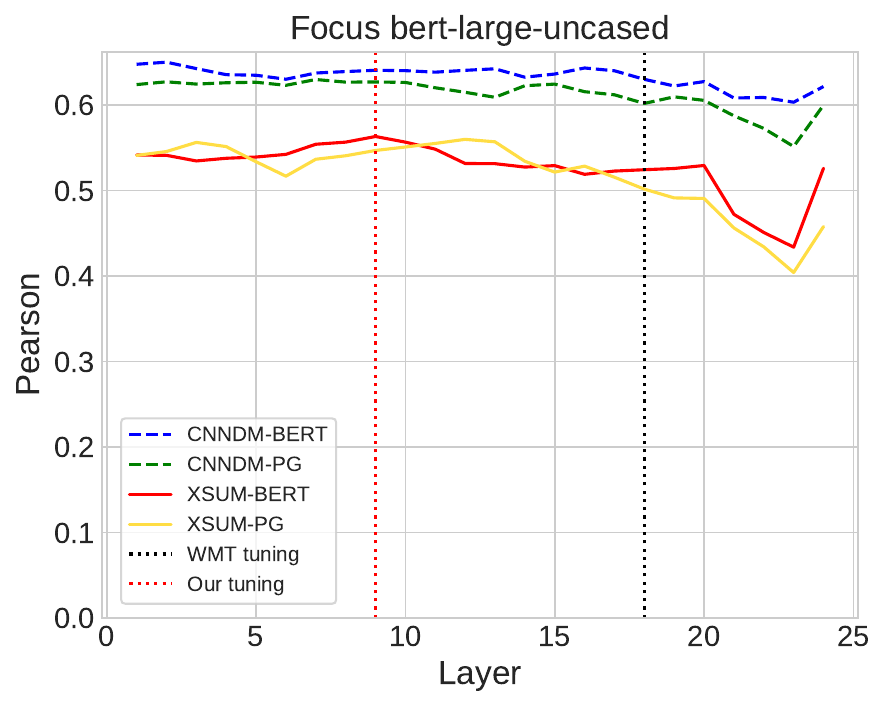}}
	\subfloat[]{\includegraphics[width = 0.33\linewidth]{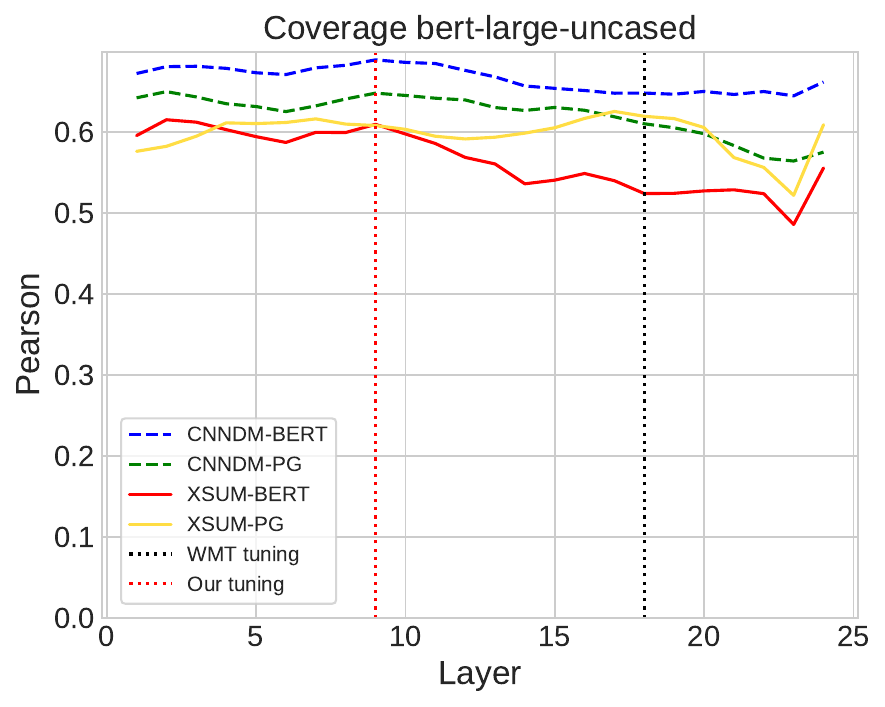}}\\
	\caption{\texttt{bert-large-uncased}}
	\label{fig:bert-large-uncased}
\end{figure}

\begin{figure}[ht]
	\subfloat[]{\includegraphics[width = 0.33\linewidth]{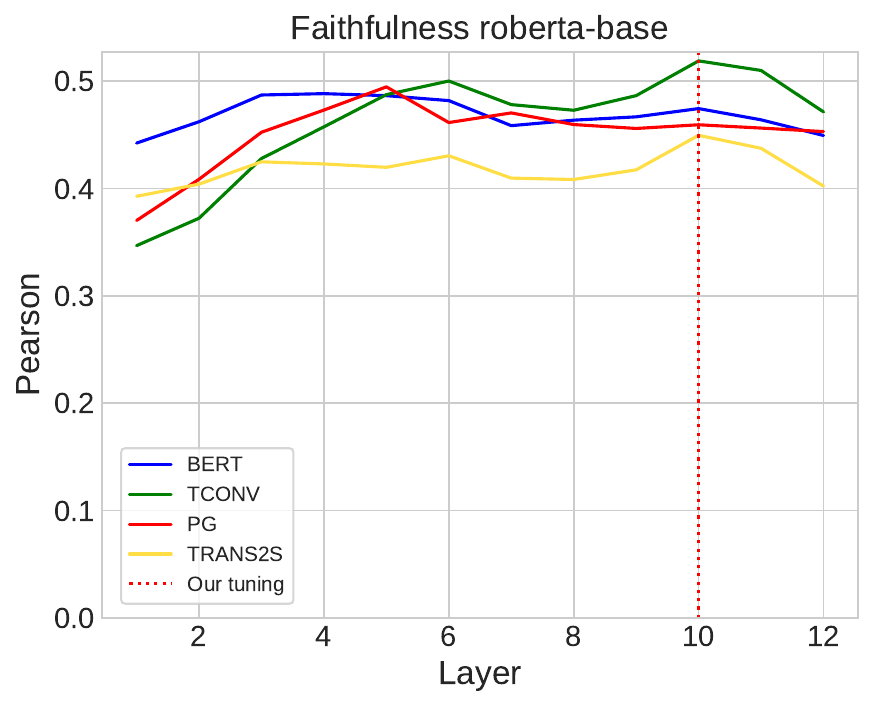}}
	\subfloat[]{\includegraphics[width = 0.33\linewidth]{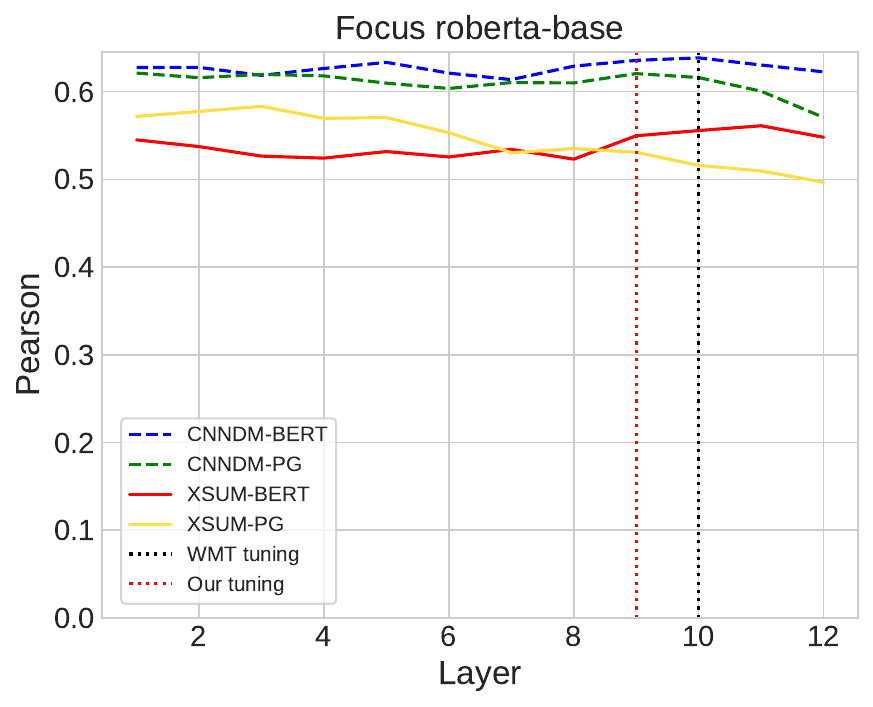}}
	\subfloat[]{\includegraphics[width = 0.33\linewidth]{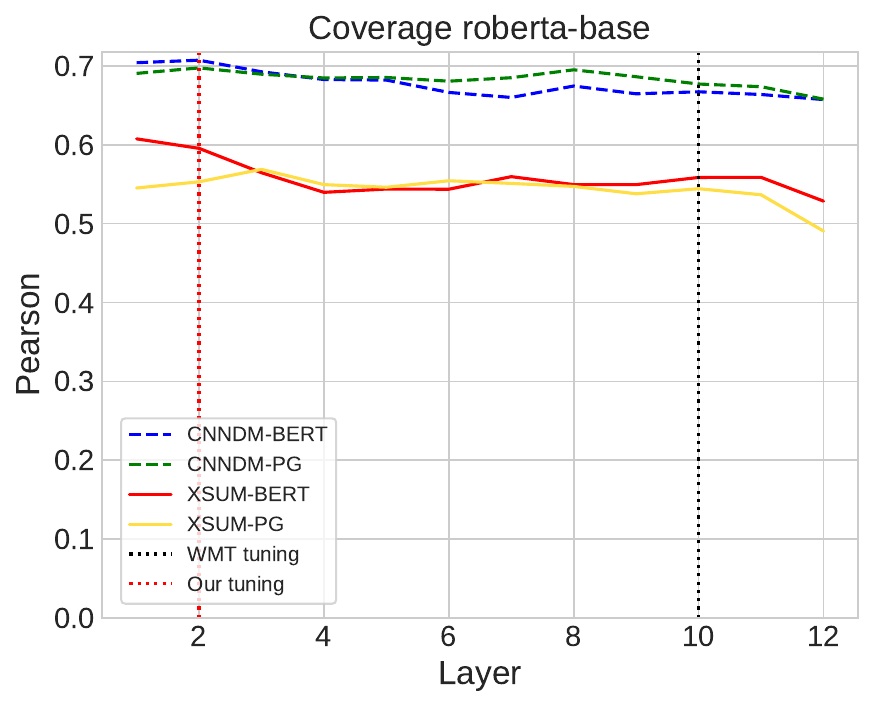}}\\
	\caption{\texttt{roberta-base}}
	\label{fig:roberta-base}
\end{figure}

\begin{figure}[ht]
	\subfloat[]{\includegraphics[width = 0.33\linewidth]{image/Faithfulness/roberta-large.pdf}}
	\subfloat[]{\includegraphics[width = 0.33\linewidth]{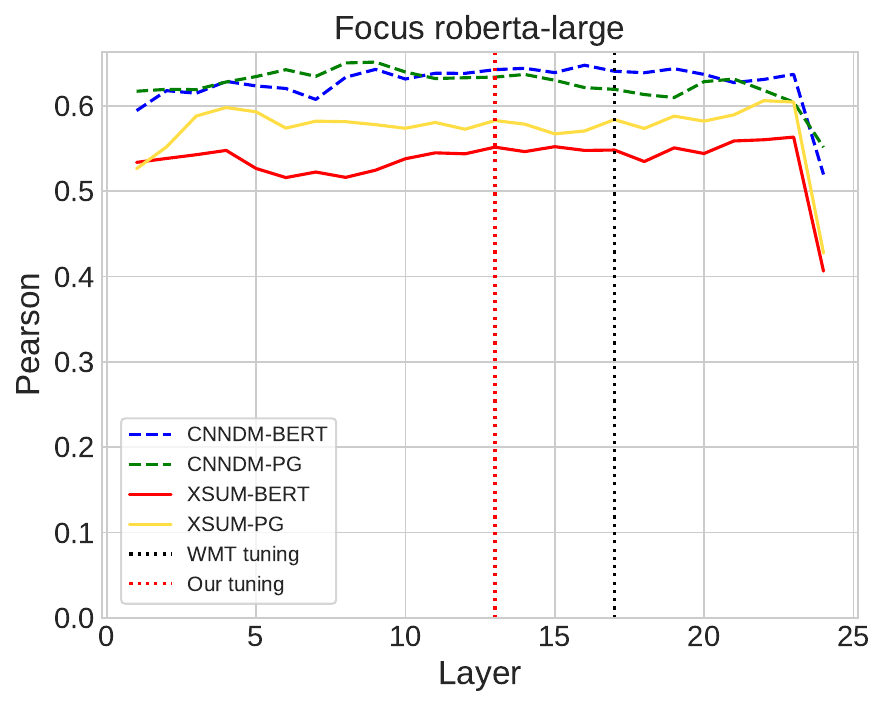}}
	\subfloat[]{\includegraphics[width = 0.33\linewidth]{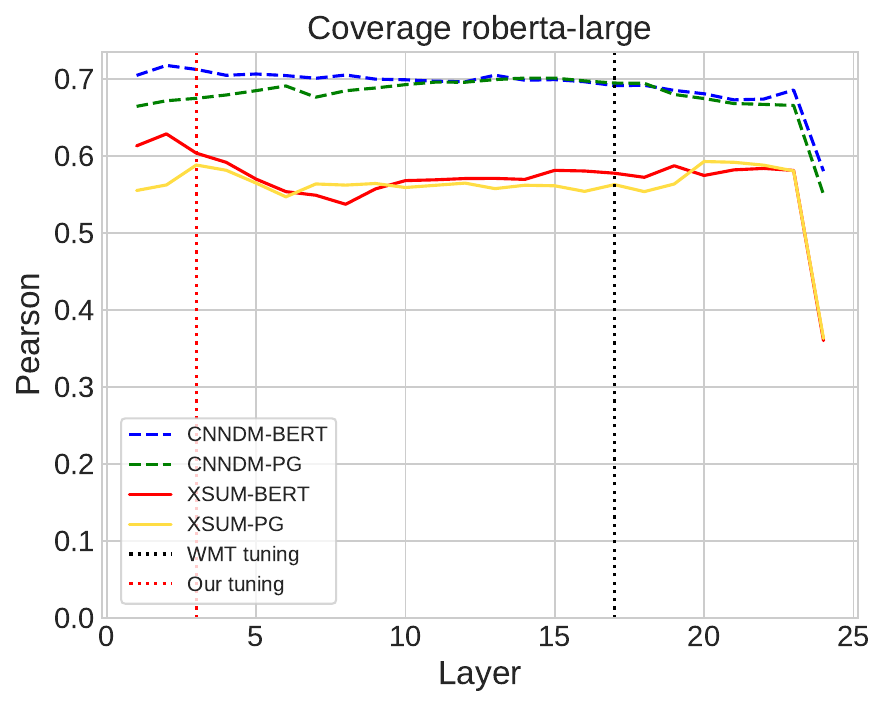}}\\
	\caption{\texttt{roberta-large}. Please note that final recommendation of \shortciteA{zhang2020bertscore} is to use layer-24, however, in their supplementary material (Appendix B), the best layer is 17. }
	\label{fig:roberta-large}
\end{figure}

\begin{figure}[ht]
	\subfloat[]{\includegraphics[width = 0.33\linewidth]{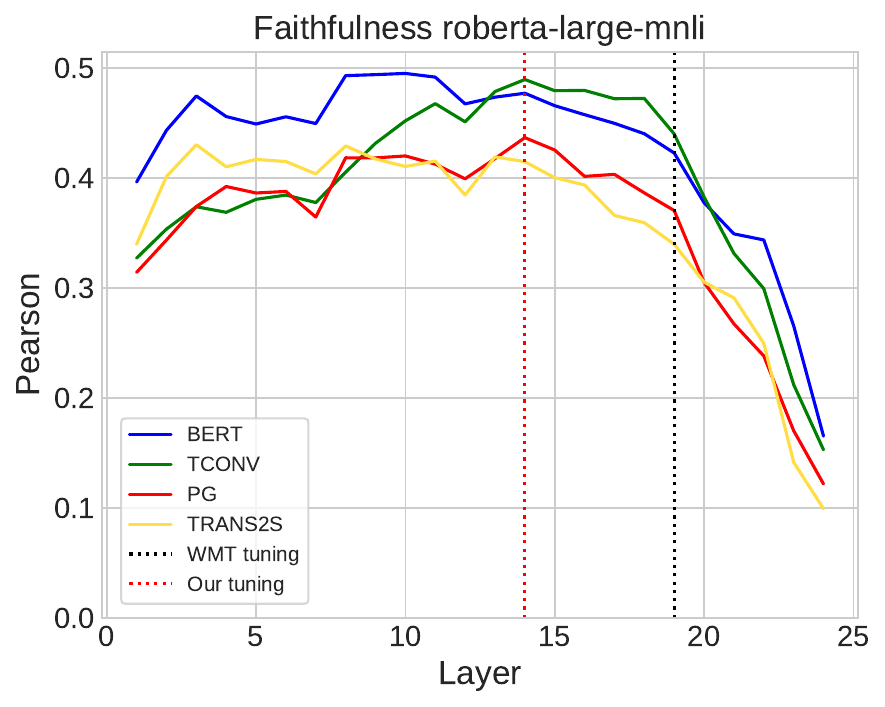}}
	\subfloat[]{\includegraphics[width = 0.33\linewidth]{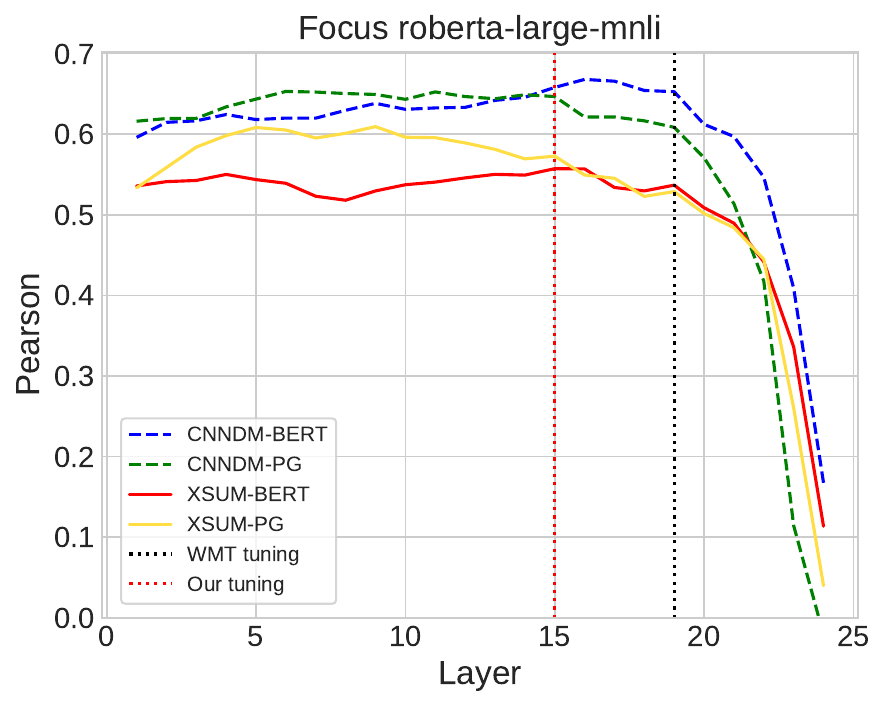}}
	\subfloat[]{\includegraphics[width = 0.33\linewidth]{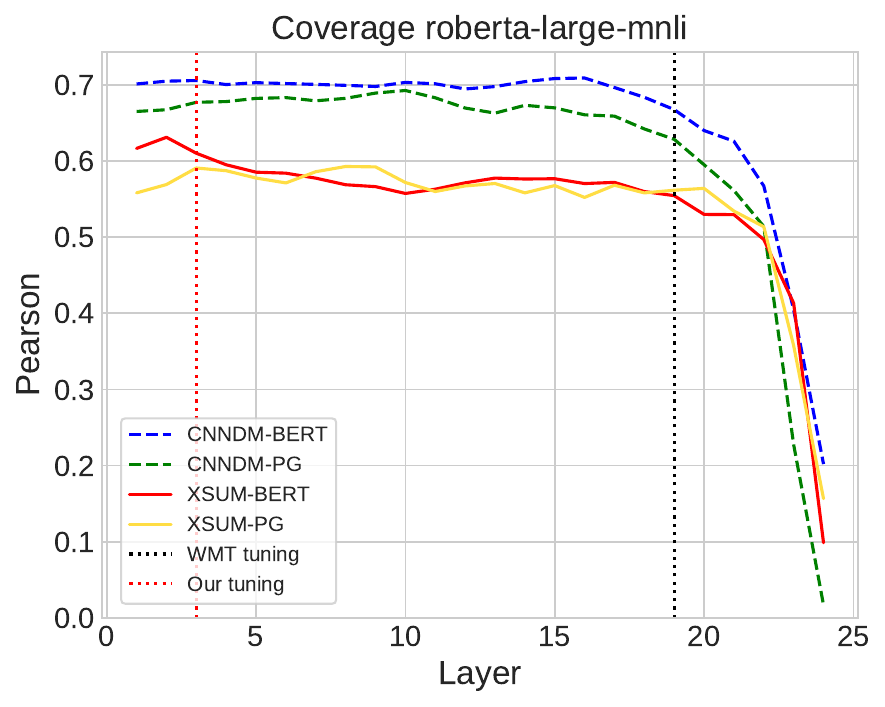}}\\
	\caption{\texttt{roberta-large-mnli}.}
	\label{fig:roberta-large-mnli}
\end{figure}

\begin{figure}[ht]
	\subfloat[]{\includegraphics[width = 0.33\linewidth]{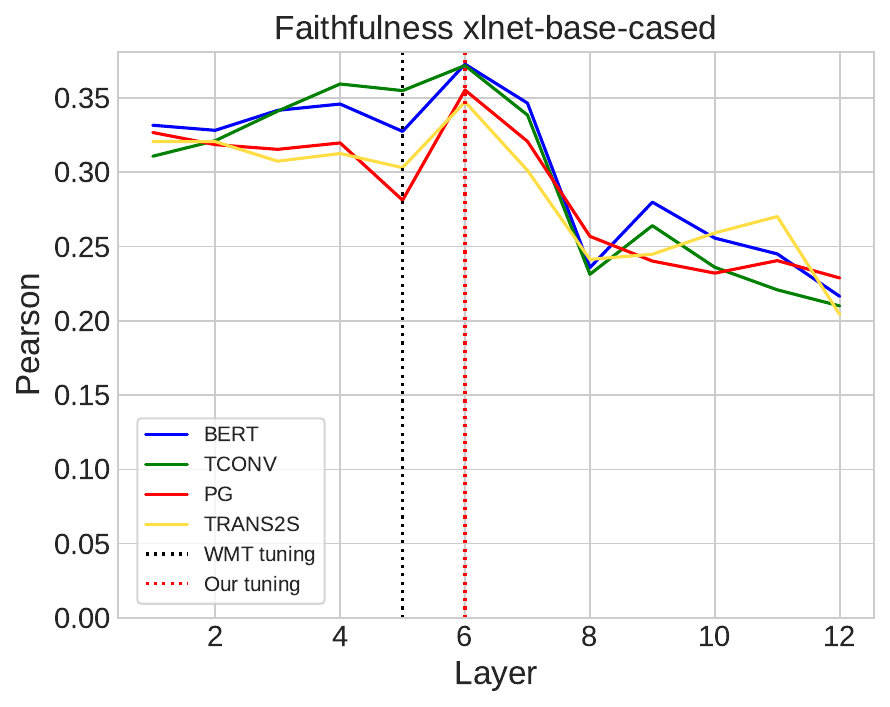}}
	\subfloat[]{\includegraphics[width = 0.33\linewidth]{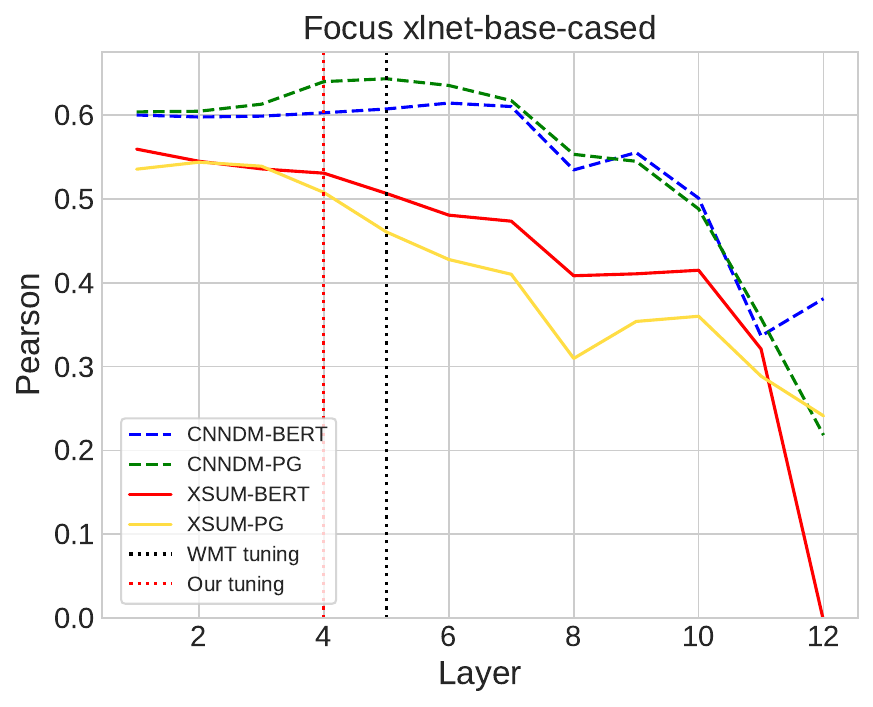}}
	\subfloat[]{\includegraphics[width = 0.33\linewidth]{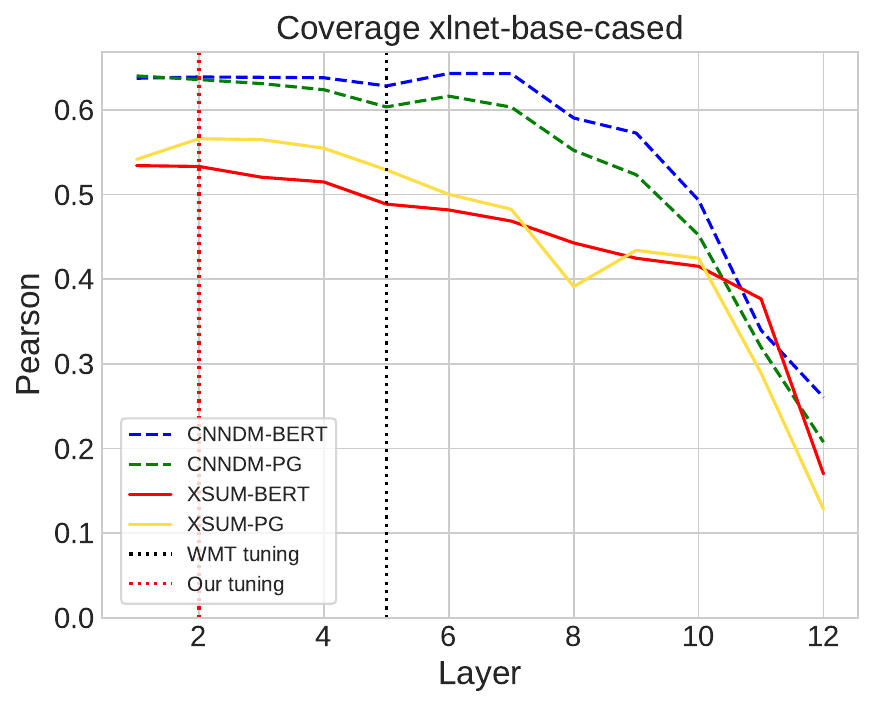}}\\
	\caption{\texttt{xlnet-base-cased}.}
	\label{fig:xlnet-base-cased}
\end{figure}

\begin{figure}[ht]
	\subfloat[]{\includegraphics[width = 0.33\linewidth]{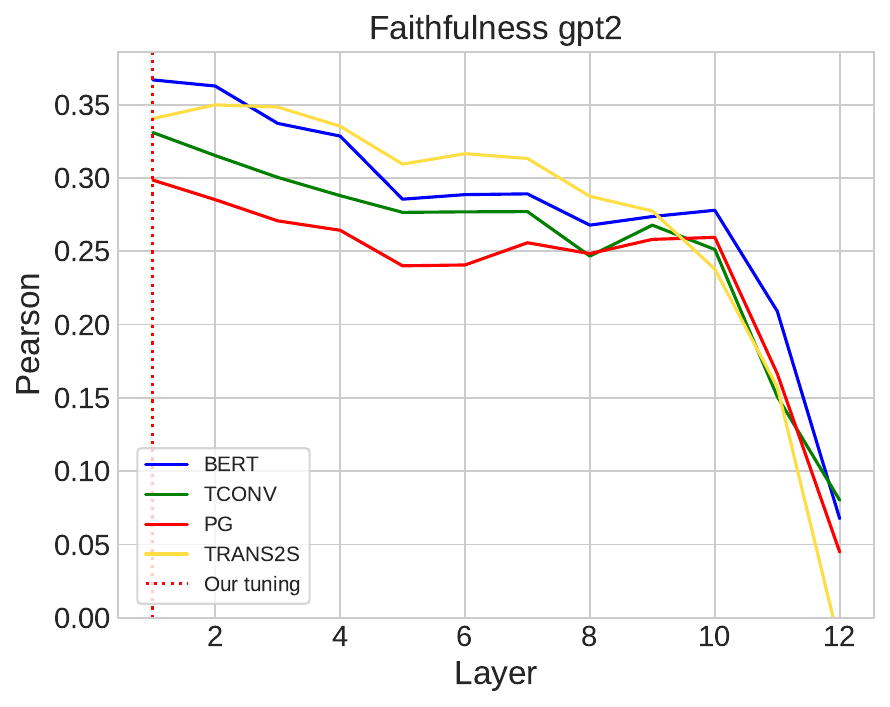}}
	\subfloat[]{\includegraphics[width = 0.33\linewidth]{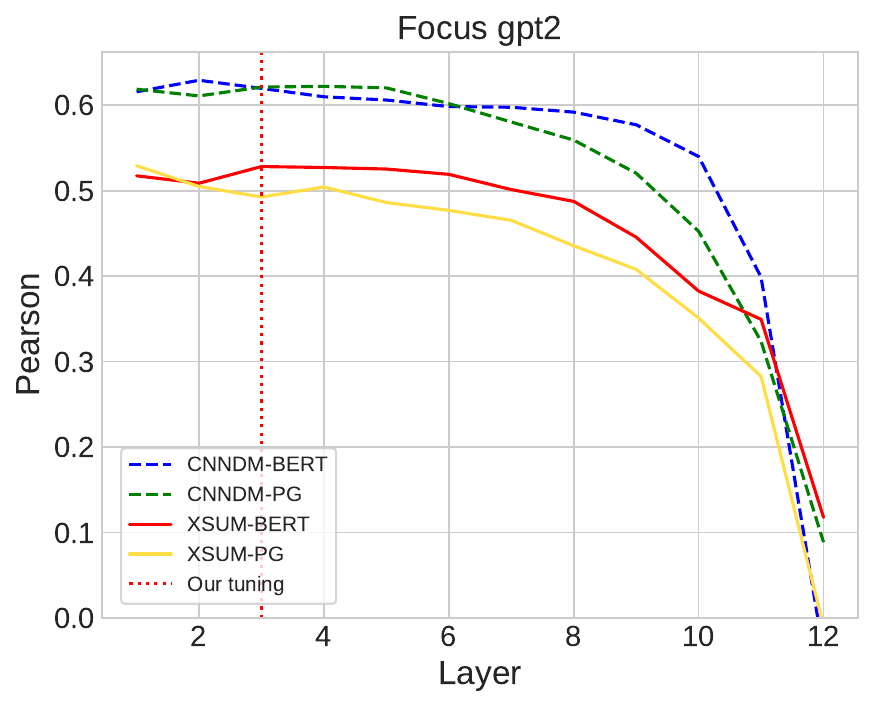}}
	\subfloat[]{\includegraphics[width = 0.33\linewidth]{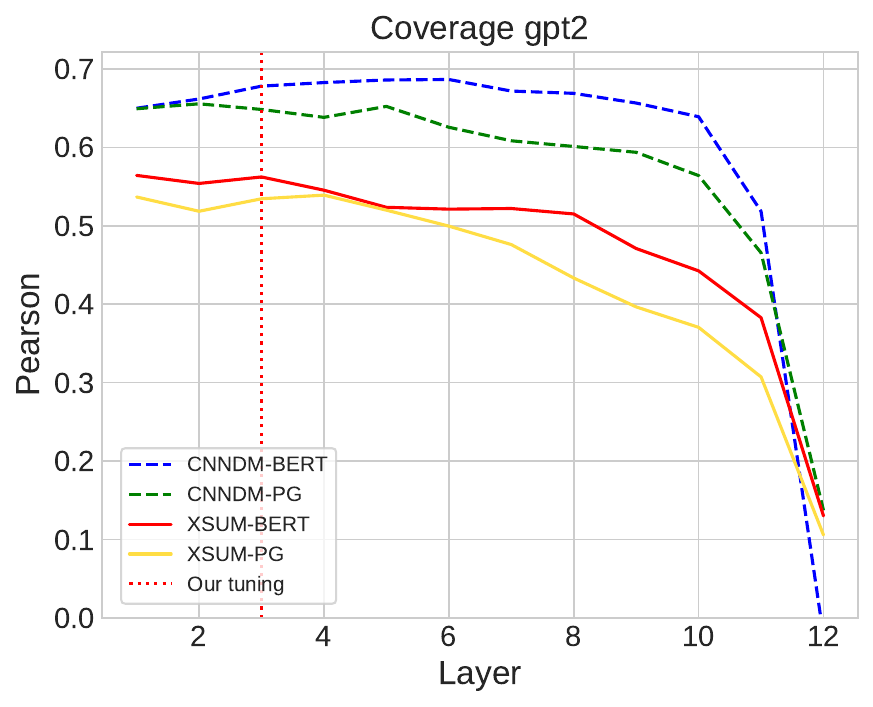}}\\
	\caption{\texttt{gpt2}.}
	\label{fig:gpt2}
\end{figure}

\begin{figure}[ht]
	\subfloat[]{\includegraphics[width = 0.33\linewidth]{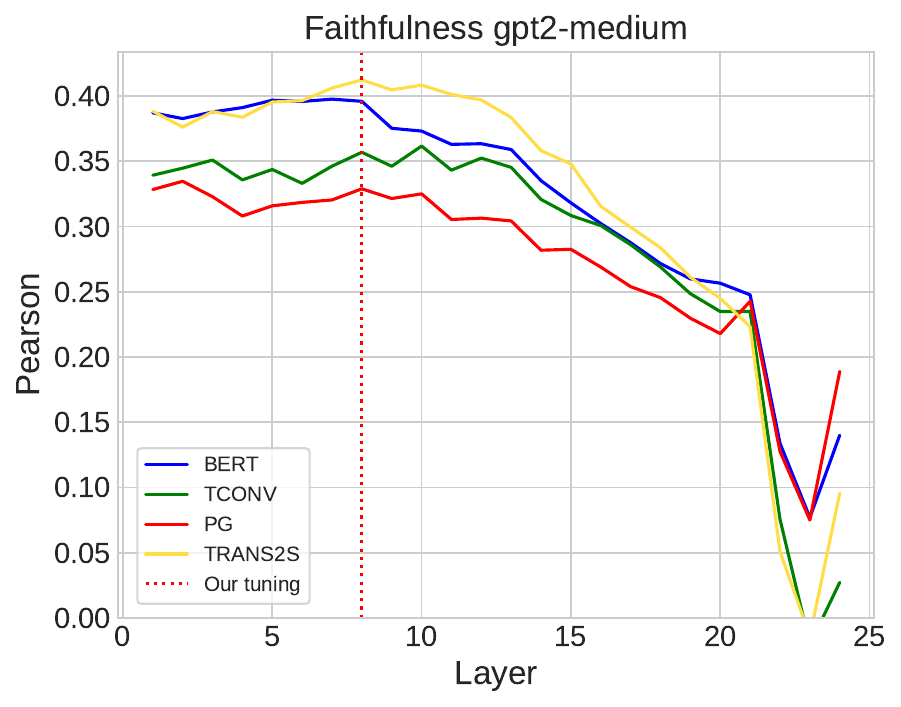}}
	\subfloat[]{\includegraphics[width = 0.33\linewidth]{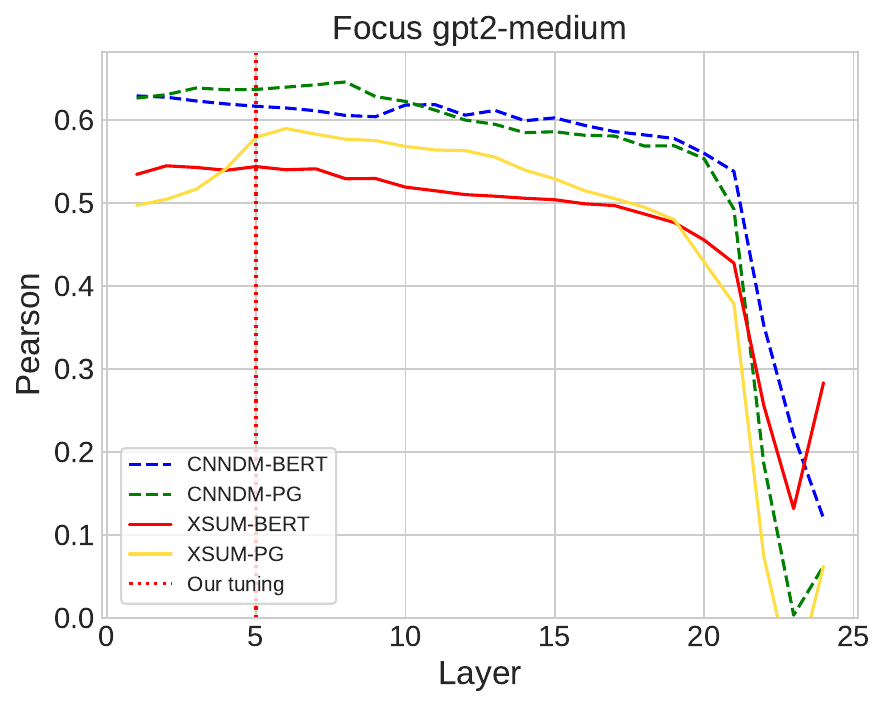}}
	\subfloat[]{\includegraphics[width = 0.33\linewidth]{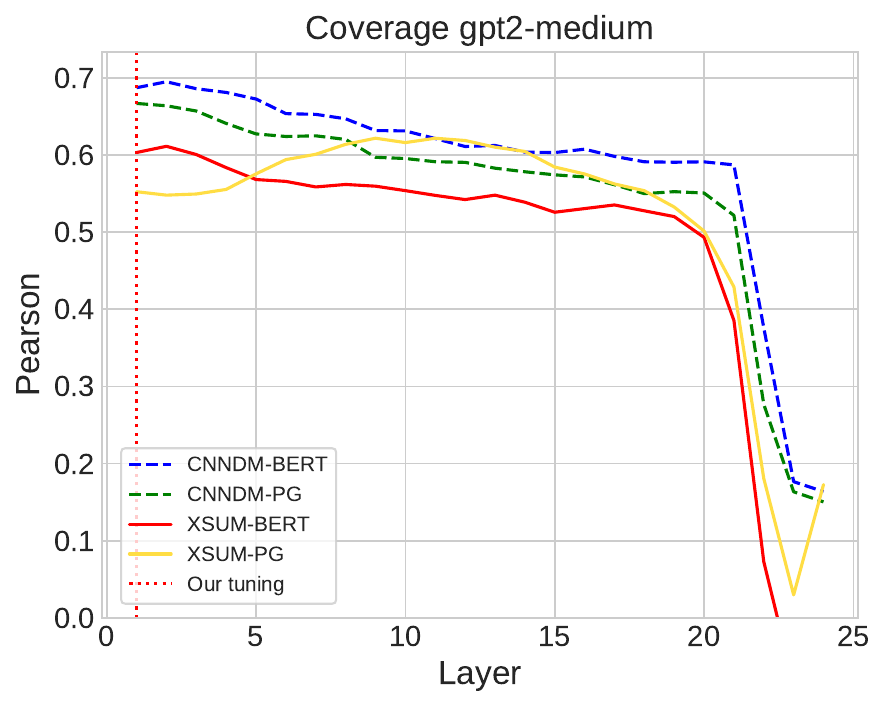}}\\
	\caption{\texttt{gpt2-medium}.}
	\label{fig:gpt2-medium}
\end{figure}

\begin{figure}[ht]
	\subfloat[]{\includegraphics[width = 0.33\linewidth]{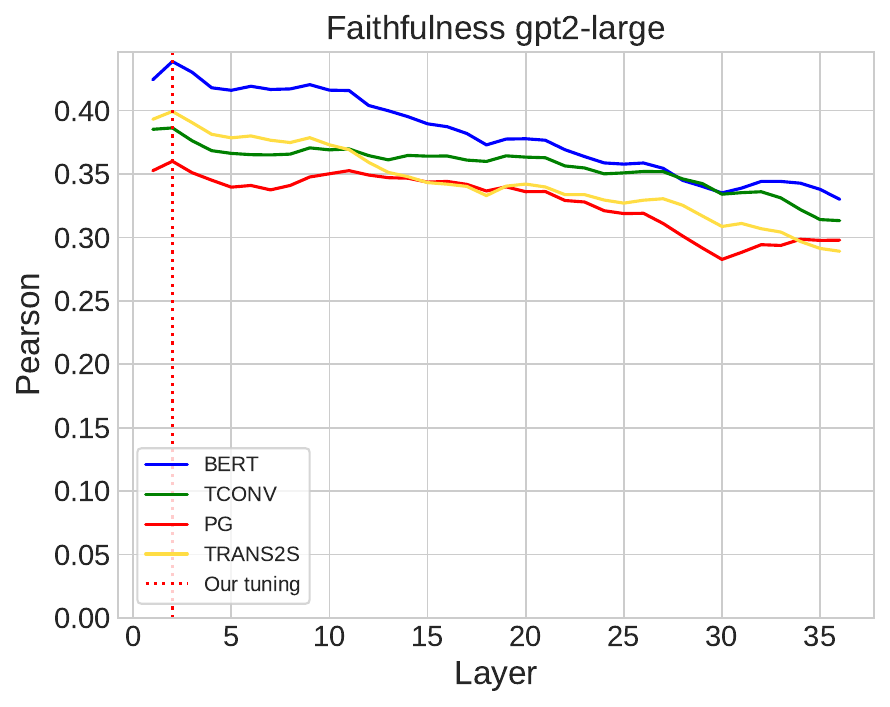}}
	\subfloat[]{\includegraphics[width = 0.33\linewidth]{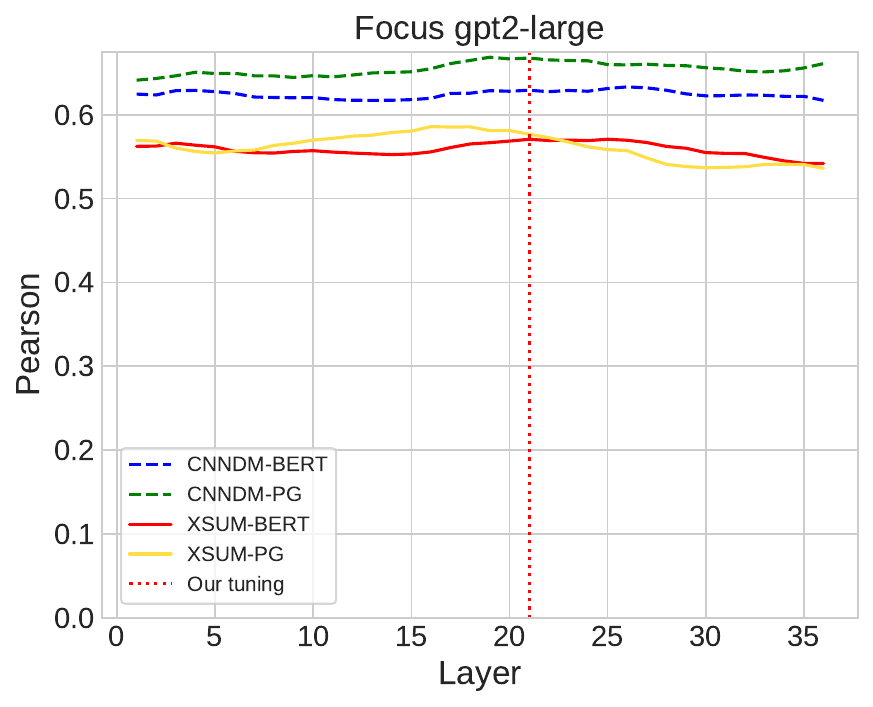}}
	\subfloat[]{\includegraphics[width = 0.33\linewidth]{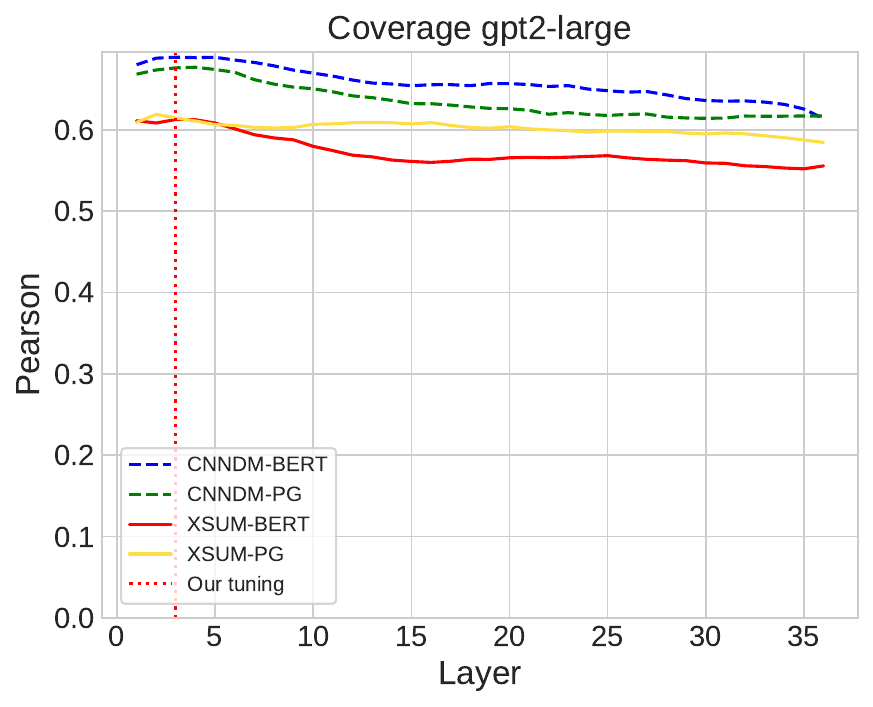}}\\
	\caption{\texttt{gpt2-large}.}
	\label{fig:gpt2-large}
\end{figure}

\begin{figure}[ht]
	\subfloat[]{\includegraphics[width = 0.33\linewidth]{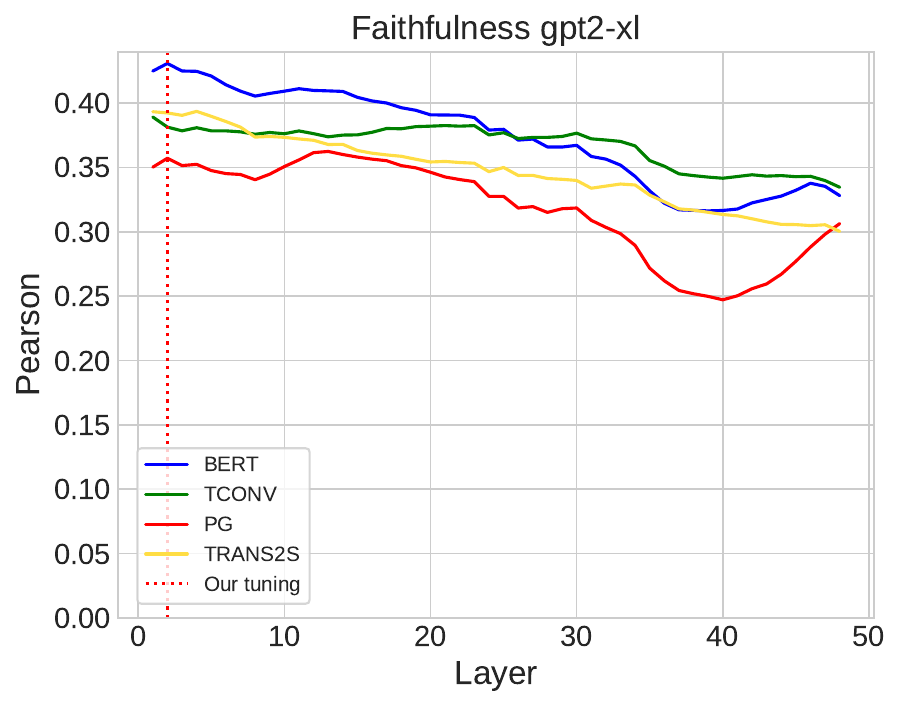}}
	\subfloat[]{\includegraphics[width = 0.33\linewidth]{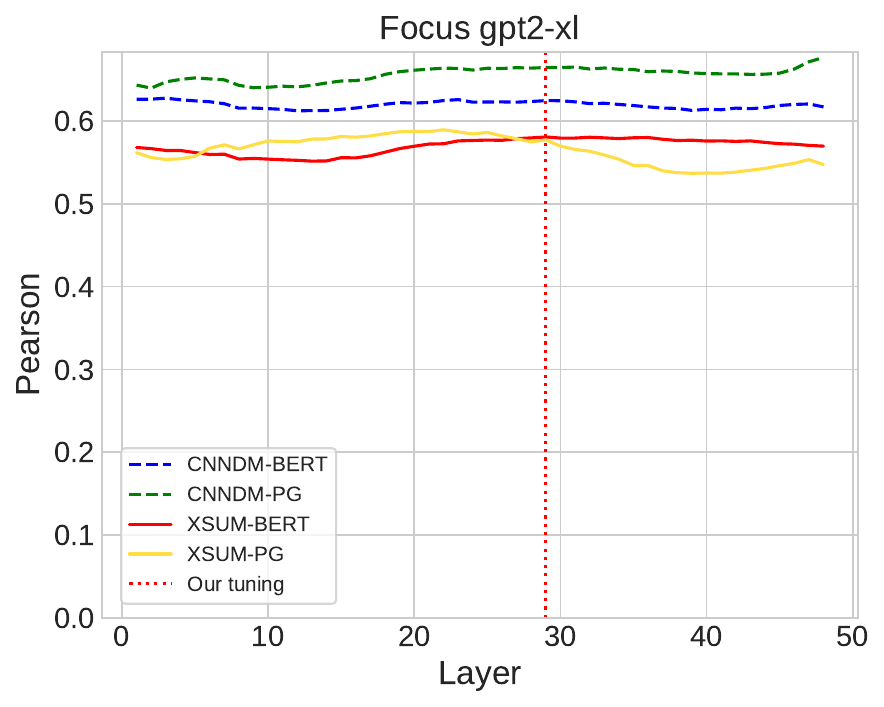}}
	\subfloat[]{\includegraphics[width = 0.33\linewidth]{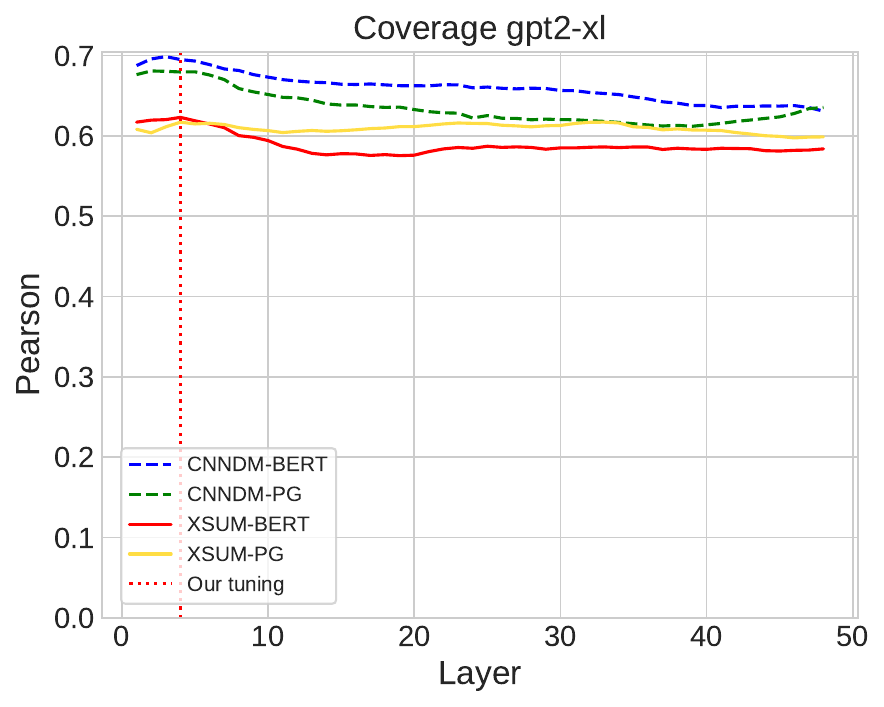}}\\
	\caption{\texttt{gpt2-xl}.}
	\label{fig:gpt2-xl}
\end{figure}

\begin{figure}[ht]
	\subfloat[]{\includegraphics[width = 0.33\linewidth]{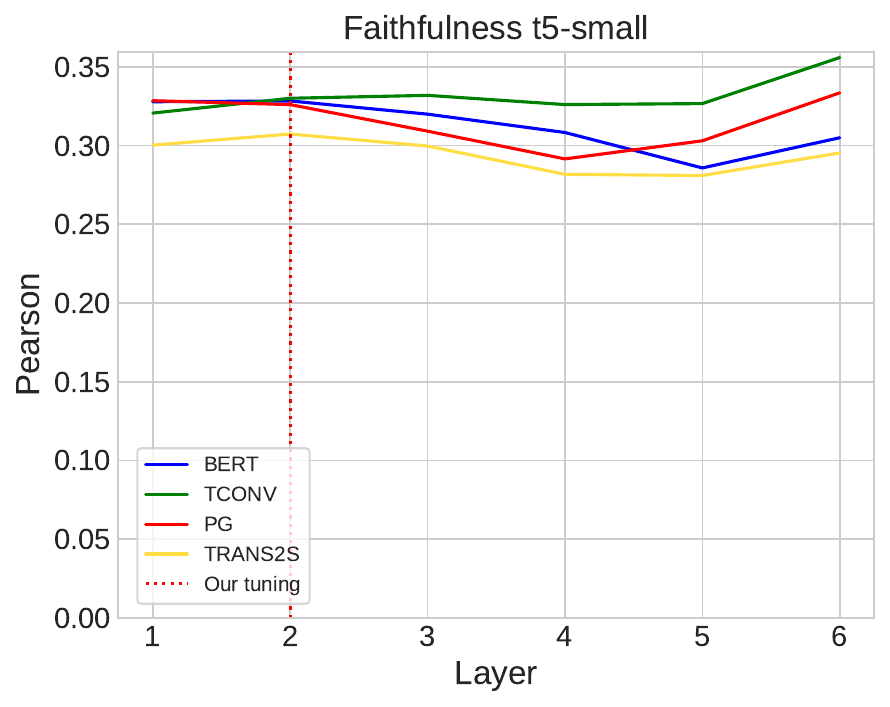}}
	\subfloat[]{\includegraphics[width = 0.33\linewidth]{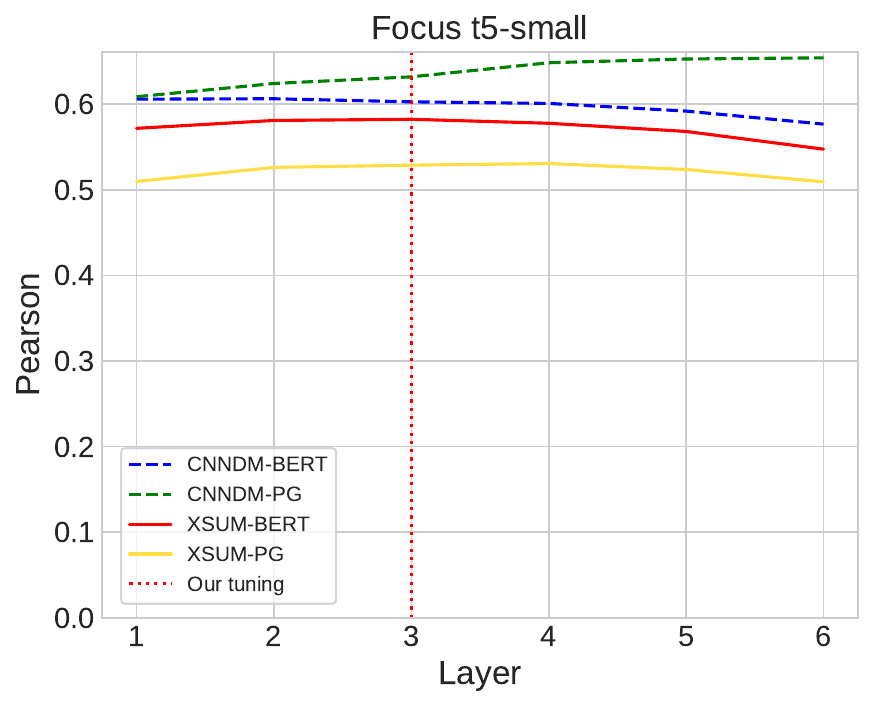}}
	\subfloat[]{\includegraphics[width = 0.33\linewidth]{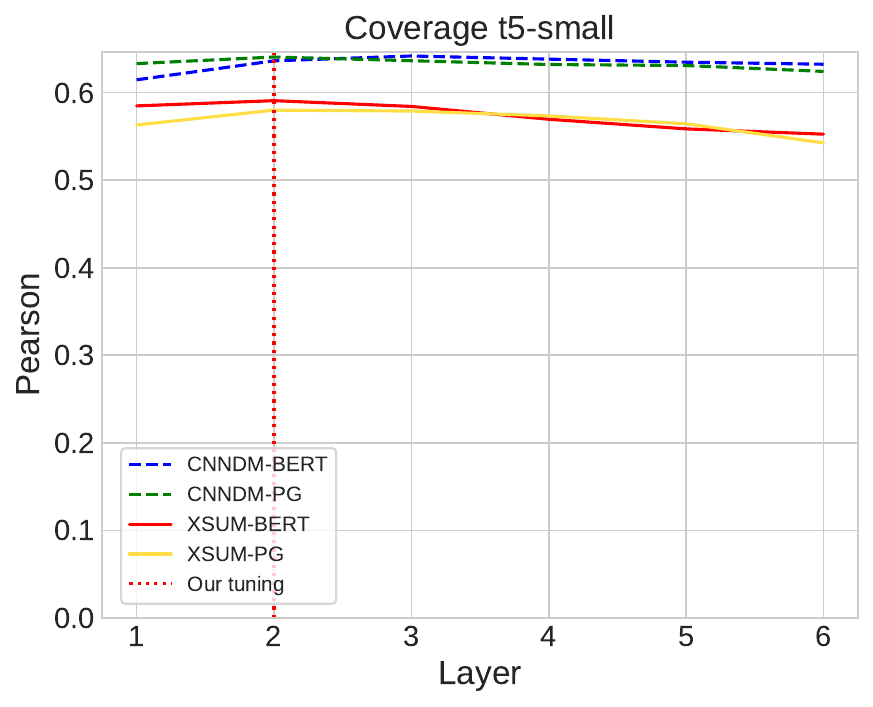}}\\
	\caption{\texttt{t5-small}.}
	\label{fig:t5-small}
\end{figure}

\begin{figure}[ht]
	\subfloat[]{\includegraphics[width = 0.33\linewidth]{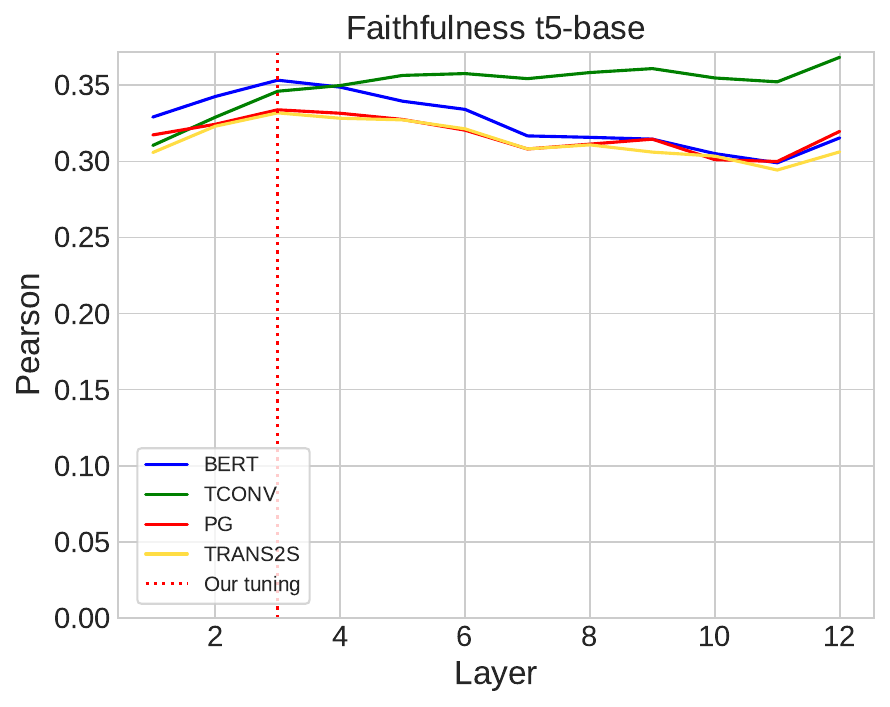}}
	\subfloat[]{\includegraphics[width = 0.33\linewidth]{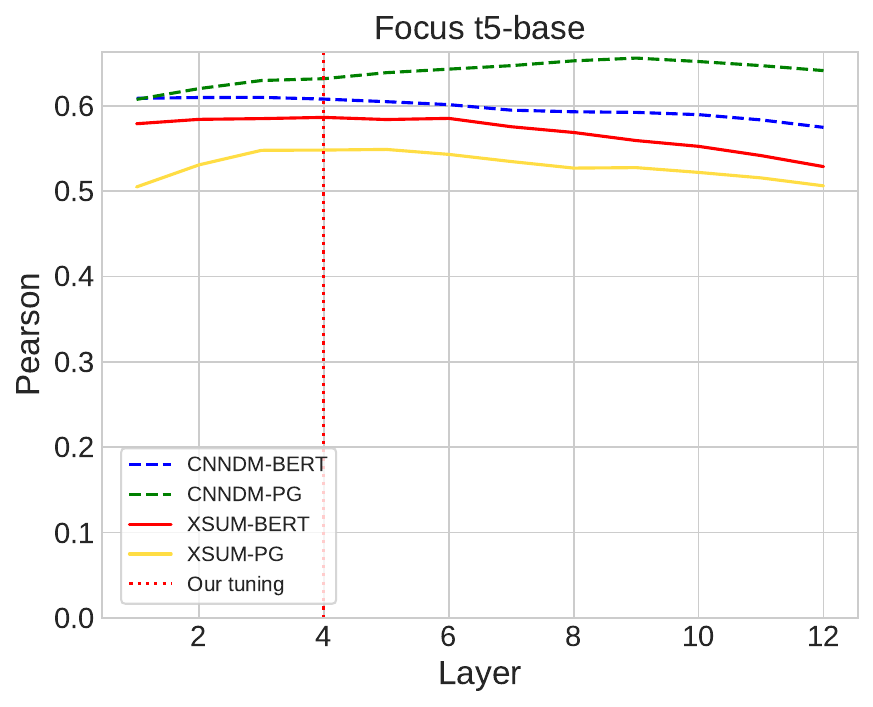}}
	\subfloat[]{\includegraphics[width = 0.33\linewidth]{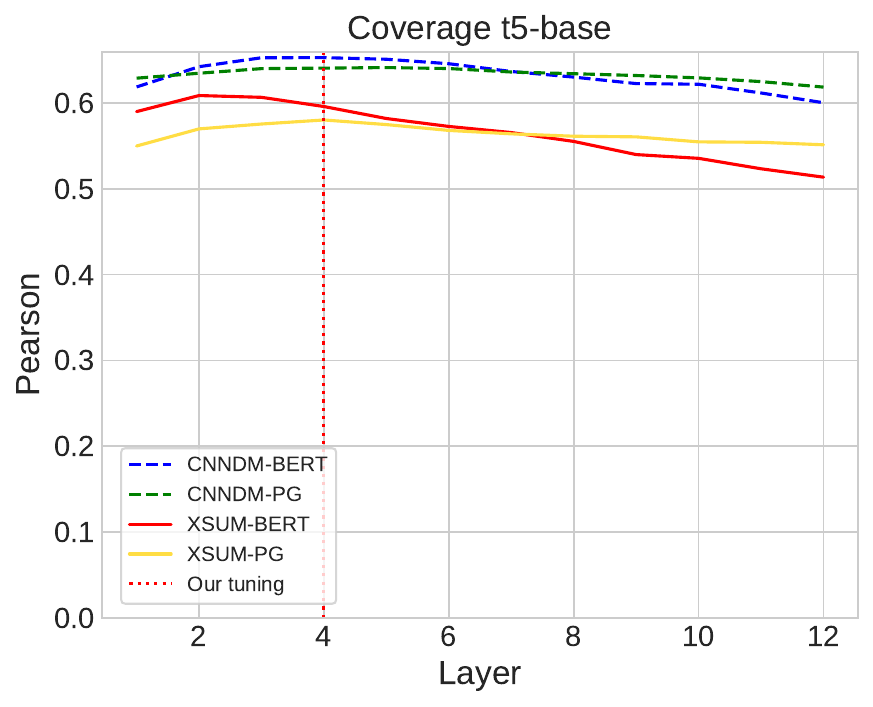}}\\
	\caption{\texttt{t5-base}.}
	\label{fig:t5-base}
\end{figure}

\begin{figure}[ht]
	\subfloat[]{\includegraphics[width = 0.33\linewidth]{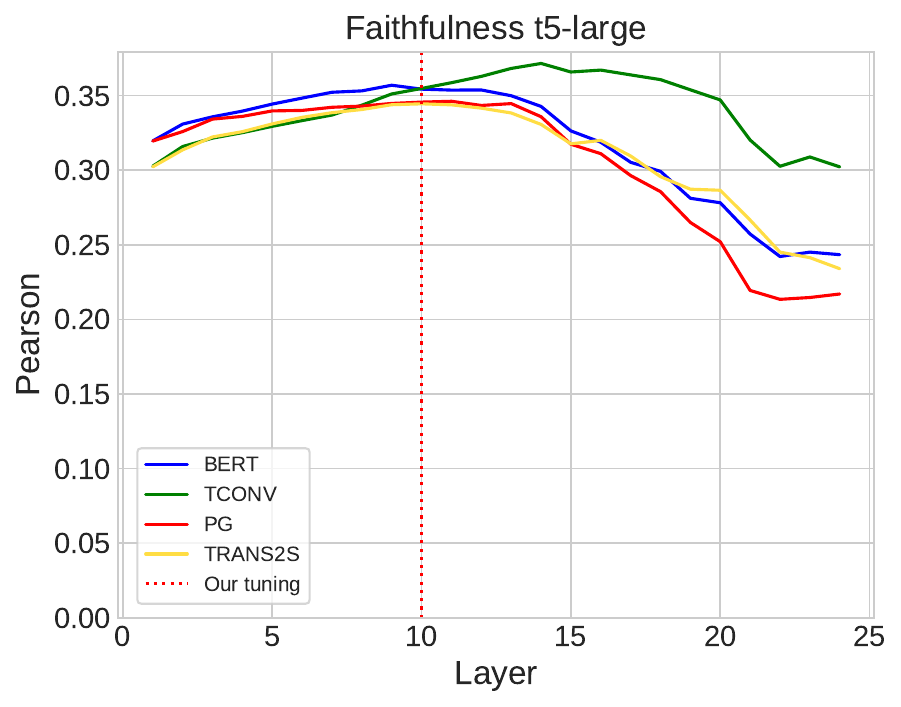}}
	\subfloat[]{\includegraphics[width = 0.33\linewidth]{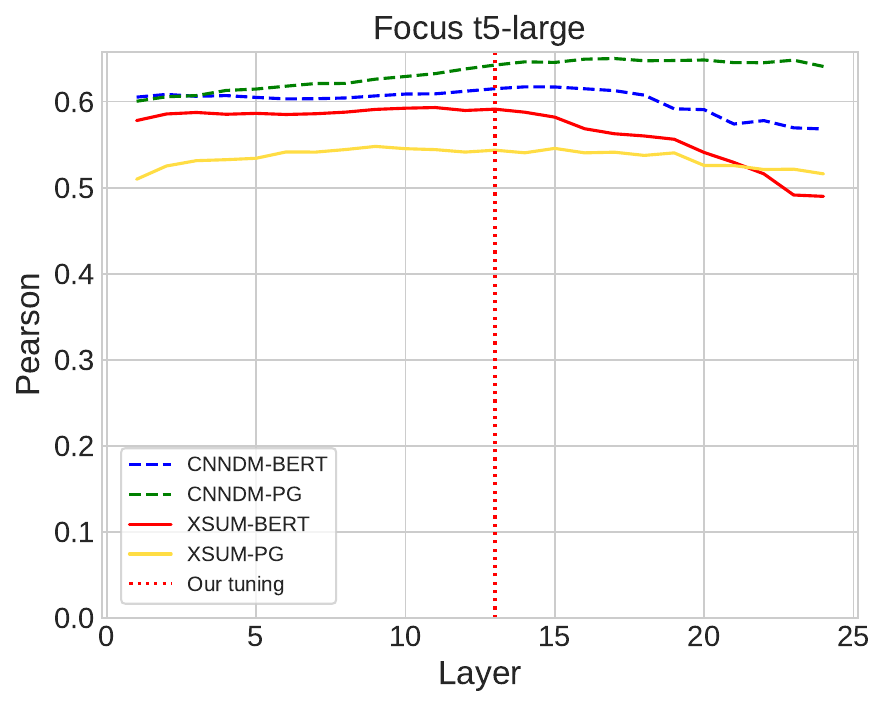}}
	\subfloat[]{\includegraphics[width = 0.33\linewidth]{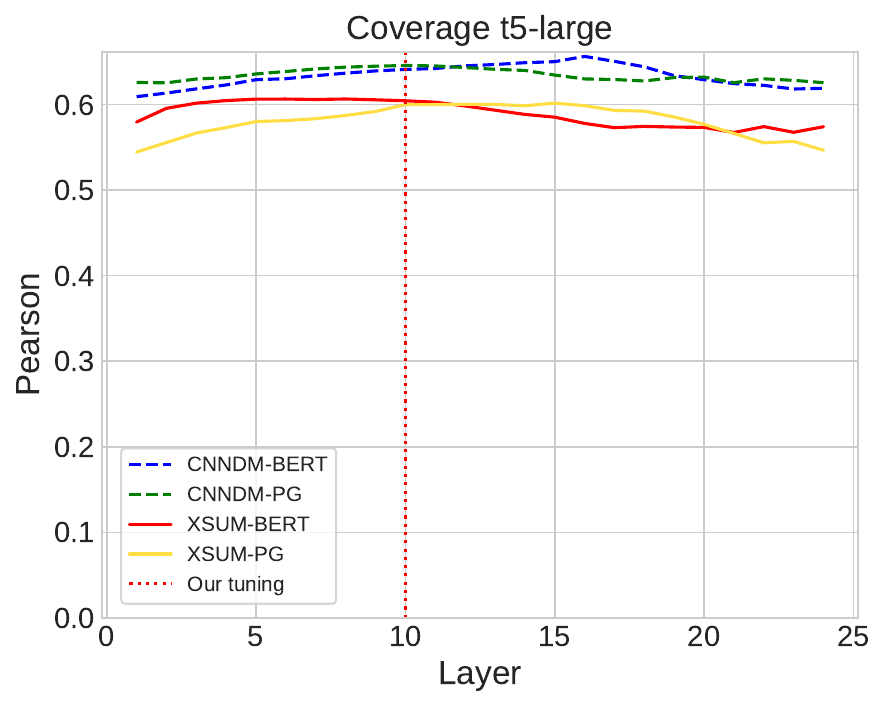}}\\
	\caption{\texttt{t5-large}.}
	\label{fig:t5-large}
\end{figure}

\begin{figure}[ht]
	\subfloat[]{\includegraphics[width = 0.33\linewidth]{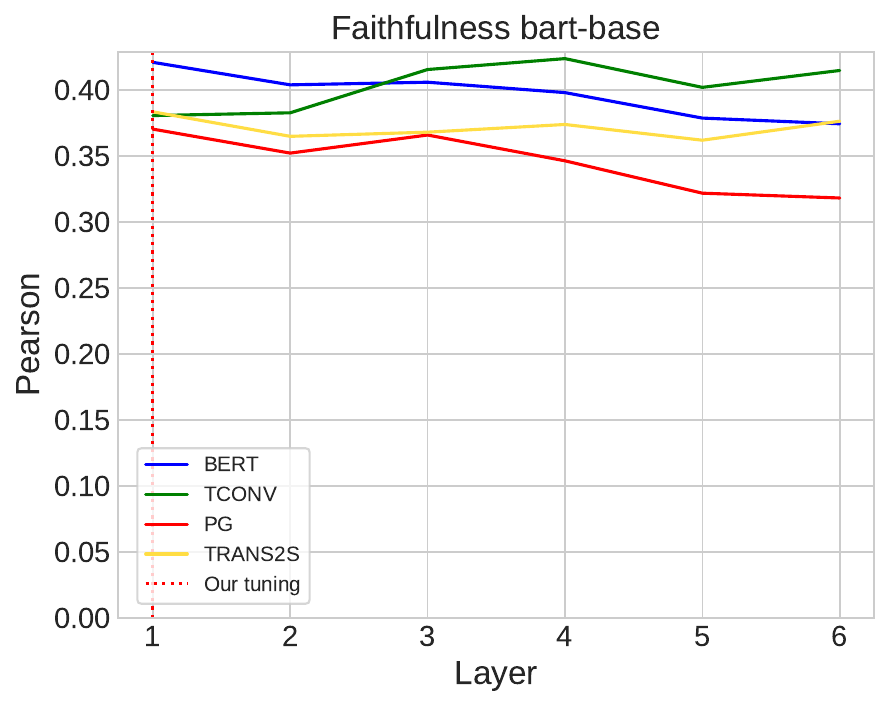}}
	\subfloat[]{\includegraphics[width = 0.33\linewidth]{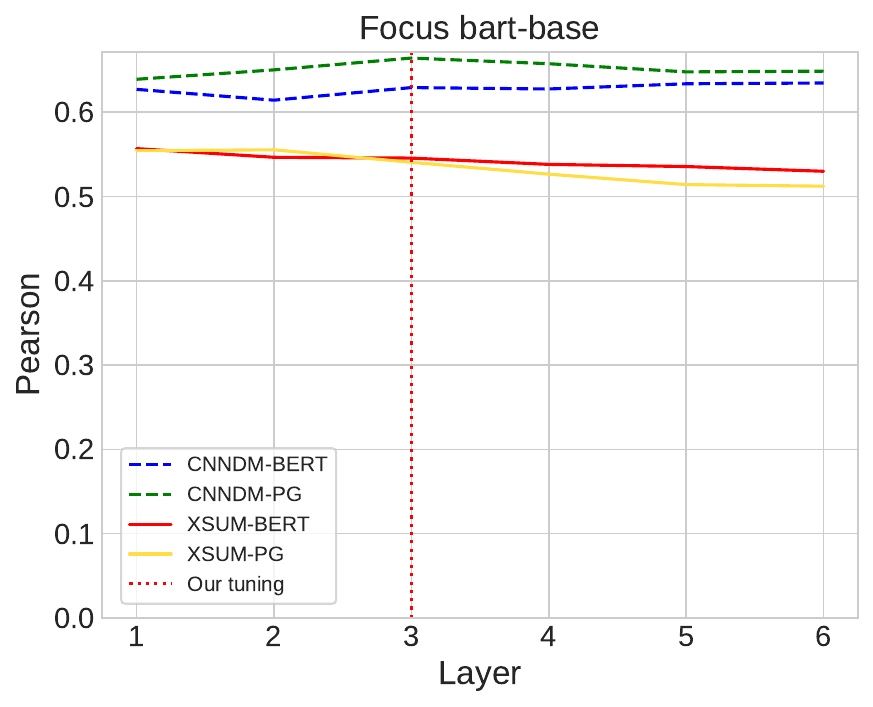}}
	\subfloat[]{\includegraphics[width = 0.33\linewidth]{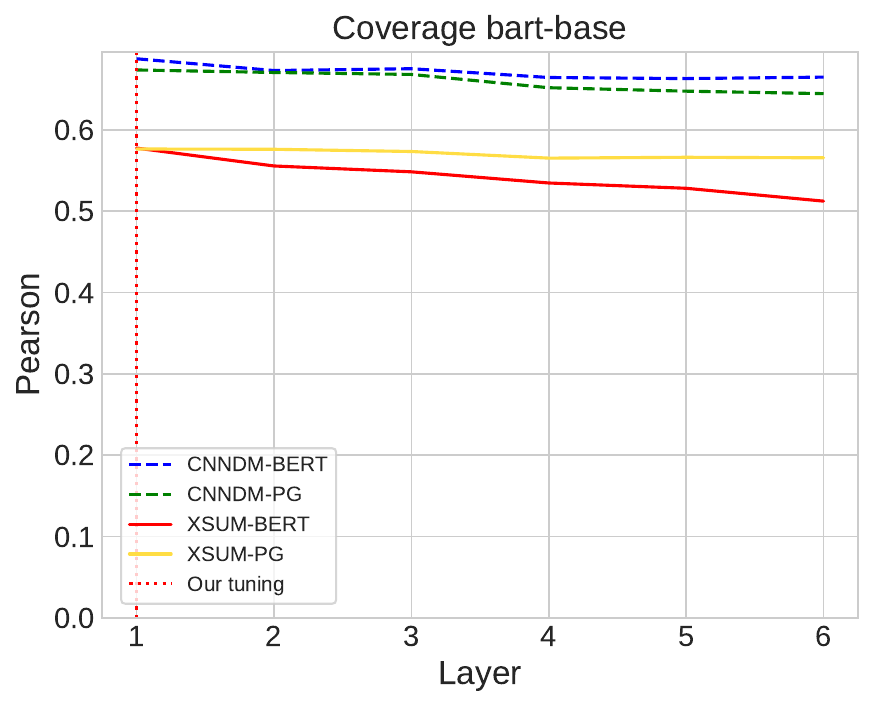}}\\
	\caption{\texttt{bart-base}.}
	\label{fig:bart-base}
\end{figure}

\begin{figure}[ht]
	\subfloat[]{\includegraphics[width = 0.33\linewidth]{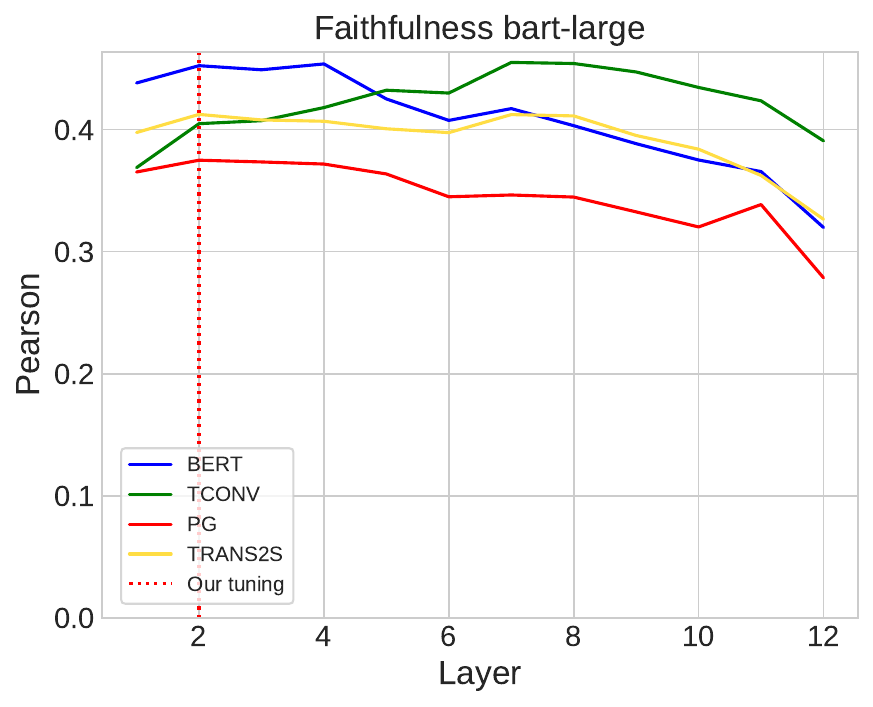}}
	\subfloat[]{\includegraphics[width = 0.33\linewidth]{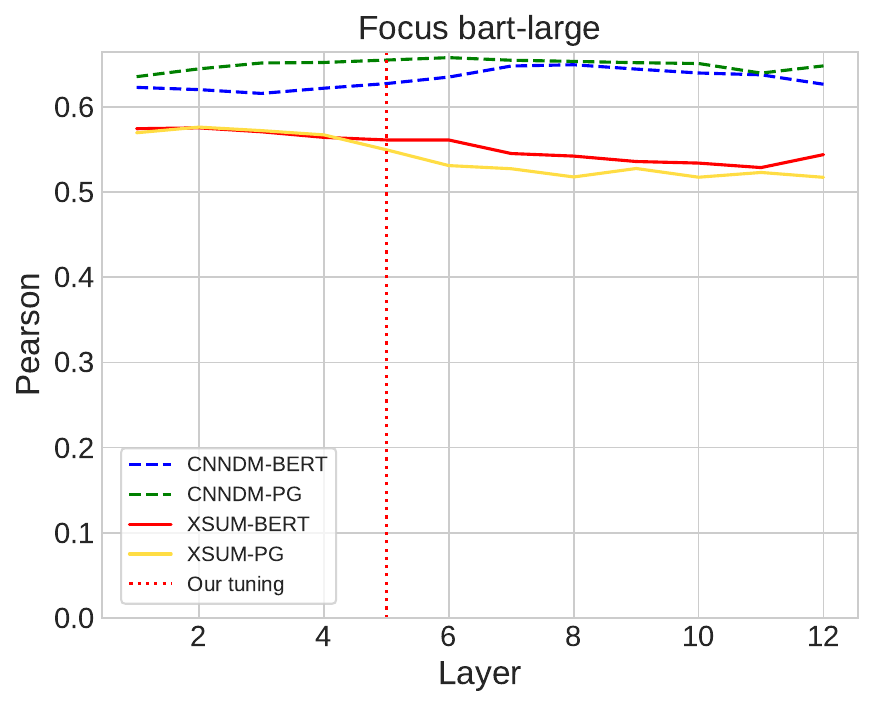}}
	\subfloat[]{\includegraphics[width = 0.33\linewidth]{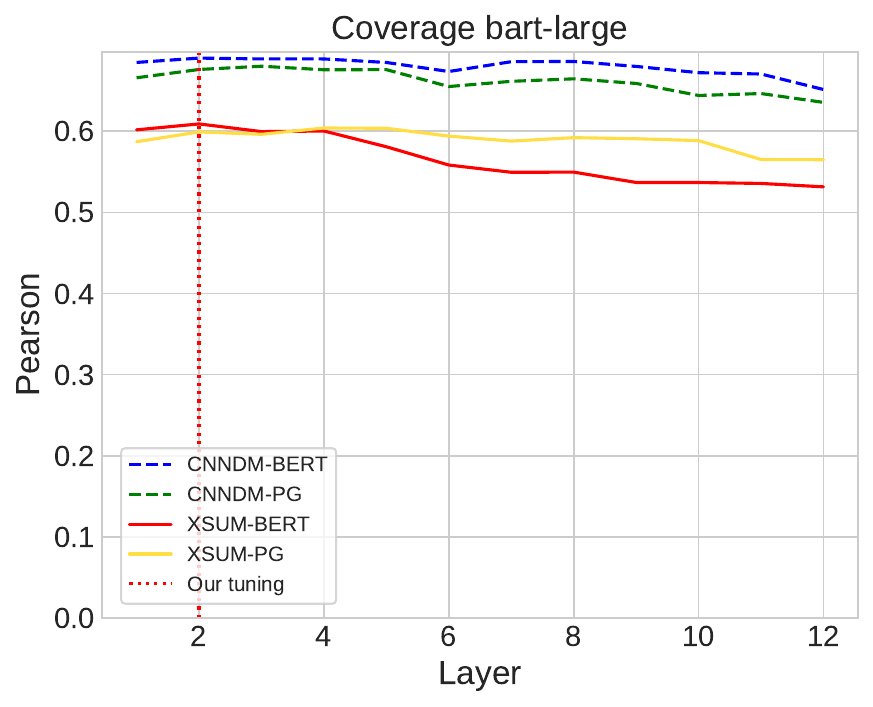}}\\
	\caption{\texttt{bart-large}.}
	\label{fig:bart-large}
\end{figure}

\begin{figure}[ht]
	\subfloat[]{\includegraphics[width = 0.33\linewidth]{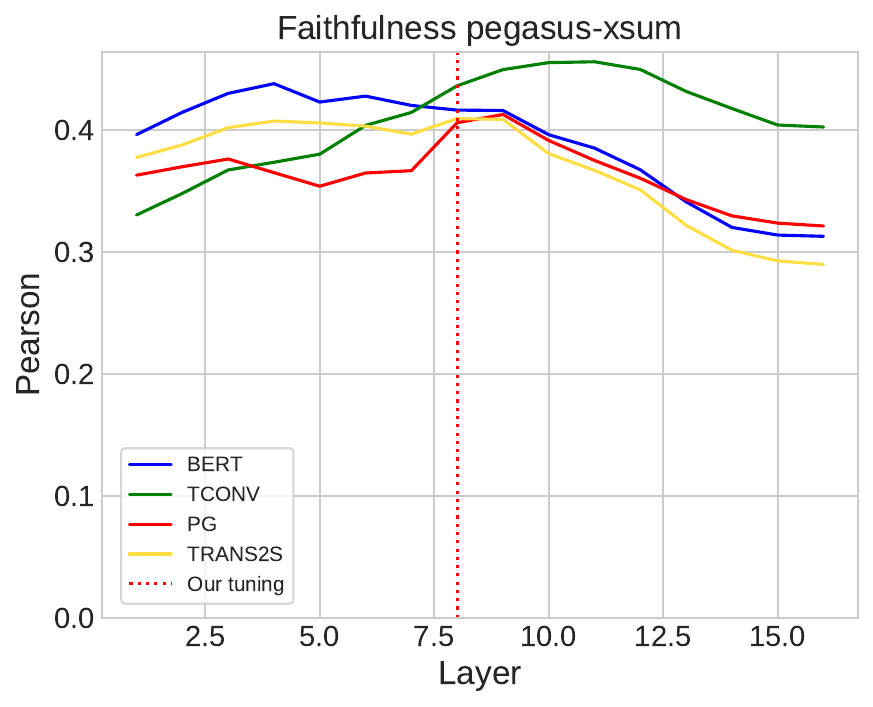}}
	\subfloat[]{\includegraphics[width = 0.33\linewidth]{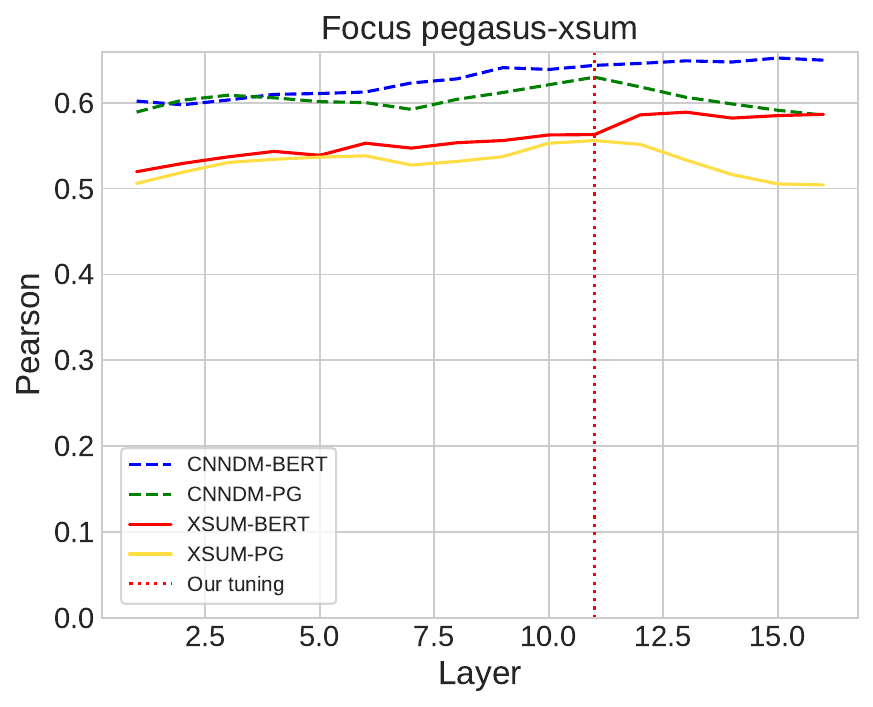}}
	\subfloat[]{\includegraphics[width = 0.33\linewidth]{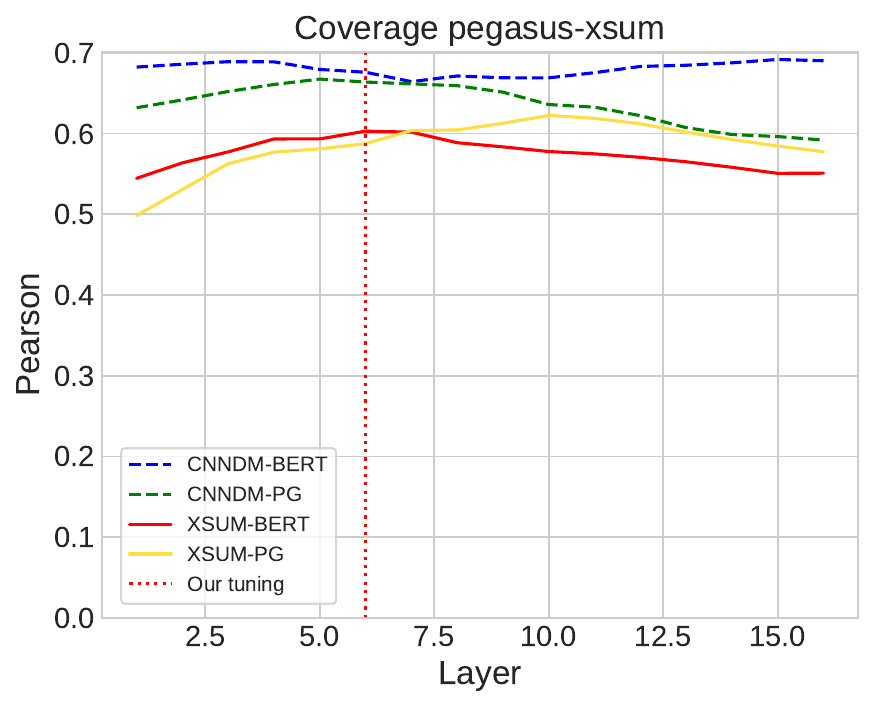}}\\
	\caption{\texttt{pegasus-xsum}.}
	\label{fig:pegasus-xsum}
\end{figure}

\begin{figure}[ht]
	\subfloat[]{\includegraphics[width = 0.33\linewidth]{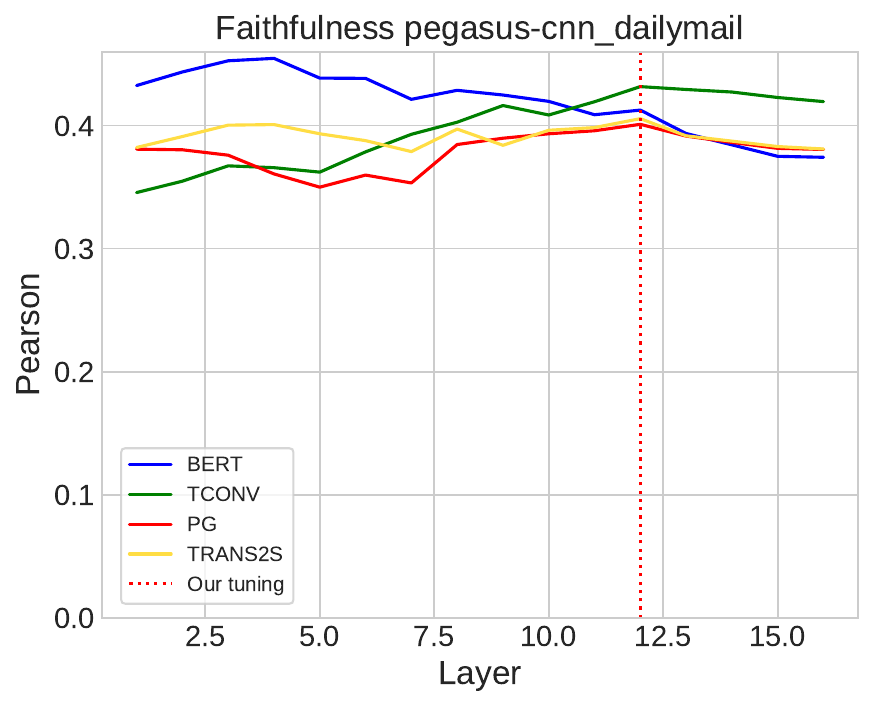}}
	\subfloat[]{\includegraphics[width = 0.33\linewidth]{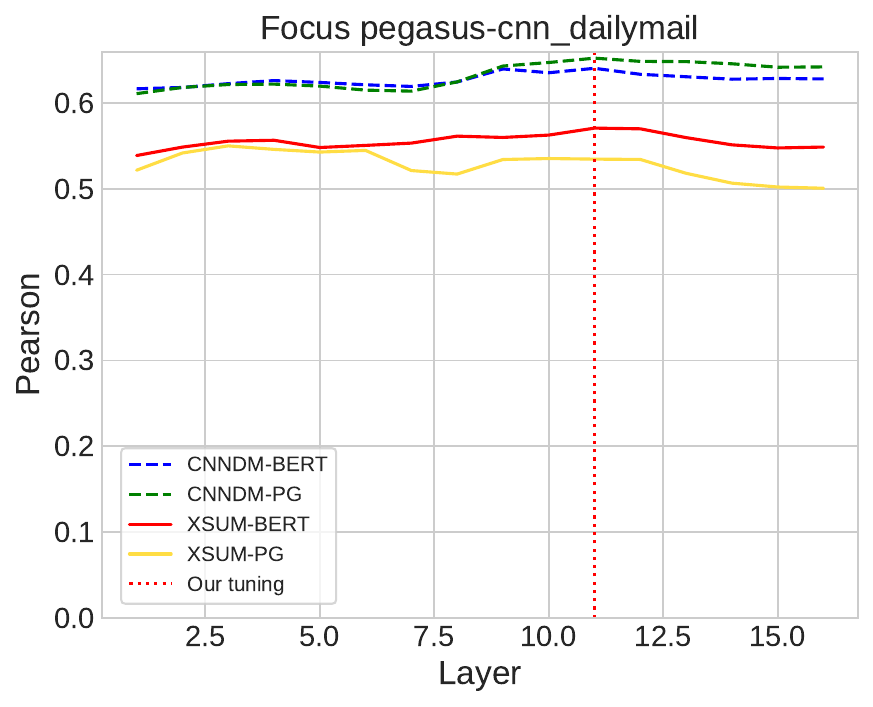}}
	\subfloat[]{\includegraphics[width = 0.33\linewidth]{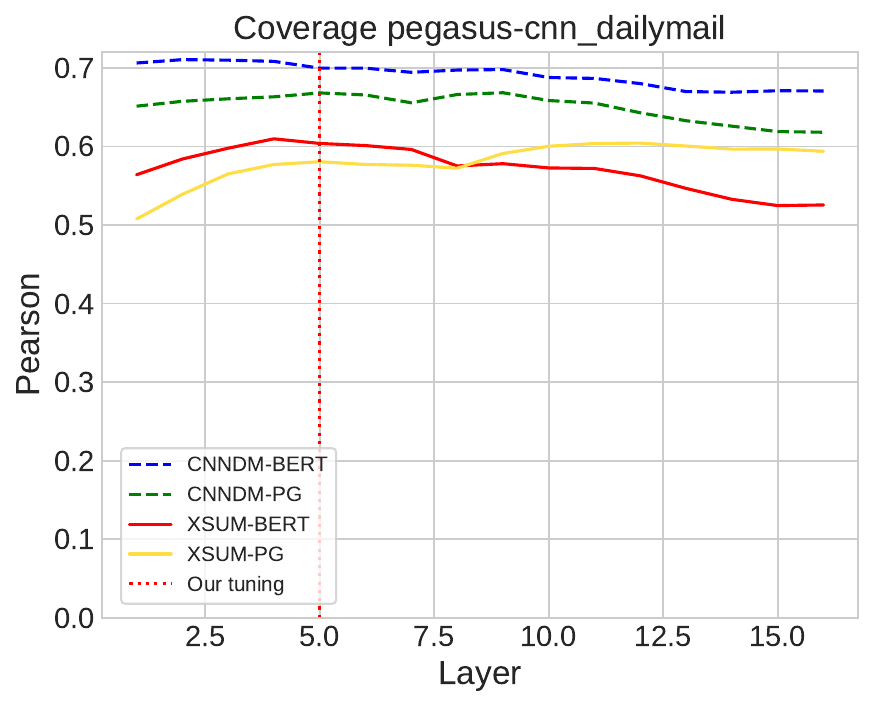}}\\
	\caption{\texttt{pegasus-cnn\_dailymail}.}
	\label{fig:pegasus-cnn_dailymail}
\end{figure}

\clearpage
\bibliography{sample,anthology,tacl2018,bibliography}

\begin{thebibliography}{}

\bibitem[\protect\BCAY{{Abacha}\ \BBA\ {Demner-Fushman}}{{Abacha}\ \BBA\
  {Demner-Fushman}}{2019}]{abacha2019on}
{Abacha}, A.~B.\BBACOMMA\  \BBA\ {Demner-Fushman}, D. \BBOP2019\BBCP.
\newblock \BBOQ On the summarization of consumer health questions\BBCQ\
\newblock In {\Bem Proceedings of the 57th Annual Meeting of the Association
  for Computational Linguistics}, \BPGS\ 2228--2234.

\bibitem[\protect\BCAY{Agirre, Cer, Diab,\ \BBA\ Gonzalez-Agirre}{Agirre
  et~al.}{2012}]{agirre-etal-2012-semeval}
Agirre, E., Cer, D., Diab, M., \BBA\ Gonzalez-Agirre, A. \BBOP2012\BBCP.
\newblock \BBOQ {S}em{E}val-2012 task 6: A pilot on semantic textual
  similarity\BBCQ\
\newblock In {\Bem *{SEM} 2012: The First Joint Conference on Lexical and
  Computational Semantics {--} Volume 1: Proceedings of the main conference and
  the shared task, and Volume 2: Proceedings of the Sixth International
  Workshop on Semantic Evaluation ({S}em{E}val 2012)}, \BPGS\ 385--393,
  Montr{\'e}al, Canada. Association for Computational Linguistics.

\bibitem[\protect\BCAY{Ahmad, Chakraborty, Ray,\ \BBA\ Chang}{Ahmad
  et~al.}{2020}]{ahmad-etal-2020-transformer}
Ahmad, W., Chakraborty, S., Ray, B., \BBA\ Chang, K.-W. \BBOP2020\BBCP.
\newblock \BBOQ A transformer-based approach for source code
  summarization\BBCQ\
\newblock In {\Bem Proceedings of the 58th Annual Meeting of the Association
  for Computational Linguistics}, \BPGS\ 4998--5007, Online. Association for
  Computational Linguistics.

\bibitem[\protect\BCAY{Amplayo\ \BBA\ Lapata}{Amplayo\ \BBA\
  Lapata}{2020}]{amplayo-lapata-2020-unsupervised}
Amplayo, R.~K.\BBACOMMA\  \BBA\ Lapata, M. \BBOP2020\BBCP.
\newblock \BBOQ Unsupervised opinion summarization with noising and
  denoising\BBCQ\
\newblock In {\Bem Proceedings of the 58th Annual Meeting of the Association
  for Computational Linguistics}, \BPGS\ 1934--1945, Online. Association for
  Computational Linguistics.

\bibitem[\protect\BCAY{Amplayo, Lim,\ \BBA\ Hwang}{Amplayo
  et~al.}{2018}]{amplayo-etal-2018-entity}
Amplayo, R.~K., Lim, S., \BBA\ Hwang, S.-w. \BBOP2018\BBCP.
\newblock \BBOQ Entity commonsense representation for neural abstractive
  summarization\BBCQ\
\newblock In {\Bem Proceedings of the 2018 Conference of the North {A}merican
  Chapter of the Association for Computational Linguistics: Human Language
  Technologies, Volume 1 (Long Papers)}, \BPGS\ 697--707, New Orleans,
  Louisiana. Association for Computational Linguistics.

\bibitem[\protect\BCAY{{Anderson}, {Fernando}, {Johnson},\ \BBA\
  {Gould}}{{Anderson} et~al.}{2016}]{anderson2016spice}
{Anderson}, P., {Fernando}, B., {Johnson}, M., \BBA\ {Gould}, S.
  \BBOP2016\BBCP.
\newblock \BBOQ {SPICE}: Semantic propositional image caption evaluation\BBCQ\
\newblock In {\Bem European Conference on Computer Vision (14th : 2016)},
  \BPGS\ 382--398.

\bibitem[\protect\BCAY{Bhandari, Gour, Ashfaq, Liu,\ \BBA\ Neubig}{Bhandari
  et~al.}{2020}]{bhandari-etal-2020-evaluating}
Bhandari, M., Gour, P.~N., Ashfaq, A., Liu, P., \BBA\ Neubig, G.
  \BBOP2020\BBCP.
\newblock \BBOQ Re-evaluating evaluation in text summarization\BBCQ\
\newblock In {\Bem Proceedings of the 2020 Conference on Empirical Methods in
  Natural Language Processing (EMNLP)}, \BPGS\ 9347--9359, Online. Association
  for Computational Linguistics.

\bibitem[\protect\BCAY{Bommasani\ \BBA\ Cardie}{Bommasani\ \BBA\
  Cardie}{2020}]{bommasani-cardie-2020-intrinsic}
Bommasani, R.\BBACOMMA\  \BBA\ Cardie, C. \BBOP2020\BBCP.
\newblock \BBOQ Intrinsic evaluation of summarization datasets\BBCQ\
\newblock In {\Bem Proceedings of the 2020 Conference on Empirical Methods in
  Natural Language Processing (EMNLP)}, \BPGS\ 8075--8096, Online. Association
  for Computational Linguistics.

\bibitem[\protect\BCAY{Bra{\v{z}}inskas, Lapata,\ \BBA\ Titov}{Bra{\v{z}}inskas
  et~al.}{2020}]{brazinskas-etal-2020-unsupervised}
Bra{\v{z}}inskas, A., Lapata, M., \BBA\ Titov, I. \BBOP2020\BBCP.
\newblock \BBOQ Unsupervised opinion summarization as copycat-review
  generation\BBCQ\
\newblock In {\Bem Proceedings of the 58th Annual Meeting of the Association
  for Computational Linguistics}, \BPGS\ 5151--5169, Online. Association for
  Computational Linguistics.

\bibitem[\protect\BCAY{Cao, Dong, Wu,\ \BBA\ Cheung}{Cao
  et~al.}{2020}]{cao-etal-2020-factual}
Cao, M., Dong, Y., Wu, J., \BBA\ Cheung, J. C.~K. \BBOP2020\BBCP.
\newblock \BBOQ Factual error correction for abstractive summarization
  models\BBCQ\
\newblock In {\Bem Proceedings of the 2020 Conference on Empirical Methods in
  Natural Language Processing (EMNLP)}, \BPGS\ 6251--6258, Online. Association
  for Computational Linguistics.

\bibitem[\protect\BCAY{Cao, Li, Li,\ \BBA\ Wei}{Cao
  et~al.}{2018}]{cao-etal-2018-retrieve}
Cao, Z., Li, W., Li, S., \BBA\ Wei, F. \BBOP2018\BBCP.
\newblock \BBOQ Retrieve, rerank and rewrite: Soft template based neural
  summarization\BBCQ\
\newblock In {\Bem Proceedings of the 56th Annual Meeting of the Association
  for Computational Linguistics (Volume 1: Long Papers)}, \BPGS\ 152--161,
  Melbourne, Australia. Association for Computational Linguistics.

\bibitem[\protect\BCAY{Celikyilmaz, Bosselut, He,\ \BBA\ Choi}{Celikyilmaz
  et~al.}{2018}]{celikyilmaz-etal-2018-deep}
Celikyilmaz, A., Bosselut, A., He, X., \BBA\ Choi, Y. \BBOP2018\BBCP.
\newblock \BBOQ Deep communicating agents for abstractive summarization\BBCQ\
\newblock In {\Bem Proceedings of the 2018 Conference of the North {A}merican
  Chapter of the Association for Computational Linguistics: Human Language
  Technologies, Volume 1 (Long Papers)}, \BPGS\ 1662--1675, New Orleans,
  Louisiana. Association for Computational Linguistics.

\bibitem[\protect\BCAY{Chen\ \BBA\ Bansal}{Chen\ \BBA\
  Bansal}{2018}]{chen-bansal-2018-fast}
Chen, Y.-C.\BBACOMMA\  \BBA\ Bansal, M. \BBOP2018\BBCP.
\newblock \BBOQ Fast abstractive summarization with reinforce-selected sentence
  rewriting\BBCQ\
\newblock In {\Bem Proceedings of the 56th Annual Meeting of the Association
  for Computational Linguistics (Volume 1: Long Papers)}, \BPGS\ 675--686,
  Melbourne, Australia. Association for Computational Linguistics.

\bibitem[\protect\BCAY{Cohan, Dernoncourt, Kim, Bui, Kim, Chang,\ \BBA\
  Goharian}{Cohan et~al.}{2018}]{cohan-etal-2018-discourse}
Cohan, A., Dernoncourt, F., Kim, D.~S., Bui, T., Kim, S., Chang, W., \BBA\
  Goharian, N. \BBOP2018\BBCP.
\newblock \BBOQ A discourse-aware attention model for abstractive summarization
  of long documents\BBCQ\
\newblock In {\Bem Proceedings of the 2018 Conference of the North {A}merican
  Chapter of the Association for Computational Linguistics: Human Language
  Technologies, Volume 2 (Short Papers)}, \BPGS\ 615--621, New Orleans,
  Louisiana. Association for Computational Linguistics.

\bibitem[\protect\BCAY{Cui, Hu,\ \BBA\ Liu}{Cui
  et~al.}{2020}]{cui-etal-2020-enhancing}
Cui, P., Hu, L., \BBA\ Liu, Y. \BBOP2020\BBCP.
\newblock \BBOQ Enhancing extractive text summarization with topic-aware graph
  neural networks\BBCQ\
\newblock In {\Bem Proceedings of the 28th International Conference on
  Computational Linguistics}, \BPGS\ 5360--5371, Barcelona, Spain (Online).
  International Committee on Computational Linguistics.

\bibitem[\protect\BCAY{Deng, Zhang,\ \BBA\ Lam}{Deng
  et~al.}{2020}]{deng-etal-2020-multi}
Deng, Y., Zhang, W., \BBA\ Lam, W. \BBOP2020\BBCP.
\newblock \BBOQ Multi-hop inference for question-driven summarization\BBCQ\
\newblock In {\Bem Proceedings of the 2020 Conference on Empirical Methods in
  Natural Language Processing (EMNLP)}, \BPGS\ 6734--6744, Online. Association
  for Computational Linguistics.

\bibitem[\protect\BCAY{Desai, Xu,\ \BBA\ Durrett}{Desai
  et~al.}{2020}]{desai-etal-2020-compressive}
Desai, S., Xu, J., \BBA\ Durrett, G. \BBOP2020\BBCP.
\newblock \BBOQ Compressive summarization with plausibility and salience
  modeling\BBCQ\
\newblock In {\Bem Proceedings of the 2020 Conference on Empirical Methods in
  Natural Language Processing (EMNLP)}, \BPGS\ 6259--6274, Online. Association
  for Computational Linguistics.

\bibitem[\protect\BCAY{Devlin, Chang, Lee,\ \BBA\ Toutanova}{Devlin
  et~al.}{2019}]{devlin-etal-2019-bert}
Devlin, J., Chang, M.-W., Lee, K., \BBA\ Toutanova, K. \BBOP2019\BBCP.
\newblock \BBOQ {BERT}: Pre-training of deep bidirectional transformers for
  language understanding\BBCQ\
\newblock In {\Bem Proceedings of the 2019 Conference of the North {A}merican
  Chapter of the Association for Computational Linguistics: Human Language
  Technologies, Volume 1 (Long and Short Papers)}, \BPGS\ 4171--4186,
  Minneapolis, Minnesota. Association for Computational Linguistics.

\bibitem[\protect\BCAY{{Dong}, {Shen}, {Crawford}, van {Hoof},\ \BBA\
  {Cheung}}{{Dong} et~al.}{2018}]{dong2018banditsum}
{Dong}, Y., {Shen}, Y., {Crawford}, E., van {Hoof}, H., \BBA\ {Cheung}, J.
  C.~K. \BBOP2018\BBCP.
\newblock \BBOQ Banditsum: Extractive summarization as a contextual
  bandit\BBCQ\
\newblock In {\Bem EMNLP 2018: 2018 Conference on Empirical Methods in Natural
  Language Processing}, \BPGS\ 3739--3748.

\bibitem[\protect\BCAY{Du, Shao,\ \BBA\ Cardie}{Du
  et~al.}{2017}]{du-etal-2017-learning}
Du, X., Shao, J., \BBA\ Cardie, C. \BBOP2017\BBCP.
\newblock \BBOQ Learning to ask: Neural question generation for reading
  comprehension\BBCQ\
\newblock In {\Bem Proceedings of the 55th Annual Meeting of the Association
  for Computational Linguistics (Volume 1: Long Papers)}, \BPGS\ 1342--1352,
  Vancouver, Canada. Association for Computational Linguistics.

\bibitem[\protect\BCAY{Duan, Yin, Zhang, Chen,\ \BBA\ Luo}{Duan
  et~al.}{2019a}]{duan-etal-2019-zero}
Duan, X., Yin, M., Zhang, M., Chen, B., \BBA\ Luo, W. \BBOP2019a\BBCP.
\newblock \BBOQ Zero-shot cross-lingual abstractive sentence summarization
  through teaching generation and attention\BBCQ\
\newblock In {\Bem Proceedings of the 57th Annual Meeting of the Association
  for Computational Linguistics}, \BPGS\ 3162--3172, Florence, Italy.
  Association for Computational Linguistics.

\bibitem[\protect\BCAY{Duan, Yu, Yin, Zhang, Luo,\ \BBA\ Zhang}{Duan
  et~al.}{2019b}]{duan-etal-2019-contrastive}
Duan, X., Yu, H., Yin, M., Zhang, M., Luo, W., \BBA\ Zhang, Y. \BBOP2019b\BBCP.
\newblock \BBOQ Contrastive attention mechanism for abstractive sentence
  summarization\BBCQ\
\newblock In {\Bem Proceedings of the 2019 Conference on Empirical Methods in
  Natural Language Processing and the 9th International Joint Conference on
  Natural Language Processing (EMNLP-IJCNLP)}, \BPGS\ 3044--3053, Hong Kong,
  China. Association for Computational Linguistics.

\bibitem[\protect\BCAY{Durmus, He,\ \BBA\ Diab}{Durmus
  et~al.}{2020}]{durmus-etal-2020-feqa}
Durmus, E., He, H., \BBA\ Diab, M. \BBOP2020\BBCP.
\newblock \BBOQ {FEQA}: A question answering evaluation framework for
  faithfulness assessment in abstractive summarization\BBCQ\
\newblock In {\Bem Proceedings of the 58th Annual Meeting of the Association
  for Computational Linguistics}, \BPGS\ 5055--5070, Online. Association for
  Computational Linguistics.

\bibitem[\protect\BCAY{Fabbri, Li, She, Li,\ \BBA\ Radev}{Fabbri
  et~al.}{2019}]{fabbri-etal-2019-multi}
Fabbri, A., Li, I., She, T., Li, S., \BBA\ Radev, D. \BBOP2019\BBCP.
\newblock \BBOQ Multi-news: A large-scale multi-document summarization dataset
  and abstractive hierarchical model\BBCQ\
\newblock In {\Bem Proceedings of the 57th Annual Meeting of the Association
  for Computational Linguistics}, \BPGS\ 1074--1084, Florence, Italy.
  Association for Computational Linguistics.

\bibitem[\protect\BCAY{Fabbri, Kry{\'s}ci{\'n}ski, McCann, Xiong, Socher,\
  \BBA\ Radev}{Fabbri et~al.}{2020}]{fabbri2020summeval}
Fabbri, A.~R., Kry{\'s}ci{\'n}ski, W., McCann, B., Xiong, C., Socher, R., \BBA\
  Radev, D. \BBOP2020\BBCP.
\newblock \BBOQ Summeval: Re-evaluating summarization evaluation\BBCQ\
\newblock {\Bem arXiv preprint arXiv:2007.12626}.

\bibitem[\protect\BCAY{Falke\ \BBA\ Gurevych}{Falke\ \BBA\
  Gurevych}{2019}]{falke-gurevych-2019-fast}
Falke, T.\BBACOMMA\  \BBA\ Gurevych, I. \BBOP2019\BBCP.
\newblock \BBOQ Fast concept mention grouping for concept map-based
  multi-document summarization\BBCQ\
\newblock In {\Bem Proceedings of the 2019 Conference of the North {A}merican
  Chapter of the Association for Computational Linguistics: Human Language
  Technologies, Volume 1 (Long and Short Papers)}, \BPGS\ 695--700,
  Minneapolis, Minnesota. Association for Computational Linguistics.

\bibitem[\protect\BCAY{Frermann\ \BBA\ Klementiev}{Frermann\ \BBA\
  Klementiev}{2019}]{frermann-klementiev-2019-inducing}
Frermann, L.\BBACOMMA\  \BBA\ Klementiev, A. \BBOP2019\BBCP.
\newblock \BBOQ Inducing document structure for aspect-based
  summarization\BBCQ\
\newblock In {\Bem Proceedings of the 57th Annual Meeting of the Association
  for Computational Linguistics}, \BPGS\ 6263--6273, Florence, Italy.
  Association for Computational Linguistics.

\bibitem[\protect\BCAY{Gao, Chen, Li, Chan, Zhao,\ \BBA\ Yan}{Gao
  et~al.}{2019}]{gao-etal-2019-write}
Gao, S., Chen, X., Li, P., Chan, Z., Zhao, D., \BBA\ Yan, R. \BBOP2019\BBCP.
\newblock \BBOQ How to write summaries with patterns? learning towards
  abstractive summarization through prototype editing\BBCQ\
\newblock In {\Bem Proceedings of the 2019 Conference on Empirical Methods in
  Natural Language Processing and the 9th International Joint Conference on
  Natural Language Processing (EMNLP-IJCNLP)}, \BPGS\ 3741--3751, Hong Kong,
  China. Association for Computational Linguistics.

\bibitem[\protect\BCAY{Gao, Zhao,\ \BBA\ Eger}{Gao
  et~al.}{2020}]{gao-etal-2020-supert}
Gao, Y., Zhao, W., \BBA\ Eger, S. \BBOP2020\BBCP.
\newblock \BBOQ {SUPERT}: Towards new frontiers in unsupervised evaluation
  metrics for multi-document summarization\BBCQ\
\newblock In {\Bem Proceedings of the 58th Annual Meeting of the Association
  for Computational Linguistics}, \BPGS\ 1347--1354, Online. Association for
  Computational Linguistics.

\bibitem[\protect\BCAY{Gehrmann, Deng,\ \BBA\ Rush}{Gehrmann
  et~al.}{2018}]{gehrmann-etal-2018-bottom}
Gehrmann, S., Deng, Y., \BBA\ Rush, A. \BBOP2018\BBCP.
\newblock \BBOQ Bottom-up abstractive summarization\BBCQ\
\newblock In {\Bem Proceedings of the 2018 Conference on Empirical Methods in
  Natural Language Processing}, \BPGS\ 4098--4109, Brussels, Belgium.
  Association for Computational Linguistics.

\bibitem[\protect\BCAY{Gholipour~Ghalandari, Hokamp, Pham, Glover,\ \BBA\
  Ifrim}{Gholipour~Ghalandari
  et~al.}{2020}]{gholipour-ghalandari-etal-2020-large}
Gholipour~Ghalandari, D., Hokamp, C., Pham, N.~T., Glover, J., \BBA\ Ifrim, G.
  \BBOP2020\BBCP.
\newblock \BBOQ A large-scale multi-document summarization dataset from the
  {W}ikipedia current events portal\BBCQ\
\newblock In {\Bem Proceedings of the 58th Annual Meeting of the Association
  for Computational Linguistics}, \BPGS\ 1302--1308, Online. Association for
  Computational Linguistics.

\bibitem[\protect\BCAY{Gholipour~Ghalandari\ \BBA\ Ifrim}{Gholipour~Ghalandari\
  \BBA\ Ifrim}{2020}]{gholipour-ghalandari-ifrim-2020-examining}
Gholipour~Ghalandari, D.\BBACOMMA\  \BBA\ Ifrim, G. \BBOP2020\BBCP.
\newblock \BBOQ Examining the state-of-the-art in news timeline
  summarization\BBCQ\
\newblock In {\Bem Proceedings of the 58th Annual Meeting of the Association
  for Computational Linguistics}, \BPGS\ 1322--1334, Online. Association for
  Computational Linguistics.

\bibitem[\protect\BCAY{Graham, Baldwin, Dowling, Eskevich, Lynn,\ \BBA\
  Tounsi}{Graham et~al.}{2016}]{graham-etal-2016-glitters}
Graham, Y., Baldwin, T., Dowling, M., Eskevich, M., Lynn, T., \BBA\ Tounsi, L.
  \BBOP2016\BBCP.
\newblock \BBOQ Is all that glitters in machine translation quality estimation
  really gold?\BBCQ\
\newblock In {\Bem Proceedings of {COLING} 2016, the 26th International
  Conference on Computational Linguistics: Technical Papers}, \BPGS\
  3124--3134, Osaka, Japan. The COLING 2016 Organizing Committee.

\bibitem[\protect\BCAY{Graham, Baldwin,\ \BBA\ Mathur}{Graham
  et~al.}{2015}]{graham-etal-2015-accurate}
Graham, Y., Baldwin, T., \BBA\ Mathur, N. \BBOP2015\BBCP.
\newblock \BBOQ Accurate evaluation of segment-level machine translation
  metrics\BBCQ\
\newblock In {\Bem Proceedings of the 2015 Conference of the North {A}merican
  Chapter of the Association for Computational Linguistics: Human Language
  Technologies}, \BPGS\ 1183--1191, Denver, Colorado. Association for
  Computational Linguistics.

\bibitem[\protect\BCAY{{Graham}, {Baldwin}, {Moffat},\ \BBA\ {Zobel}}{{Graham}
  et~al.}{2017}]{graham2017can}
{Graham}, Y., {Baldwin}, T., {Moffat}, A., \BBA\ {Zobel}, J. \BBOP2017\BBCP.
\newblock \BBOQ Can machine translation systems be evaluated by the crowd
  alone\BBCQ\
\newblock {\Bem Natural Language Engineering}, {\Bem 23\/}(1), 3--30.

\bibitem[\protect\BCAY{Grenander, Dong, Cheung,\ \BBA\ Louis}{Grenander
  et~al.}{2019}]{grenander-etal-2019-countering}
Grenander, M., Dong, Y., Cheung, J. C.~K., \BBA\ Louis, A. \BBOP2019\BBCP.
\newblock \BBOQ Countering the effects of lead bias in news summarization via
  multi-stage training and auxiliary losses\BBCQ\
\newblock In {\Bem Proceedings of the 2019 Conference on Empirical Methods in
  Natural Language Processing and the 9th International Joint Conference on
  Natural Language Processing (EMNLP-IJCNLP)}, \BPGS\ 6019--6024, Hong Kong,
  China. Association for Computational Linguistics.

\bibitem[\protect\BCAY{Gui, Tian, Wang,\ \BBA\ Yang}{Gui
  et~al.}{2019}]{gui-etal-2019-attention}
Gui, M., Tian, J., Wang, R., \BBA\ Yang, Z. \BBOP2019\BBCP.
\newblock \BBOQ Attention optimization for abstractive document
  summarization\BBCQ\
\newblock In {\Bem Proceedings of the 2019 Conference on Empirical Methods in
  Natural Language Processing and the 9th International Joint Conference on
  Natural Language Processing (EMNLP-IJCNLP)}, \BPGS\ 1222--1228, Hong Kong,
  China. Association for Computational Linguistics.

\bibitem[\protect\BCAY{Guo, Pasunuru,\ \BBA\ Bansal}{Guo
  et~al.}{2018}]{guo-etal-2018-soft}
Guo, H., Pasunuru, R., \BBA\ Bansal, M. \BBOP2018\BBCP.
\newblock \BBOQ Soft layer-specific multi-task summarization with entailment
  and question generation\BBCQ\
\newblock In {\Bem Proceedings of the 56th Annual Meeting of the Association
  for Computational Linguistics (Volume 1: Long Papers)}, \BPGS\ 687--697,
  Melbourne, Australia. Association for Computational Linguistics.

\bibitem[\protect\BCAY{Hardy, Narayan,\ \BBA\ Vlachos}{Hardy
  et~al.}{2019}]{hardy-etal-2019-highres}
Hardy, H., Narayan, S., \BBA\ Vlachos, A. \BBOP2019\BBCP.
\newblock \BBOQ {H}igh{RES}: Highlight-based reference-less evaluation of
  summarization\BBCQ\
\newblock In {\Bem Proceedings of the 57th Annual Meeting of the Association
  for Computational Linguistics}, \BPGS\ 3381--3392, Florence, Italy.
  Association for Computational Linguistics.

\bibitem[\protect\BCAY{Hardy\ \BBA\ Vlachos}{Hardy\ \BBA\
  Vlachos}{2018}]{hardy-vlachos-2018-guided}
Hardy, H.\BBACOMMA\  \BBA\ Vlachos, A. \BBOP2018\BBCP.
\newblock \BBOQ Guided neural language generation for abstractive summarization
  using {A}bstract {M}eaning {R}epresentation\BBCQ\
\newblock In {\Bem Proceedings of the 2018 Conference on Empirical Methods in
  Natural Language Processing}, \BPGS\ 768--773, Brussels, Belgium. Association
  for Computational Linguistics.

\bibitem[\protect\BCAY{He, Zhao,\ \BBA\ Liu}{He
  et~al.}{2020}]{he-etal-2020-tweetsum}
He, R., Zhao, L., \BBA\ Liu, H. \BBOP2020\BBCP.
\newblock \BBOQ {TWEETSUM}: Event oriented social summarization dataset\BBCQ\
\newblock In {\Bem Proceedings of the 28th International Conference on
  Computational Linguistics}, \BPGS\ 5731--5736, Barcelona, Spain (Online).
  International Committee on Computational Linguistics.

\bibitem[\protect\BCAY{{Hermann}, {Kocisky}, {Grefenstette}, {Espeholt}, {Kay},
  {Suleyman},\ \BBA\ {Blunsom}}{{Hermann} et~al.}{2015}]{hermann2015teaching}
{Hermann}, K.~M., {Kocisky}, T., {Grefenstette}, E., {Espeholt}, L., {Kay}, W.,
  {Suleyman}, M., \BBA\ {Blunsom}, P. \BBOP2015\BBCP.
\newblock \BBOQ Teaching machines to read and comprehend\BBCQ\
\newblock In {\Bem Neural Information Processing Systems}, \BPGS\ 1693--1701.

\bibitem[\protect\BCAY{Hsu, Lin, Lee, Min, Tang,\ \BBA\ Sun}{Hsu
  et~al.}{2018}]{hsu-etal-2018-unified}
Hsu, W.-T., Lin, C.-K., Lee, M.-Y., Min, K., Tang, J., \BBA\ Sun, M.
  \BBOP2018\BBCP.
\newblock \BBOQ A unified model for extractive and abstractive summarization
  using inconsistency loss\BBCQ\
\newblock In {\Bem Proceedings of the 56th Annual Meeting of the Association
  for Computational Linguistics (Volume 1: Long Papers)}, \BPGS\ 132--141,
  Melbourne, Australia. Association for Computational Linguistics.

\bibitem[\protect\BCAY{Huang, Cui, Yang, Bao, Wang, Xie,\ \BBA\ Zhang}{Huang
  et~al.}{2020a}]{huang-etal-2020-achieved}
Huang, D., Cui, L., Yang, S., Bao, G., Wang, K., Xie, J., \BBA\ Zhang, Y.
  \BBOP2020a\BBCP.
\newblock \BBOQ What have we achieved on text summarization?\BBCQ\
\newblock In {\Bem Proceedings of the 2020 Conference on Empirical Methods in
  Natural Language Processing (EMNLP)}, \BPGS\ 446--469, Online. Association
  for Computational Linguistics.

\bibitem[\protect\BCAY{Huang, Li,\ \BBA\ Chang}{Huang
  et~al.}{2020b}]{huang-etal-2020-generating}
Huang, K.-H., Li, C., \BBA\ Chang, K.-W. \BBOP2020b\BBCP.
\newblock \BBOQ Generating sports news from live commentary: A {C}hinese
  dataset for sports game summarization\BBCQ\
\newblock In {\Bem Proceedings of the 1st Conference of the Asia-Pacific
  Chapter of the Association for Computational Linguistics and the 10th
  International Joint Conference on Natural Language Processing}, \BPGS\
  609--615, Suzhou, China. Association for Computational Linguistics.

\bibitem[\protect\BCAY{Huang, Wu,\ \BBA\ Wang}{Huang
  et~al.}{2020c}]{huang-etal-2020-knowledge}
Huang, L., Wu, L., \BBA\ Wang, L. \BBOP2020c\BBCP.
\newblock \BBOQ Knowledge graph-augmented abstractive summarization with
  semantic-driven cloze reward\BBCQ\
\newblock In {\Bem Proceedings of the 58th Annual Meeting of the Association
  for Computational Linguistics}, \BPGS\ 5094--5107, Online. Association for
  Computational Linguistics.

\bibitem[\protect\BCAY{Isonuma, Mori,\ \BBA\ Sakata}{Isonuma
  et~al.}{2019}]{isonuma-etal-2019-unsupervised}
Isonuma, M., Mori, J., \BBA\ Sakata, I. \BBOP2019\BBCP.
\newblock \BBOQ Unsupervised neural single-document summarization of reviews
  via learning latent discourse structure and its ranking\BBCQ\
\newblock In {\Bem Proceedings of the 57th Annual Meeting of the Association
  for Computational Linguistics}, \BPGS\ 2142--2152, Florence, Italy.
  Association for Computational Linguistics.

\bibitem[\protect\BCAY{Jadhav\ \BBA\ Rajan}{Jadhav\ \BBA\
  Rajan}{2018}]{jadhav-rajan-2018-extractive}
Jadhav, A.\BBACOMMA\  \BBA\ Rajan, V. \BBOP2018\BBCP.
\newblock \BBOQ Extractive summarization with {SWAP}-{NET}: Sentences and words
  from alternating pointer networks\BBCQ\
\newblock In {\Bem Proceedings of the 56th Annual Meeting of the Association
  for Computational Linguistics (Volume 1: Long Papers)}, \BPGS\ 142--151,
  Melbourne, Australia. Association for Computational Linguistics.

\bibitem[\protect\BCAY{Ji\ \BBA\ Eisenstein}{Ji\ \BBA\
  Eisenstein}{2014}]{ji-eisenstein-2014-representation}
Ji, Y.\BBACOMMA\  \BBA\ Eisenstein, J. \BBOP2014\BBCP.
\newblock \BBOQ Representation learning for text-level discourse parsing\BBCQ\
\newblock In {\Bem Proceedings of the 52nd Annual Meeting of the Association
  for Computational Linguistics (Volume 1: Long Papers)}, \BPGS\ 13--24,
  Baltimore, Maryland. Association for Computational Linguistics.

\bibitem[\protect\BCAY{{Jia}, {Rajpurkar},\ \BBA\ {Liang}}{{Jia}
  et~al.}{2018}]{jia2018know}
{Jia}, R., {Rajpurkar}, P., \BBA\ {Liang}, P. \BBOP2018\BBCP.
\newblock \BBOQ Know what you don't know: Unanswerable questions for
  squad\BBCQ\
\newblock In {\Bem ACL 2018: 56th Annual Meeting of the Association for
  Computational Linguistics}, \lowercase{\BVOL}~2, \BPGS\ 784--789.

\bibitem[\protect\BCAY{Jia, Cao, Tang, Fang, Cao,\ \BBA\ Wang}{Jia
  et~al.}{2020}]{jia-etal-2020-neural}
Jia, R., Cao, Y., Tang, H., Fang, F., Cao, C., \BBA\ Wang, S. \BBOP2020\BBCP.
\newblock \BBOQ Neural extractive summarization with hierarchical attentive
  heterogeneous graph network\BBCQ\
\newblock In {\Bem Proceedings of the 2020 Conference on Empirical Methods in
  Natural Language Processing (EMNLP)}, \BPGS\ 3622--3631, Online. Association
  for Computational Linguistics.

\bibitem[\protect\BCAY{Kano, Miura, Taniguchi,\ \BBA\ Ohkuma}{Kano
  et~al.}{2020}]{kano-etal-2020-identifying}
Kano, R., Miura, Y., Taniguchi, T., \BBA\ Ohkuma, T. \BBOP2020\BBCP.
\newblock \BBOQ Identifying implicit quotes for unsupervised extractive
  summarization of conversations\BBCQ\
\newblock In {\Bem Proceedings of the 1st Conference of the Asia-Pacific
  Chapter of the Association for Computational Linguistics and the 10th
  International Joint Conference on Natural Language Processing}, \BPGS\
  291--302, Suzhou, China. Association for Computational Linguistics.

\bibitem[\protect\BCAY{Kedzie, McKeown,\ \BBA\ Daum{\'e}~III}{Kedzie
  et~al.}{2018}]{kedzie-etal-2018-content}
Kedzie, C., McKeown, K., \BBA\ Daum{\'e}~III, H. \BBOP2018\BBCP.
\newblock \BBOQ Content selection in deep learning models of
  summarization\BBCQ\
\newblock In {\Bem Proceedings of the 2018 Conference on Empirical Methods in
  Natural Language Processing}, \BPGS\ 1818--1828, Brussels, Belgium.
  Association for Computational Linguistics.

\bibitem[\protect\BCAY{Kim, Kim,\ \BBA\ Kim}{Kim
  et~al.}{2019}]{kim-etal-2019-abstractive}
Kim, B., Kim, H., \BBA\ Kim, G. \BBOP2019\BBCP.
\newblock \BBOQ Abstractive summarization of {R}eddit posts with multi-level
  memory networks\BBCQ\
\newblock In {\Bem Proceedings of the 2019 Conference of the North {A}merican
  Chapter of the Association for Computational Linguistics: Human Language
  Technologies, Volume 1 (Long and Short Papers)}, \BPGS\ 2519--2531,
  Minneapolis, Minnesota. Association for Computational Linguistics.

\bibitem[\protect\BCAY{Koto, Lau,\ \BBA\ Baldwin}{Koto
  et~al.}{2020}]{koto-etal-2020-liputan6}
Koto, F., Lau, J.~H., \BBA\ Baldwin, T. \BBOP2020\BBCP.
\newblock \BBOQ Liputan6: A large-scale {I}ndonesian dataset for text
  summarization\BBCQ\
\newblock In {\Bem Proceedings of the 1st Conference of the Asia-Pacific
  Chapter of the Association for Computational Linguistics and the 10th
  International Joint Conference on Natural Language Processing}, \BPGS\
  598--608, Suzhou, China. Association for Computational Linguistics.

\bibitem[\protect\BCAY{Kouris, Alexandridis,\ \BBA\ Stafylopatis}{Kouris
  et~al.}{2019}]{kouris-etal-2019-abstractive}
Kouris, P., Alexandridis, G., \BBA\ Stafylopatis, A. \BBOP2019\BBCP.
\newblock \BBOQ Abstractive text summarization based on deep learning and
  semantic content generalization\BBCQ\
\newblock In {\Bem Proceedings of the 57th Annual Meeting of the Association
  for Computational Linguistics}, \BPGS\ 5082--5092, Florence, Italy.
  Association for Computational Linguistics.

\bibitem[\protect\BCAY{Krishna\ \BBA\ Srinivasan}{Krishna\ \BBA\
  Srinivasan}{2018}]{krishna-srinivasan-2018-generating}
Krishna, K.\BBACOMMA\  \BBA\ Srinivasan, B.~V. \BBOP2018\BBCP.
\newblock \BBOQ Generating topic-oriented summaries using neural
  attention\BBCQ\
\newblock In {\Bem Proceedings of the 2018 Conference of the North {A}merican
  Chapter of the Association for Computational Linguistics: Human Language
  Technologies, Volume 1 (Long Papers)}, \BPGS\ 1697--1705, New Orleans,
  Louisiana. Association for Computational Linguistics.

\bibitem[\protect\BCAY{Kryscinski, McCann, Xiong,\ \BBA\ Socher}{Kryscinski
  et~al.}{2020}]{kryscinski-etal-2020-evaluating}
Kryscinski, W., McCann, B., Xiong, C., \BBA\ Socher, R. \BBOP2020\BBCP.
\newblock \BBOQ Evaluating the factual consistency of abstractive text
  summarization\BBCQ\
\newblock In {\Bem Proceedings of the 2020 Conference on Empirical Methods in
  Natural Language Processing (EMNLP)}, \BPGS\ 9332--9346, Online. Association
  for Computational Linguistics.

\bibitem[\protect\BCAY{Kry{\'s}ci{\'n}ski, Paulus, Xiong,\ \BBA\
  Socher}{Kry{\'s}ci{\'n}ski et~al.}{2018}]{kryscinski-etal-2018-improving}
Kry{\'s}ci{\'n}ski, W., Paulus, R., Xiong, C., \BBA\ Socher, R. \BBOP2018\BBCP.
\newblock \BBOQ Improving abstraction in text summarization\BBCQ\
\newblock In {\Bem Proceedings of the 2018 Conference on Empirical Methods in
  Natural Language Processing}, \BPGS\ 1808--1817, Brussels, Belgium.
  Association for Computational Linguistics.

\bibitem[\protect\BCAY{Ladhak, Li, Al-Onaizan,\ \BBA\ McKeown}{Ladhak
  et~al.}{2020}]{ladhak-etal-2020-exploring}
Ladhak, F., Li, B., Al-Onaizan, Y., \BBA\ McKeown, K. \BBOP2020\BBCP.
\newblock \BBOQ Exploring content selection in summarization of novel
  chapters\BBCQ\
\newblock In {\Bem Proceedings of the 58th Annual Meeting of the Association
  for Computational Linguistics}, \BPGS\ 5043--5054, Online. Association for
  Computational Linguistics.

\bibitem[\protect\BCAY{Lavie\ \BBA\ Agarwal}{Lavie\ \BBA\
  Agarwal}{2007}]{lavie-agarwal-2007-meteor}
Lavie, A.\BBACOMMA\  \BBA\ Agarwal, A. \BBOP2007\BBCP.
\newblock \BBOQ {METEOR}: An automatic metric for {MT} evaluation with high
  levels of correlation with human judgments\BBCQ\
\newblock In {\Bem Proceedings of the Second Workshop on Statistical Machine
  Translation}, \BPGS\ 228--231, Prague, Czech Republic. Association for
  Computational Linguistics.

\bibitem[\protect\BCAY{Lebanoff, Dernoncourt, Kim, Chang,\ \BBA\ Liu}{Lebanoff
  et~al.}{2020}]{lebanoff-etal-2020-cascade}
Lebanoff, L., Dernoncourt, F., Kim, D.~S., Chang, W., \BBA\ Liu, F.
  \BBOP2020\BBCP.
\newblock \BBOQ A cascade approach to neural abstractive summarization with
  content selection and fusion\BBCQ\
\newblock In {\Bem Proceedings of the 1st Conference of the Asia-Pacific
  Chapter of the Association for Computational Linguistics and the 10th
  International Joint Conference on Natural Language Processing}, \BPGS\
  529--535, Suzhou, China. Association for Computational Linguistics.

\bibitem[\protect\BCAY{Lebanoff, Song, Dernoncourt, Kim, Kim, Chang,\ \BBA\
  Liu}{Lebanoff et~al.}{2019}]{lebanoff-etal-2019-scoring}
Lebanoff, L., Song, K., Dernoncourt, F., Kim, D.~S., Kim, S., Chang, W., \BBA\
  Liu, F. \BBOP2019\BBCP.
\newblock \BBOQ Scoring sentence singletons and pairs for abstractive
  summarization\BBCQ\
\newblock In {\Bem Proceedings of the 57th Annual Meeting of the Association
  for Computational Linguistics}, \BPGS\ 2175--2189, Florence, Italy.
  Association for Computational Linguistics.

\bibitem[\protect\BCAY{Lee, Shin, Whang, Cho, Ko, Lee, Kim,\ \BBA\ Jo}{Lee
  et~al.}{2020}]{lee-etal-2020-reference}
Lee, D., Shin, M.~C., Whang, T., Cho, S., Ko, B., Lee, D., Kim, E., \BBA\ Jo,
  J. \BBOP2020\BBCP.
\newblock \BBOQ Reference and document aware semantic evaluation methods for
  {K}orean language summarization\BBCQ\
\newblock In {\Bem Proceedings of the 28th International Conference on
  Computational Linguistics}, \BPGS\ 5604--5616, Barcelona, Spain (Online).
  International Committee on Computational Linguistics.

\bibitem[\protect\BCAY{Lev, Shmueli-Scheuer, Herzig, Jerbi,\ \BBA\
  Konopnicki}{Lev et~al.}{2019}]{lev-etal-2019-talksumm}
Lev, G., Shmueli-Scheuer, M., Herzig, J., Jerbi, A., \BBA\ Konopnicki, D.
  \BBOP2019\BBCP.
\newblock \BBOQ {T}alk{S}umm: A dataset and scalable annotation method for
  scientific paper summarization based on conference talks\BBCQ\
\newblock In {\Bem Proceedings of the 57th Annual Meeting of the Association
  for Computational Linguistics}, \BPGS\ 2125--2131, Florence, Italy.
  Association for Computational Linguistics.

\bibitem[\protect\BCAY{{Lewis}, {Liu}, {Goyal}, {Ghazvininejad}, {Mohamed},
  {Levy}, {Stoyanov},\ \BBA\ {Zettlemoyer}}{{Lewis}
  et~al.}{2020}]{lewis2019bart}
{Lewis}, M., {Liu}, Y., {Goyal}, N., {Ghazvininejad}, M., {Mohamed}, A.,
  {Levy}, O., {Stoyanov}, V., \BBA\ {Zettlemoyer}, L. \BBOP2020\BBCP.
\newblock \BBOQ {BART}: Denoising sequence-to-sequence pre-training for natural
  language generation, translation, and comprehension\BBCQ\
\newblock In {\Bem Proceedings of the 58th Annual Meeting of the Association
  for Computational Linguistics}, \BPGS\ 7871--7880.

\bibitem[\protect\BCAY{Li, Xu, Li,\ \BBA\ Gao}{Li
  et~al.}{2018a}]{li-etal-2018-guiding}
Li, C., Xu, W., Li, S., \BBA\ Gao, S. \BBOP2018a\BBCP.
\newblock \BBOQ Guiding generation for abstractive text summarization based on
  key information guide network\BBCQ\
\newblock In {\Bem Proceedings of the 2018 Conference of the North {A}merican
  Chapter of the Association for Computational Linguistics: Human Language
  Technologies, Volume 2 (Short Papers)}, \BPGS\ 55--60, New Orleans,
  Louisiana. Association for Computational Linguistics.

\bibitem[\protect\BCAY{Li, Zhu, Zhang,\ \BBA\ Zong}{Li
  et~al.}{2018b}]{li-etal-2018-ensure}
Li, H., Zhu, J., Zhang, J., \BBA\ Zong, C. \BBOP2018b\BBCP.
\newblock \BBOQ Ensure the correctness of the summary: Incorporate entailment
  knowledge into abstractive sentence summarization\BBCQ\
\newblock In {\Bem Proceedings of the 27th International Conference on
  Computational Linguistics}, \BPGS\ 1430--1441, Santa Fe, New Mexico, USA.
  Association for Computational Linguistics.

\bibitem[\protect\BCAY{Li, Zhang, Ji,\ \BBA\ Radke}{Li
  et~al.}{2019a}]{li-etal-2019-keep}
Li, M., Zhang, L., Ji, H., \BBA\ Radke, R.~J. \BBOP2019a\BBCP.
\newblock \BBOQ Keep meeting summaries on topic: Abstractive multi-modal
  meeting summarization\BBCQ\
\newblock In {\Bem Proceedings of the 57th Annual Meeting of the Association
  for Computational Linguistics}, \BPGS\ 2190--2196, Florence, Italy.
  Association for Computational Linguistics.

\bibitem[\protect\BCAY{Li, Lei, Qin,\ \BBA\ Wang}{Li
  et~al.}{2019b}]{li-etal-2019-deep}
Li, S., Lei, D., Qin, P., \BBA\ Wang, W.~Y. \BBOP2019b\BBCP.
\newblock \BBOQ Deep reinforcement learning with distributional semantic
  rewards for abstractive summarization\BBCQ\
\newblock In {\Bem Proceedings of the 2019 Conference on Empirical Methods in
  Natural Language Processing and the 9th International Joint Conference on
  Natural Language Processing (EMNLP-IJCNLP)}, \BPGS\ 6038--6044, Hong Kong,
  China. Association for Computational Linguistics.

\bibitem[\protect\BCAY{Li, Wu,\ \BBA\ Li}{Li
  et~al.}{2020}]{li-etal-2020-composing}
Li, Z., Wu, W., \BBA\ Li, S. \BBOP2020\BBCP.
\newblock \BBOQ Composing elementary discourse units in abstractive
  summarization\BBCQ\
\newblock In {\Bem Proceedings of the 58th Annual Meeting of the Association
  for Computational Linguistics}, \BPGS\ 6191--6196, Online. Association for
  Computational Linguistics.

\bibitem[\protect\BCAY{Liao, Lebanoff,\ \BBA\ Liu}{Liao
  et~al.}{2018}]{liao-etal-2018-abstract}
Liao, K., Lebanoff, L., \BBA\ Liu, F. \BBOP2018\BBCP.
\newblock \BBOQ {A}bstract {M}eaning {R}epresentation for multi-document
  summarization\BBCQ\
\newblock In {\Bem Proceedings of the 27th International Conference on
  Computational Linguistics}, \BPGS\ 1178--1190, Santa Fe, New Mexico, USA.
  Association for Computational Linguistics.

\bibitem[\protect\BCAY{Lin}{Lin}{2004}]{lin-2004-rouge}
Lin, C.-Y. \BBOP2004\BBCP.
\newblock \BBOQ {ROUGE}: A package for automatic evaluation of summaries\BBCQ\
\newblock In {\Bem Text Summarization Branches Out}, \BPGS\ 74--81, Barcelona,
  Spain. Association for Computational Linguistics.

\bibitem[\protect\BCAY{Lin, Sun, Ma,\ \BBA\ Su}{Lin
  et~al.}{2018}]{lin-etal-2018-global}
Lin, J., Sun, X., Ma, S., \BBA\ Su, Q. \BBOP2018\BBCP.
\newblock \BBOQ Global encoding for abstractive summarization\BBCQ\
\newblock In {\Bem Proceedings of the 56th Annual Meeting of the Association
  for Computational Linguistics (Volume 2: Short Papers)}, \BPGS\ 163--169,
  Melbourne, Australia. Association for Computational Linguistics.

\bibitem[\protect\BCAY{Liu\ \BBA\ Lapata}{Liu\ \BBA\
  Lapata}{2019}]{liu-lapata-2019-text}
Liu, Y.\BBACOMMA\  \BBA\ Lapata, M. \BBOP2019\BBCP.
\newblock \BBOQ Text summarization with pretrained encoders\BBCQ\
\newblock In {\Bem Proceedings of the 2019 Conference on Empirical Methods in
  Natural Language Processing and the 9th International Joint Conference on
  Natural Language Processing (EMNLP-IJCNLP)}, \BPGS\ 3730--3740, Hong Kong,
  China. Association for Computational Linguistics.

\bibitem[\protect\BCAY{Liu, Titov,\ \BBA\ Lapata}{Liu
  et~al.}{2019a}]{liu-etal-2019-single}
Liu, Y., Titov, I., \BBA\ Lapata, M. \BBOP2019a\BBCP.
\newblock \BBOQ Single document summarization as tree induction\BBCQ\
\newblock In {\Bem Proceedings of the 2019 Conference of the North {A}merican
  Chapter of the Association for Computational Linguistics: Human Language
  Technologies, Volume 1 (Long and Short Papers)}, \BPGS\ 1745--1755,
  Minneapolis, Minnesota. Association for Computational Linguistics.

\bibitem[\protect\BCAY{Liu, Ott, Goyal, Du, Joshi, Chen, Levy, Lewis,
  Zettlemoyer,\ \BBA\ Stoyanov}{Liu et~al.}{2019b}]{liu2019roberta}
Liu, Y., Ott, M., Goyal, N., Du, J., Joshi, M., Chen, D., Levy, O., Lewis, M.,
  Zettlemoyer, L., \BBA\ Stoyanov, V. \BBOP2019b\BBCP.
\newblock \BBOQ {RoBERTa}: A robustly optimized {BERT} pretraining
  approach\BBCQ\
\newblock {\Bem arXiv preprint arXiv:1907.11692}.

\bibitem[\protect\BCAY{Lu, Dong,\ \BBA\ Charlin}{Lu
  et~al.}{2020}]{lu-etal-2020-multi-xscience}
Lu, Y., Dong, Y., \BBA\ Charlin, L. \BBOP2020\BBCP.
\newblock \BBOQ Multi-{XS}cience: A large-scale dataset for extreme
  multi-document summarization of scientific articles\BBCQ\
\newblock In {\Bem Proceedings of the 2020 Conference on Empirical Methods in
  Natural Language Processing (EMNLP)}, \BPGS\ 8068--8074, Online. Association
  for Computational Linguistics.

\bibitem[\protect\BCAY{Luo, Ao, Song, Pan, Yang,\ \BBA\ He}{Luo
  et~al.}{2019}]{luo-etal-2019-reading}
Luo, L., Ao, X., Song, Y., Pan, F., Yang, M., \BBA\ He, Q. \BBOP2019\BBCP.
\newblock \BBOQ Reading like {HER}: Human reading inspired extractive
  summarization\BBCQ\
\newblock In {\Bem Proceedings of the 2019 Conference on Empirical Methods in
  Natural Language Processing and the 9th International Joint Conference on
  Natural Language Processing (EMNLP-IJCNLP)}, \BPGS\ 3033--3043, Hong Kong,
  China. Association for Computational Linguistics.

\bibitem[\protect\BCAY{Makino, Iwakura, Takamura,\ \BBA\ Okumura}{Makino
  et~al.}{2019}]{makino-etal-2019-global}
Makino, T., Iwakura, T., Takamura, H., \BBA\ Okumura, M. \BBOP2019\BBCP.
\newblock \BBOQ Global optimization under length constraint for neural text
  summarization\BBCQ\
\newblock In {\Bem Proceedings of the 57th Annual Meeting of the Association
  for Computational Linguistics}, \BPGS\ 1039--1048, Florence, Italy.
  Association for Computational Linguistics.

\bibitem[\protect\BCAY{Mann}{Mann}{1984}]{mann-1984-discourse}
Mann, W.~C. \BBOP1984\BBCP.
\newblock \BBOQ Discourse structures for text generation\BBCQ\
\newblock In {\Bem 10th International Conference on Computational Linguistics
  and 22nd Annual Meeting of the Association for Computational Linguistics},
  \BPGS\ 367--375, Stanford, California, USA. Association for Computational
  Linguistics.

\bibitem[\protect\BCAY{Mao, Liu, Zhu, Ren,\ \BBA\ Han}{Mao
  et~al.}{2020a}]{mao-etal-2020-facet}
Mao, Y., Liu, L., Zhu, Q., Ren, X., \BBA\ Han, J. \BBOP2020a\BBCP.
\newblock \BBOQ Facet-aware evaluation for extractive summarization\BBCQ\
\newblock In {\Bem Proceedings of the 58th Annual Meeting of the Association
  for Computational Linguistics}, \BPGS\ 4941--4957, Online. Association for
  Computational Linguistics.

\bibitem[\protect\BCAY{Mao, Qu, Xie, Ren,\ \BBA\ Han}{Mao
  et~al.}{2020b}]{mao-etal-2020-multi}
Mao, Y., Qu, Y., Xie, Y., Ren, X., \BBA\ Han, J. \BBOP2020b\BBCP.
\newblock \BBOQ Multi-document summarization with maximal marginal
  relevance-guided reinforcement learning\BBCQ\
\newblock In {\Bem Proceedings of the 2020 Conference on Empirical Methods in
  Natural Language Processing (EMNLP)}, \BPGS\ 1737--1751, Online. Association
  for Computational Linguistics.

\bibitem[\protect\BCAY{Mathur, Baldwin,\ \BBA\ Cohn}{Mathur
  et~al.}{2019}]{mathur-etal-2019-putting}
Mathur, N., Baldwin, T., \BBA\ Cohn, T. \BBOP2019\BBCP.
\newblock \BBOQ Putting evaluation in context: Contextual embeddings improve
  machine translation evaluation\BBCQ\
\newblock In {\Bem Proceedings of the 57th Annual Meeting of the Association
  for Computational Linguistics}, \BPGS\ 2799--2808, Florence, Italy.
  Association for Computational Linguistics.

\bibitem[\protect\BCAY{Maynez, Narayan, Bohnet,\ \BBA\ McDonald}{Maynez
  et~al.}{2020}]{maynez-etal-2020-faithfulness}
Maynez, J., Narayan, S., Bohnet, B., \BBA\ McDonald, R. \BBOP2020\BBCP.
\newblock \BBOQ On faithfulness and factuality in abstractive
  summarization\BBCQ\
\newblock In {\Bem Proceedings of the 58th Annual Meeting of the Association
  for Computational Linguistics}, \BPGS\ 1906--1919, Online. Association for
  Computational Linguistics.

\bibitem[\protect\BCAY{Mendes, Narayan, Miranda, Marinho, Martins,\ \BBA\
  Cohen}{Mendes et~al.}{2019}]{mendes-etal-2019-jointly}
Mendes, A., Narayan, S., Miranda, S., Marinho, Z., Martins, A. F.~T., \BBA\
  Cohen, S.~B. \BBOP2019\BBCP.
\newblock \BBOQ Jointly extracting and compressing documents with summary state
  representations\BBCQ\
\newblock In {\Bem Proceedings of the 2019 Conference of the North {A}merican
  Chapter of the Association for Computational Linguistics: Human Language
  Technologies, Volume 1 (Long and Short Papers)}, \BPGS\ 3955--3966,
  Minneapolis, Minnesota. Association for Computational Linguistics.

\bibitem[\protect\BCAY{Mikolov, Yih,\ \BBA\ Zweig}{Mikolov
  et~al.}{2013}]{mikolov-etal-2013-linguistic}
Mikolov, T., Yih, W.-t., \BBA\ Zweig, G. \BBOP2013\BBCP.
\newblock \BBOQ Linguistic regularities in continuous space word
  representations\BBCQ\
\newblock In {\Bem Proceedings of the 2013 Conference of the North {A}merican
  Chapter of the Association for Computational Linguistics: Human Language
  Technologies}, \BPGS\ 746--751, Atlanta, Georgia. Association for
  Computational Linguistics.

\bibitem[\protect\BCAY{Nallapati, Zhou, dos Santos, Gul\c{c}ehre,\ \BBA\
  Xiang}{Nallapati et~al.}{2016}]{nallapati-etal-2016-abstractive}
Nallapati, R., Zhou, B., dos Santos, C., Gul\c{c}ehre, {\c{C}}., \BBA\ Xiang,
  B. \BBOP2016\BBCP.
\newblock \BBOQ Abstractive text summarization using sequence-to-sequence
  {RNN}s and beyond\BBCQ\
\newblock In {\Bem Proceedings of The 20th {SIGNLL} Conference on Computational
  Natural Language Learning}, \BPGS\ 280--290, Berlin, Germany. Association for
  Computational Linguistics.

\bibitem[\protect\BCAY{Narayan, Cardenas, Papasarantopoulos, Cohen, Lapata,
  Yu,\ \BBA\ Chang}{Narayan et~al.}{2018a}]{narayan-etal-2018-document}
Narayan, S., Cardenas, R., Papasarantopoulos, N., Cohen, S.~B., Lapata, M., Yu,
  J., \BBA\ Chang, Y. \BBOP2018a\BBCP.
\newblock \BBOQ Document modeling with external attention for sentence
  extraction\BBCQ\
\newblock In {\Bem Proceedings of the 56th Annual Meeting of the Association
  for Computational Linguistics (Volume 1: Long Papers)}, \BPGS\ 2020--2030,
  Melbourne, Australia. Association for Computational Linguistics.

\bibitem[\protect\BCAY{Narayan, Cohen,\ \BBA\ Lapata}{Narayan
  et~al.}{2018b}]{narayan-etal-2018-dont}
Narayan, S., Cohen, S.~B., \BBA\ Lapata, M. \BBOP2018b\BBCP.
\newblock \BBOQ Don{'}t give me the details, just the summary! topic-aware
  convolutional neural networks for extreme summarization\BBCQ\
\newblock In {\Bem Proceedings of the 2018 Conference on Empirical Methods in
  Natural Language Processing}, \BPGS\ 1797--1807, Brussels, Belgium.
  Association for Computational Linguistics.

\bibitem[\protect\BCAY{Narayan, Cohen,\ \BBA\ Lapata}{Narayan
  et~al.}{2018c}]{narayan-etal-2018-ranking}
Narayan, S., Cohen, S.~B., \BBA\ Lapata, M. \BBOP2018c\BBCP.
\newblock \BBOQ Ranking sentences for extractive summarization with
  reinforcement learning\BBCQ\
\newblock In {\Bem Proceedings of the 2018 Conference of the North {A}merican
  Chapter of the Association for Computational Linguistics: Human Language
  Technologies, Volume 1 (Long Papers)}, \BPGS\ 1747--1759, New Orleans,
  Louisiana. Association for Computational Linguistics.

\bibitem[\protect\BCAY{Nayeem\ \BBA\ Chali}{Nayeem\ \BBA\
  Chali}{2017}]{nayeem-chali-2017-extract}
Nayeem, M.~T.\BBACOMMA\  \BBA\ Chali, Y. \BBOP2017\BBCP.
\newblock \BBOQ Extract with order for coherent multi-document
  summarization\BBCQ\
\newblock In {\Bem Proceedings of {T}ext{G}raphs-11: the Workshop on
  Graph-based Methods for Natural Language Processing}, \BPGS\ 51--56,
  Vancouver, Canada. Association for Computational Linguistics.

\bibitem[\protect\BCAY{Nenkova\ \BBA\ Passonneau}{Nenkova\ \BBA\
  Passonneau}{2004}]{nenkova-passonneau-2004-evaluating}
Nenkova, A.\BBACOMMA\  \BBA\ Passonneau, R. \BBOP2004\BBCP.
\newblock \BBOQ Evaluating content selection in summarization: The pyramid
  method\BBCQ\
\newblock In {\Bem Proceedings of the Human Language Technology Conference of
  the North {A}merican Chapter of the Association for Computational
  Linguistics: {HLT}-{NAACL} 2004}, \BPGS\ 145--152, Boston, Massachusetts,
  USA. Association for Computational Linguistics.

\bibitem[\protect\BCAY{Ng\ \BBA\ Abrecht}{Ng\ \BBA\
  Abrecht}{2015}]{ng-abrecht-2015-better}
Ng, J.-P.\BBACOMMA\  \BBA\ Abrecht, V. \BBOP2015\BBCP.
\newblock \BBOQ Better summarization evaluation with word embeddings for
  {ROUGE}\BBCQ\
\newblock In {\Bem Proceedings of the 2015 Conference on Empirical Methods in
  Natural Language Processing}, \BPGS\ 1925--1930, Lisbon, Portugal.
  Association for Computational Linguistics.

\bibitem[\protect\BCAY{Ouyang, Song,\ \BBA\ McKeown}{Ouyang
  et~al.}{2019}]{ouyang-etal-2019-robust}
Ouyang, J., Song, B., \BBA\ McKeown, K. \BBOP2019\BBCP.
\newblock \BBOQ A robust abstractive system for cross-lingual
  summarization\BBCQ\
\newblock In {\Bem Proceedings of the 2019 Conference of the North {A}merican
  Chapter of the Association for Computational Linguistics: Human Language
  Technologies, Volume 1 (Long and Short Papers)}, \BPGS\ 2025--2031,
  Minneapolis, Minnesota. Association for Computational Linguistics.

\bibitem[\protect\BCAY{Pagnoni, Balachandran,\ \BBA\ Tsvetkov}{Pagnoni
  et~al.}{2021}]{pagnoni-etal-2021-understanding}
Pagnoni, A., Balachandran, V., \BBA\ Tsvetkov, Y. \BBOP2021\BBCP.
\newblock \BBOQ Understanding factuality in abstractive summarization with
  {FRANK}: A benchmark for factuality metrics\BBCQ\
\newblock In {\Bem Proceedings of the 2021 Conference of the North American
  Chapter of the Association for Computational Linguistics: Human Language
  Technologies}, \BPGS\ 4812--4829, Online. Association for Computational
  Linguistics.

\bibitem[\protect\BCAY{Palaskar, Libovick{\'y}, Gella,\ \BBA\ Metze}{Palaskar
  et~al.}{2019}]{palaskar-etal-2019-multimodal}
Palaskar, S., Libovick{\'y}, J., Gella, S., \BBA\ Metze, F. \BBOP2019\BBCP.
\newblock \BBOQ Multimodal abstractive summarization for how2 videos\BBCQ\
\newblock In {\Bem Proceedings of the 57th Annual Meeting of the Association
  for Computational Linguistics}, \BPGS\ 6587--6596, Florence, Italy.
  Association for Computational Linguistics.

\bibitem[\protect\BCAY{Papalampidi, Keller, Frermann,\ \BBA\
  Lapata}{Papalampidi et~al.}{2020}]{papalampidi-etal-2020-screenplay}
Papalampidi, P., Keller, F., Frermann, L., \BBA\ Lapata, M. \BBOP2020\BBCP.
\newblock \BBOQ Screenplay summarization using latent narrative structure\BBCQ\
\newblock In {\Bem Proceedings of the 58th Annual Meeting of the Association
  for Computational Linguistics}, \BPGS\ 1920--1933, Online. Association for
  Computational Linguistics.

\bibitem[\protect\BCAY{Papalampidi, Keller,\ \BBA\ Lapata}{Papalampidi
  et~al.}{2019}]{papalampidi-etal-2019-movie}
Papalampidi, P., Keller, F., \BBA\ Lapata, M. \BBOP2019\BBCP.
\newblock \BBOQ Movie plot analysis via turning point identification\BBCQ\
\newblock In {\Bem Proceedings of the 2019 Conference on Empirical Methods in
  Natural Language Processing and the 9th International Joint Conference on
  Natural Language Processing (EMNLP-IJCNLP)}, \BPGS\ 1707--1717, Hong Kong,
  China. Association for Computational Linguistics.

\bibitem[\protect\BCAY{Papineni, Roukos, Ward,\ \BBA\ Zhu}{Papineni
  et~al.}{2002}]{papineni-etal-2002-bleu}
Papineni, K., Roukos, S., Ward, T., \BBA\ Zhu, W.-J. \BBOP2002\BBCP.
\newblock \BBOQ {B}leu: a method for automatic evaluation of machine
  translation\BBCQ\
\newblock In {\Bem Proceedings of the 40th Annual Meeting of the Association
  for Computational Linguistics}, \BPGS\ 311--318, Philadelphia, Pennsylvania,
  USA. Association for Computational Linguistics.

\bibitem[\protect\BCAY{Parida\ \BBA\ Motlicek}{Parida\ \BBA\
  Motlicek}{2019}]{parida-motlicek-2019-abstract}
Parida, S.\BBACOMMA\  \BBA\ Motlicek, P. \BBOP2019\BBCP.
\newblock \BBOQ Abstract text summarization: A low resource challenge\BBCQ\
\newblock In {\Bem Proceedings of the 2019 Conference on Empirical Methods in
  Natural Language Processing and the 9th International Joint Conference on
  Natural Language Processing (EMNLP-IJCNLP)}, \BPGS\ 5994--5998, Hong Kong,
  China. Association for Computational Linguistics.

\bibitem[\protect\BCAY{Pasunuru\ \BBA\ Bansal}{Pasunuru\ \BBA\
  Bansal}{2018}]{pasunuru-bansal-2018-multi}
Pasunuru, R.\BBACOMMA\  \BBA\ Bansal, M. \BBOP2018\BBCP.
\newblock \BBOQ Multi-reward reinforced summarization with saliency and
  entailment\BBCQ\
\newblock In {\Bem Proceedings of the 2018 Conference of the North {A}merican
  Chapter of the Association for Computational Linguistics: Human Language
  Technologies, Volume 2 (Short Papers)}, \BPGS\ 646--653, New Orleans,
  Louisiana. Association for Computational Linguistics.

\bibitem[\protect\BCAY{Pennington, Socher,\ \BBA\ Manning}{Pennington
  et~al.}{2014}]{pennington-etal-2014-glove}
Pennington, J., Socher, R., \BBA\ Manning, C. \BBOP2014\BBCP.
\newblock \BBOQ {G}lo{V}e: Global vectors for word representation\BBCQ\
\newblock In {\Bem Proceedings of the 2014 Conference on Empirical Methods in
  Natural Language Processing ({EMNLP})}, \BPGS\ 1532--1543, Doha, Qatar.
  Association for Computational Linguistics.

\bibitem[\protect\BCAY{Peyrard}{Peyrard}{2019a}]{peyrard-2019-simple}
Peyrard, M. \BBOP2019a\BBCP.
\newblock \BBOQ A simple theoretical model of importance for
  summarization\BBCQ\
\newblock In {\Bem Proceedings of the 57th Annual Meeting of the Association
  for Computational Linguistics}, \BPGS\ 1059--1073, Florence, Italy.
  Association for Computational Linguistics.

\bibitem[\protect\BCAY{Peyrard}{Peyrard}{2019b}]{peyrard-2019-studying}
Peyrard, M. \BBOP2019b\BBCP.
\newblock \BBOQ Studying summarization evaluation metrics in the appropriate
  scoring range\BBCQ\
\newblock In {\Bem Proceedings of the 57th Annual Meeting of the Association
  for Computational Linguistics}, \BPGS\ 5093--5100, Florence, Italy.
  Association for Computational Linguistics.

\bibitem[\protect\BCAY{Peyrard\ \BBA\ Gurevych}{Peyrard\ \BBA\
  Gurevych}{2018}]{peyrard-gurevych-2018-objective}
Peyrard, M.\BBACOMMA\  \BBA\ Gurevych, I. \BBOP2018\BBCP.
\newblock \BBOQ Objective function learning to match human judgements for
  optimization-based summarization\BBCQ\
\newblock In {\Bem Proceedings of the 2018 Conference of the North {A}merican
  Chapter of the Association for Computational Linguistics: Human Language
  Technologies, Volume 2 (Short Papers)}, \BPGS\ 654--660, New Orleans,
  Louisiana. Association for Computational Linguistics.

\bibitem[\protect\BCAY{Pilault, Li, Subramanian,\ \BBA\ Pal}{Pilault
  et~al.}{2020}]{pilault-etal-2020-extractive}
Pilault, J., Li, R., Subramanian, S., \BBA\ Pal, C. \BBOP2020\BBCP.
\newblock \BBOQ On extractive and abstractive neural document summarization
  with transformer language models\BBCQ\
\newblock In {\Bem Proceedings of the 2020 Conference on Empirical Methods in
  Natural Language Processing (EMNLP)}, \BPGS\ 9308--9319, Online. Association
  for Computational Linguistics.

\bibitem[\protect\BCAY{Post}{Post}{2018}]{post-2018-call}
Post, M. \BBOP2018\BBCP.
\newblock \BBOQ A call for clarity in reporting {BLEU} scores\BBCQ\
\newblock In {\Bem Proceedings of the Third Conference on Machine Translation:
  Research Papers}, \BPGS\ 186--191, Brussels, Belgium. Association for
  Computational Linguistics.

\bibitem[\protect\BCAY{Radford, Wu, Child, Luan, Amodei,\ \BBA\
  Sutskever}{Radford et~al.}{2019}]{radford2019language}
Radford, A., Wu, J., Child, R., Luan, D., Amodei, D., \BBA\ Sutskever, I.
  \BBOP2019\BBCP.
\newblock \BBOQ Language models are unsupervised multitask learners\BBCQ\
\newblock {\Bem OpenAI Blog}, {\Bem 1\/}(8), 9.

\bibitem[\protect\BCAY{{Raffel}, {Shazeer}, {Roberts}, {Lee}, {Narang},
  {Matena}, {Zhou}, {Li},\ \BBA\ {Liu}}{{Raffel}
  et~al.}{2019}]{raffel2019exploring}
{Raffel}, C., {Shazeer}, N., {Roberts}, A., {Lee}, K., {Narang}, S., {Matena},
  M., {Zhou}, Y., {Li}, W., \BBA\ {Liu}, P.~J. \BBOP2019\BBCP.
\newblock \BBOQ Exploring the limits of transfer learning with a unified
  text-to-text transformer\BBCQ\
\newblock {\Bem Journal of Machine Learning Research}, {\Bem 21\/}(140), 1--67.

\bibitem[\protect\BCAY{Reimers\ \BBA\ Gurevych}{Reimers\ \BBA\
  Gurevych}{2019a}]{reimers-gurevych-2019-sentence}
Reimers, N.\BBACOMMA\  \BBA\ Gurevych, I. \BBOP2019a\BBCP.
\newblock \BBOQ Sentence-{BERT}: Sentence embeddings using {S}iamese
  {BERT}-networks\BBCQ\
\newblock In {\Bem Proceedings of the 2019 Conference on Empirical Methods in
  Natural Language Processing and the 9th International Joint Conference on
  Natural Language Processing (EMNLP-IJCNLP)}, \BPGS\ 3982--3992, Hong Kong,
  China. Association for Computational Linguistics.

\bibitem[\protect\BCAY{Reimers\ \BBA\ Gurevych}{Reimers\ \BBA\
  Gurevych}{2019b}]{reimers-2019-sentence-bert}
Reimers, N.\BBACOMMA\  \BBA\ Gurevych, I. \BBOP2019b\BBCP.
\newblock \BBOQ Sentence-bert: Sentence embeddings using siamese
  bert-networks\BBCQ\
\newblock In {\Bem Proceedings of the 2019 Conference on Empirical Methods in
  Natural Language Processing}. Association for Computational Linguistics.

\bibitem[\protect\BCAY{Sakaue, Hirao, Nishino,\ \BBA\ Nagata}{Sakaue
  et~al.}{2018}]{sakaue-etal-2018-provable}
Sakaue, S., Hirao, T., Nishino, M., \BBA\ Nagata, M. \BBOP2018\BBCP.
\newblock \BBOQ Provable fast greedy compressive summarization with any
  monotone submodular function\BBCQ\
\newblock In {\Bem Proceedings of the 2018 Conference of the North {A}merican
  Chapter of the Association for Computational Linguistics: Human Language
  Technologies, Volume 1 (Long Papers)}, \BPGS\ 1737--1746, New Orleans,
  Louisiana. Association for Computational Linguistics.

\bibitem[\protect\BCAY{Schumann, Mou, Lu, Vechtomova,\ \BBA\ Markert}{Schumann
  et~al.}{2020}]{schumann-etal-2020-discrete}
Schumann, R., Mou, L., Lu, Y., Vechtomova, O., \BBA\ Markert, K.
  \BBOP2020\BBCP.
\newblock \BBOQ Discrete optimization for unsupervised sentence summarization
  with word-level extraction\BBCQ\
\newblock In {\Bem Proceedings of the 58th Annual Meeting of the Association
  for Computational Linguistics}, \BPGS\ 5032--5042, Online. Association for
  Computational Linguistics.

\bibitem[\protect\BCAY{Scialom, Dray, Lamprier, Piwowarski,\ \BBA\
  Staiano}{Scialom et~al.}{2020}]{scialom-etal-2020-mlsum}
Scialom, T., Dray, P.-A., Lamprier, S., Piwowarski, B., \BBA\ Staiano, J.
  \BBOP2020\BBCP.
\newblock \BBOQ {MLSUM}: The multilingual summarization corpus\BBCQ\
\newblock In {\Bem Proceedings of the 2020 Conference on Empirical Methods in
  Natural Language Processing (EMNLP)}, \BPGS\ 8051--8067, Online. Association
  for Computational Linguistics.

\bibitem[\protect\BCAY{See, Liu,\ \BBA\ Manning}{See
  et~al.}{2017}]{see-etal-2017-get}
See, A., Liu, P.~J., \BBA\ Manning, C.~D. \BBOP2017\BBCP.
\newblock \BBOQ Get to the point: Summarization with pointer-generator
  networks\BBCQ\
\newblock In {\Bem Proceedings of the 55th Annual Meeting of the Association
  for Computational Linguistics (Volume 1: Long Papers)}, \BPGS\ 1073--1083,
  Vancouver, Canada. Association for Computational Linguistics.

\bibitem[\protect\BCAY{ShafieiBavani, Ebrahimi, Wong,\ \BBA\
  Chen}{ShafieiBavani et~al.}{2018}]{shafieibavani-etal-2018-summarization}
ShafieiBavani, E., Ebrahimi, M., Wong, R., \BBA\ Chen, F. \BBOP2018\BBCP.
\newblock \BBOQ Summarization evaluation in the absence of human model
  summaries using the compositionality of word embeddings\BBCQ\
\newblock In {\Bem Proceedings of the 27th International Conference on
  Computational Linguistics}, \BPGS\ 905--914, Santa Fe, New Mexico, USA.
  Association for Computational Linguistics.

\bibitem[\protect\BCAY{Shapira, Gabay, Gao, Ronen, Pasunuru, Bansal,
  Amsterdamer,\ \BBA\ Dagan}{Shapira
  et~al.}{2019}]{shapira-etal-2019-crowdsourcing}
Shapira, O., Gabay, D., Gao, Y., Ronen, H., Pasunuru, R., Bansal, M.,
  Amsterdamer, Y., \BBA\ Dagan, I. \BBOP2019\BBCP.
\newblock \BBOQ Crowdsourcing lightweight pyramids for manual summary
  evaluation\BBCQ\
\newblock In {\Bem Proceedings of the 2019 Conference of the North {A}merican
  Chapter of the Association for Computational Linguistics: Human Language
  Technologies, Volume 1 (Long and Short Papers)}, \BPGS\ 682--687,
  Minneapolis, Minnesota. Association for Computational Linguistics.

\bibitem[\protect\BCAY{Sharma, Li,\ \BBA\ Wang}{Sharma
  et~al.}{2019}]{sharma-etal-2019-bigpatent}
Sharma, E., Li, C., \BBA\ Wang, L. \BBOP2019\BBCP.
\newblock \BBOQ {BIGPATENT}: A large-scale dataset for abstractive and coherent
  summarization\BBCQ\
\newblock In {\Bem Proceedings of the 57th Annual Meeting of the Association
  for Computational Linguistics}, \BPGS\ 2204--2213, Florence, Italy.
  Association for Computational Linguistics.

\bibitem[\protect\BCAY{Shen\ \BBA\ Baldwin}{Shen\ \BBA\
  Baldwin}{2021}]{shen2021simple}
Shen, A.\BBACOMMA\  \BBA\ Baldwin, T. \BBOP2021\BBCP.
\newblock \BBOQ A simple yet effective method for sentence ordering\BBCQ\
\newblock In {\Bem Proceedings of the 22nd Annual Meeting of the Special
  Interest Group on Discourse and Dialogue}, \BPGS\ 154--160.

\bibitem[\protect\BCAY{Shen, Zhao, Su,\ \BBA\ Klakow}{Shen
  et~al.}{2019}]{shen-etal-2019-improving}
Shen, X., Zhao, Y., Su, H., \BBA\ Klakow, D. \BBOP2019\BBCP.
\newblock \BBOQ Improving latent alignment in text summarization by
  generalizing the pointer generator\BBCQ\
\newblock In {\Bem Proceedings of the 2019 Conference on Empirical Methods in
  Natural Language Processing and the 9th International Joint Conference on
  Natural Language Processing (EMNLP-IJCNLP)}, \BPGS\ 3762--3773, Hong Kong,
  China. Association for Computational Linguistics.

\bibitem[\protect\BCAY{Song, Zhao,\ \BBA\ Liu}{Song
  et~al.}{2018}]{song-etal-2018-structure}
Song, K., Zhao, L., \BBA\ Liu, F. \BBOP2018\BBCP.
\newblock \BBOQ Structure-infused copy mechanisms for abstractive
  summarization\BBCQ\
\newblock In {\Bem Proceedings of the 27th International Conference on
  Computational Linguistics}, \BPGS\ 1717--1729, Santa Fe, New Mexico, USA.
  Association for Computational Linguistics.

\bibitem[\protect\BCAY{Sotudeh~Gharebagh, Goharian,\ \BBA\
  Filice}{Sotudeh~Gharebagh et~al.}{2020}]{sotudeh-gharebagh-etal-2020-attend}
Sotudeh~Gharebagh, S., Goharian, N., \BBA\ Filice, R. \BBOP2020\BBCP.
\newblock \BBOQ Attend to medical ontologies: Content selection for clinical
  abstractive summarization\BBCQ\
\newblock In {\Bem Proceedings of the 58th Annual Meeting of the Association
  for Computational Linguistics}, \BPGS\ 1899--1905, Online. Association for
  Computational Linguistics.

\bibitem[\protect\BCAY{Suhara, Wang, Angelidis,\ \BBA\ Tan}{Suhara
  et~al.}{2020}]{suhara-etal-2020-opiniondigest}
Suhara, Y., Wang, X., Angelidis, S., \BBA\ Tan, W.-C. \BBOP2020\BBCP.
\newblock \BBOQ {O}pinion{D}igest: A simple framework for opinion
  summarization\BBCQ\
\newblock In {\Bem Proceedings of the 58th Annual Meeting of the Association
  for Computational Linguistics}, \BPGS\ 5789--5798, Online. Association for
  Computational Linguistics.

\bibitem[\protect\BCAY{Sun\ \BBA\ Nenkova}{Sun\ \BBA\
  Nenkova}{2019}]{sun-nenkova-2019-feasibility}
Sun, S.\BBACOMMA\  \BBA\ Nenkova, A. \BBOP2019\BBCP.
\newblock \BBOQ The feasibility of embedding based automatic evaluation for
  single document summarization\BBCQ\
\newblock In {\Bem Proceedings of the 2019 Conference on Empirical Methods in
  Natural Language Processing and the 9th International Joint Conference on
  Natural Language Processing (EMNLP-IJCNLP)}, \BPGS\ 1216--1221, Hong Kong,
  China. Association for Computational Linguistics.

\bibitem[\protect\BCAY{Tan, Qin, Xing,\ \BBA\ Hu}{Tan
  et~al.}{2020}]{tan-etal-2020-summarizing}
Tan, B., Qin, L., Xing, E., \BBA\ Hu, Z. \BBOP2020\BBCP.
\newblock \BBOQ Summarizing text on any aspects: A knowledge-informed
  weakly-supervised approach\BBCQ\
\newblock In {\Bem Proceedings of the 2020 Conference on Empirical Methods in
  Natural Language Processing (EMNLP)}, \BPGS\ 6301--6309, Online. Association
  for Computational Linguistics.

\bibitem[\protect\BCAY{{Trischler}, {Wang}, {Yuan}, {Harris}, {Sordoni},
  {Bachman},\ \BBA\ {Suleman}}{{Trischler} et~al.}{2017}]{trischler2017newsqa}
{Trischler}, A., {Wang}, T., {Yuan}, X., {Harris}, J., {Sordoni}, A.,
  {Bachman}, P., \BBA\ {Suleman}, K. \BBOP2017\BBCP.
\newblock \BBOQ Newsqa: A machine comprehension dataset\BBCQ\
\newblock In {\Bem Proceedings of the 2nd Workshop on Representation Learning
  for NLP}, \BPGS\ 191--200.

\bibitem[\protect\BCAY{{Vaswani}, {Shazeer}, {Parmar}, {Uszkoreit}, {Jones},
  {Gomez}, {Kaiser},\ \BBA\ {Polosukhin}}{{Vaswani}
  et~al.}{2017}]{vaswani2017attention}
{Vaswani}, A., {Shazeer}, N., {Parmar}, N., {Uszkoreit}, J., {Jones}, L.,
  {Gomez}, A.~N., {Kaiser}, L., \BBA\ {Polosukhin}, I. \BBOP2017\BBCP.
\newblock \BBOQ Attention is all you need\BBCQ\
\newblock In {\Bem Proceedings of the 31st International Conference on Neural
  Information Processing Systems}, \BPGS\ 5998--6008.

\bibitem[\protect\BCAY{{Vedantam}, {Zitnick},\ \BBA\ {Parikh}}{{Vedantam}
  et~al.}{2015}]{vedantam2015cider}
{Vedantam}, R., {Zitnick}, C.~L., \BBA\ {Parikh}, D. \BBOP2015\BBCP.
\newblock \BBOQ {CIDE}r: Consensus-based image description evaluation\BBCQ\
\newblock In {\Bem 2015 IEEE Conference on Computer Vision and Pattern
  Recognition (CVPR)}, \BPGS\ 4566--4575.

\bibitem[\protect\BCAY{{Wang}, {Cho},\ \BBA\ {Lewis}}{{Wang}
  et~al.}{2020}]{wang2020asking}
{Wang}, A., {Cho}, K., \BBA\ {Lewis}, M. \BBOP2020\BBCP.
\newblock \BBOQ Asking and answering questions to evaluate the factual
  consistency of summaries\BBCQ\
\newblock In {\Bem Proceedings of the 58th Annual Meeting of the Association
  for Computational Linguistics}, \BPGS\ 5008--5020. Association for
  Computational Linguistics.

\bibitem[\protect\BCAY{Wang, Wang, Xiong, Yu, Guo, Chang,\ \BBA\ Wang}{Wang
  et~al.}{2019a}]{wang-etal-2019-self-supervised}
Wang, H., Wang, X., Xiong, W., Yu, M., Guo, X., Chang, S., \BBA\ Wang, W.~Y.
  \BBOP2019a\BBCP.
\newblock \BBOQ Self-supervised learning for contextualized extractive
  summarization\BBCQ\
\newblock In {\Bem Proceedings of the 57th Annual Meeting of the Association
  for Computational Linguistics}, \BPGS\ 2221--2227, Florence, Italy.
  Association for Computational Linguistics.

\bibitem[\protect\BCAY{Wang, Quan,\ \BBA\ Wang}{Wang
  et~al.}{2019b}]{wang-etal-2019-biset}
Wang, K., Quan, X., \BBA\ Wang, R. \BBOP2019b\BBCP.
\newblock \BBOQ {B}i{SET}: Bi-directional selective encoding with template for
  abstractive summarization\BBCQ\
\newblock In {\Bem Proceedings of the 57th Annual Meeting of the Association
  for Computational Linguistics}, \BPGS\ 2153--2162, Florence, Italy.
  Association for Computational Linguistics.

\bibitem[\protect\BCAY{Wang, Chang,\ \BBA\ Sui}{Wang
  et~al.}{2020a}]{wang-etal-2020-spectral}
Wang, K., Chang, B., \BBA\ Sui, Z. \BBOP2020a\BBCP.
\newblock \BBOQ A spectral method for unsupervised multi-document
  summarization\BBCQ\
\newblock In {\Bem Proceedings of the 2020 Conference on Empirical Methods in
  Natural Language Processing (EMNLP)}, \BPGS\ 435--445, Online. Association
  for Computational Linguistics.

\bibitem[\protect\BCAY{Wang, Duan, Zhang, Wang, Tian, Chen,\ \BBA\ Zhou}{Wang
  et~al.}{2020b}]{wang-etal-2020-friendly}
Wang, Z., Duan, Z., Zhang, H., Wang, C., Tian, L., Chen, B., \BBA\ Zhou, M.
  \BBOP2020b\BBCP.
\newblock \BBOQ Friendly topic assistant for transformer based abstractive
  summarization\BBCQ\
\newblock In {\Bem Proceedings of the 2020 Conference on Empirical Methods in
  Natural Language Processing (EMNLP)}, \BPGS\ 485--497, Online. Association
  for Computational Linguistics.

\bibitem[\protect\BCAY{West, Holtzman, Buys,\ \BBA\ Choi}{West
  et~al.}{2019}]{west-etal-2019-bottlesum}
West, P., Holtzman, A., Buys, J., \BBA\ Choi, Y. \BBOP2019\BBCP.
\newblock \BBOQ {B}ottle{S}um: Unsupervised and self-supervised sentence
  summarization using the information bottleneck principle\BBCQ\
\newblock In {\Bem Proceedings of the 2019 Conference on Empirical Methods in
  Natural Language Processing and the 9th International Joint Conference on
  Natural Language Processing (EMNLP-IJCNLP)}, \BPGS\ 3752--3761, Hong Kong,
  China. Association for Computational Linguistics.

\bibitem[\protect\BCAY{Wu, Ma, Wu, Manyumwa,\ \BBA\ Ji}{Wu
  et~al.}{2020}]{wu-etal-2020-unsupervised}
Wu, H., Ma, T., Wu, L., Manyumwa, T., \BBA\ Ji, S. \BBOP2020\BBCP.
\newblock \BBOQ Unsupervised reference-free summary quality evaluation via
  contrastive learning\BBCQ\
\newblock In {\Bem Proceedings of the 2020 Conference on Empirical Methods in
  Natural Language Processing (EMNLP)}, \BPGS\ 3612--3621, Online. Association
  for Computational Linguistics.

\bibitem[\protect\BCAY{Xiao, Wang, He,\ \BBA\ Jin}{Xiao
  et~al.}{2020}]{xiao-etal-2020-modeling}
Xiao, L., Wang, L., He, H., \BBA\ Jin, Y. \BBOP2020\BBCP.
\newblock \BBOQ Modeling content importance for summarization with pre-trained
  language models\BBCQ\
\newblock In {\Bem Proceedings of the 2020 Conference on Empirical Methods in
  Natural Language Processing (EMNLP)}, \BPGS\ 3606--3611, Online. Association
  for Computational Linguistics.

\bibitem[\protect\BCAY{Xiao\ \BBA\ Carenini}{Xiao\ \BBA\
  Carenini}{2019}]{xiao-carenini-2019-extractive}
Xiao, W.\BBACOMMA\  \BBA\ Carenini, G. \BBOP2019\BBCP.
\newblock \BBOQ Extractive summarization of long documents by combining global
  and local context\BBCQ\
\newblock In {\Bem Proceedings of the 2019 Conference on Empirical Methods in
  Natural Language Processing and the 9th International Joint Conference on
  Natural Language Processing (EMNLP-IJCNLP)}, \BPGS\ 3011--3021, Hong Kong,
  China. Association for Computational Linguistics.

\bibitem[\protect\BCAY{Xiao\ \BBA\ Carenini}{Xiao\ \BBA\
  Carenini}{2020}]{xiao-carenini-2020-systematically}
Xiao, W.\BBACOMMA\  \BBA\ Carenini, G. \BBOP2020\BBCP.
\newblock \BBOQ Systematically exploring redundancy reduction in summarizing
  long documents\BBCQ\
\newblock In {\Bem Proceedings of the 1st Conference of the Asia-Pacific
  Chapter of the Association for Computational Linguistics and the 10th
  International Joint Conference on Natural Language Processing}, \BPGS\
  516--528, Suzhou, China. Association for Computational Linguistics.

\bibitem[\protect\BCAY{Xu, Gan, Cheng,\ \BBA\ Liu}{Xu
  et~al.}{2020}]{xu-etal-2020-discourse}
Xu, J., Gan, Z., Cheng, Y., \BBA\ Liu, J. \BBOP2020\BBCP.
\newblock \BBOQ Discourse-aware neural extractive text summarization\BBCQ\
\newblock In {\Bem Proceedings of the 58th Annual Meeting of the Association
  for Computational Linguistics}, \BPGS\ 5021--5031, Online. Association for
  Computational Linguistics.

\bibitem[\protect\BCAY{{Xu}, {Ba}, {Kiros}, {Cho}, {Courville}, {Salakhudinov},
  {Zemel},\ \BBA\ {Bengio}}{{Xu} et~al.}{2015}]{xu2015show}
{Xu}, K., {Ba}, J., {Kiros}, R., {Cho}, K., {Courville}, A., {Salakhudinov},
  R., {Zemel}, R., \BBA\ {Bengio}, Y. \BBOP2015\BBCP.
\newblock \BBOQ Show, attend and tell: Neural image caption generation with
  visual attention\BBCQ\
\newblock In {\Bem Proceedings of The 32nd International Conference on Machine
  Learning}, \lowercase{\BVOL}~3, \BPGS\ 2048--2057.

\bibitem[\protect\BCAY{Xu, Zhu, Shi, Zeng,\ \BBA\ Huang}{Xu
  et~al.}{2020a}]{xu-etal-2020-mixed}
Xu, R., Zhu, C., Shi, Y., Zeng, M., \BBA\ Huang, X. \BBOP2020a\BBCP.
\newblock \BBOQ Mixed-lingual pre-training for cross-lingual
  summarization\BBCQ\
\newblock In {\Bem Proceedings of the 1st Conference of the Asia-Pacific
  Chapter of the Association for Computational Linguistics and the 10th
  International Joint Conference on Natural Language Processing}, \BPGS\
  536--541, Suzhou, China. Association for Computational Linguistics.

\bibitem[\protect\BCAY{Xu, Li, Yuan, Wu, He,\ \BBA\ Zhou}{Xu
  et~al.}{2020b}]{xu-etal-2020-self}
Xu, S., Li, H., Yuan, P., Wu, Y., He, X., \BBA\ Zhou, B. \BBOP2020b\BBCP.
\newblock \BBOQ Self-attention guided copy mechanism for abstractive
  summarization\BBCQ\
\newblock In {\Bem Proceedings of the 58th Annual Meeting of the Association
  for Computational Linguistics}, \BPGS\ 1355--1362, Online. Association for
  Computational Linguistics.

\bibitem[\protect\BCAY{Xu\ \BBA\ Lapata}{Xu\ \BBA\
  Lapata}{2020}]{xu-lapata-2020-coarse}
Xu, Y.\BBACOMMA\  \BBA\ Lapata, M. \BBOP2020\BBCP.
\newblock \BBOQ Coarse-to-fine query focused multi-document summarization\BBCQ\
\newblock In {\Bem Proceedings of the 2020 Conference on Empirical Methods in
  Natural Language Processing (EMNLP)}, \BPGS\ 3632--3645, Online. Association
  for Computational Linguistics.

\bibitem[\protect\BCAY{Yan, Qi, Gong, Liu, Duan, Chen, Zhang,\ \BBA\ Zhou}{Yan
  et~al.}{2020}]{yan2020prophetnet}
Yan, Y., Qi, W., Gong, Y., Liu, D., Duan, N., Chen, J., Zhang, R., \BBA\ Zhou,
  M. \BBOP2020\BBCP.
\newblock \BBOQ Prophetnet: Predicting future n-gram for sequence-to-sequence
  pre-training\BBCQ\
\newblock {\Bem arXiv preprint arXiv:2001.04063}.

\bibitem[\protect\BCAY{Yang, Bao,\ \BBA\ Nenkova}{Yang
  et~al.}{2017}]{yang-etal-2017-detecting}
Yang, Y., Bao, F., \BBA\ Nenkova, A. \BBOP2017\BBCP.
\newblock \BBOQ Detecting (un)important content for single-document news
  summarization\BBCQ\
\newblock In {\Bem Proceedings of the 15th Conference of the {E}uropean Chapter
  of the Association for Computational Linguistics: Volume 2, Short Papers},
  \BPGS\ 707--712, Valencia, Spain. Association for Computational Linguistics.

\bibitem[\protect\BCAY{{Yang}, {Dai}, {Yang}, {Carbonell}, {Salakhutdinov},\
  \BBA\ {Le}}{{Yang} et~al.}{2019}]{yang2019xlnet}
{Yang}, Z., {Dai}, Z., {Yang}, Y., {Carbonell}, J., {Salakhutdinov}, R., \BBA\
  {Le}, Q.~V. \BBOP2019\BBCP.
\newblock \BBOQ Xlnet: Generalized autoregressive pretraining for language
  understanding\BBCQ\
\newblock In {\Bem NeurIPS 2019 : Thirty-third Conference on Neural Information
  Processing Systems}, \BPGS\ 5753--5763.

\bibitem[\protect\BCAY{Yin, Meng, Su, Ge, Song, Zhou,\ \BBA\ Luo}{Yin
  et~al.}{2020}]{yin2020enhancing}
Yin, Y., Meng, F., Su, J., Ge, Y., Song, L., Zhou, J., \BBA\ Luo, J.
  \BBOP2020\BBCP.
\newblock \BBOQ Enhancing pointer network for sentence ordering with pairwise
  ordering predictions\BBCQ\
\newblock In {\Bem Proceedings of the AAAI Conference on Artificial
  Intelligence}, \BPGS\ 9482--9489.

\bibitem[\protect\BCAY{You, Jia, Liu,\ \BBA\ Yang}{You
  et~al.}{2019}]{you-etal-2019-improving}
You, Y., Jia, W., Liu, T., \BBA\ Yang, W. \BBOP2019\BBCP.
\newblock \BBOQ Improving abstractive document summarization with salient
  information modeling\BBCQ\
\newblock In {\Bem Proceedings of the 57th Annual Meeting of the Association
  for Computational Linguistics}, \BPGS\ 2132--2141, Florence, Italy.
  Association for Computational Linguistics.

\bibitem[\protect\BCAY{{Zhang}, {Zhao}, {Saleh},\ \BBA\ {Liu}}{{Zhang}
  et~al.}{2020a}]{zhang2020pegasus}
{Zhang}, J., {Zhao}, Y., {Saleh}, M., \BBA\ {Liu}, P. \BBOP2020a\BBCP.
\newblock \BBOQ Pegasus: Pre-training with extracted gap-sentences for
  abstractive summarization\BBCQ\
\newblock In {\Bem ICML 2020: 37th International Conference on Machine
  Learning}.

\bibitem[\protect\BCAY{{Zhang}, {Kishore}, {Wu}, {Weinberger},\ \BBA\
  {Artzi}}{{Zhang} et~al.}{2020b}]{zhang2020bertscore}
{Zhang}, T., {Kishore}, V., {Wu}, F., {Weinberger}, K.~Q., \BBA\ {Artzi}, Y.
  \BBOP2020b\BBCP.
\newblock \BBOQ Bertscore: Evaluating text generation with bert\BBCQ\
\newblock In {\Bem ICLR 2020 : Eighth International Conference on Learning
  Representations}.

\bibitem[\protect\BCAY{Zhang, Wei,\ \BBA\ Zhou}{Zhang
  et~al.}{2019}]{zhang-etal-2019-hibert}
Zhang, X., Wei, F., \BBA\ Zhou, M. \BBOP2019\BBCP.
\newblock \BBOQ {HIBERT}: Document level pre-training of hierarchical
  bidirectional transformers for document summarization\BBCQ\
\newblock In {\Bem Proceedings of the 57th Annual Meeting of the Association
  for Computational Linguistics}, \BPGS\ 5059--5069, Florence, Italy.
  Association for Computational Linguistics.

\bibitem[\protect\BCAY{Zhao, Xu,\ \BBA\ Guo}{Zhao
  et~al.}{2020}]{zhao-etal-2020-improving}
Zhao, L., Xu, W., \BBA\ Guo, J. \BBOP2020\BBCP.
\newblock \BBOQ Improving abstractive dialogue summarization with graph
  structures and topic words\BBCQ\
\newblock In {\Bem Proceedings of the 28th International Conference on
  Computational Linguistics}, \BPGS\ 437--449, Barcelona, Spain (Online).
  International Committee on Computational Linguistics.

\bibitem[\protect\BCAY{Zhao, Peyrard, Liu, Gao, Meyer,\ \BBA\ Eger}{Zhao
  et~al.}{2019}]{zhao-etal-2019-moverscore}
Zhao, W., Peyrard, M., Liu, F., Gao, Y., Meyer, C.~M., \BBA\ Eger, S.
  \BBOP2019\BBCP.
\newblock \BBOQ {M}over{S}core: Text generation evaluating with contextualized
  embeddings and earth mover distance\BBCQ\
\newblock In {\Bem Proceedings of the 2019 Conference on Empirical Methods in
  Natural Language Processing and the 9th International Joint Conference on
  Natural Language Processing (EMNLP-IJCNLP)}, \BPGS\ 563--578, Hong Kong,
  China. Association for Computational Linguistics.

\bibitem[\protect\BCAY{Zheng\ \BBA\ Lapata}{Zheng\ \BBA\
  Lapata}{2019}]{zheng-lapata-2019-sentence}
Zheng, H.\BBACOMMA\  \BBA\ Lapata, M. \BBOP2019\BBCP.
\newblock \BBOQ Sentence centrality revisited for unsupervised
  summarization\BBCQ\
\newblock In {\Bem Proceedings of the 57th Annual Meeting of the Association
  for Computational Linguistics}, \BPGS\ 6236--6247, Florence, Italy.
  Association for Computational Linguistics.

\bibitem[\protect\BCAY{Zhong, Liu, Chen, Wang, Qiu,\ \BBA\ Huang}{Zhong
  et~al.}{2020}]{zhong-etal-2020-extractive}
Zhong, M., Liu, P., Chen, Y., Wang, D., Qiu, X., \BBA\ Huang, X.
  \BBOP2020\BBCP.
\newblock \BBOQ Extractive summarization as text matching\BBCQ\
\newblock In {\Bem Proceedings of the 58th Annual Meeting of the Association
  for Computational Linguistics}, \BPGS\ 6197--6208, Online. Association for
  Computational Linguistics.

\bibitem[\protect\BCAY{Zhong, Liu, Wang, Qiu,\ \BBA\ Huang}{Zhong
  et~al.}{2019}]{zhong-etal-2019-searching}
Zhong, M., Liu, P., Wang, D., Qiu, X., \BBA\ Huang, X. \BBOP2019\BBCP.
\newblock \BBOQ Searching for effective neural extractive summarization: What
  works and what{'}s next\BBCQ\
\newblock In {\Bem Proceedings of the 57th Annual Meeting of the Association
  for Computational Linguistics}, \BPGS\ 1049--1058, Florence, Italy.
  Association for Computational Linguistics.

\bibitem[\protect\BCAY{Zhou\ \BBA\ Rush}{Zhou\ \BBA\
  Rush}{2019}]{zhou-rush-2019-simple}
Zhou, J.\BBACOMMA\  \BBA\ Rush, A. \BBOP2019\BBCP.
\newblock \BBOQ Simple unsupervised summarization by contextual matching\BBCQ\
\newblock In {\Bem Proceedings of the 57th Annual Meeting of the Association
  for Computational Linguistics}, \BPGS\ 5101--5106, Florence, Italy.
  Association for Computational Linguistics.

\bibitem[\protect\BCAY{Zhu, Wang, Wang, Zhou, Zhang, Wang,\ \BBA\ Zong}{Zhu
  et~al.}{2019}]{zhu-etal-2019-ncls}
Zhu, J., Wang, Q., Wang, Y., Zhou, Y., Zhang, J., Wang, S., \BBA\ Zong, C.
  \BBOP2019\BBCP.
\newblock \BBOQ {NCLS}: Neural cross-lingual summarization\BBCQ\
\newblock In {\Bem Proceedings of the 2019 Conference on Empirical Methods in
  Natural Language Processing and the 9th International Joint Conference on
  Natural Language Processing (EMNLP-IJCNLP)}, \BPGS\ 3054--3064, Hong Kong,
  China. Association for Computational Linguistics.

\bibitem[\protect\BCAY{Zhu, Zhou, Zhang,\ \BBA\ Zong}{Zhu
  et~al.}{2020}]{zhu-etal-2020-attend}
Zhu, J., Zhou, Y., Zhang, J., \BBA\ Zong, C. \BBOP2020\BBCP.
\newblock \BBOQ Attend, translate and summarize: An efficient method for neural
  cross-lingual summarization\BBCQ\
\newblock In {\Bem Proceedings of the 58th Annual Meeting of the Association
  for Computational Linguistics}, \BPGS\ 1309--1321, Online. Association for
  Computational Linguistics.

\bibitem[\protect\BCAY{Zou, Zhang, Lu, Wei,\ \BBA\ Zhou}{Zou
  et~al.}{2020}]{zou-etal-2020-pre}
Zou, Y., Zhang, X., Lu, W., Wei, F., \BBA\ Zhou, M. \BBOP2020\BBCP.
\newblock \BBOQ Pre-training for abstractive document summarization by
  reinstating source text\BBCQ\
\newblock In {\Bem Proceedings of the 2020 Conference on Empirical Methods in
  Natural Language Processing (EMNLP)}, \BPGS\ 3646--3660, Online. Association
  for Computational Linguistics.

\end{thebibliography}
\bibliographystyle{theapa}

\end{document}